\documentclass[10pt, journal, compsoc]{IEEEtran}
\IEEEoverridecommandlockouts
\usepackage[english]{babel}
\usepackage[utf8]{inputenc}
\ifCLASSOPTIONcompsoc
\usepackage[nocompress]{cite}
\else
\usepackage{cite}
\fi
\usepackage{amsmath,amssymb,amsfonts}
\usepackage{mathtools}
\usepackage{algpseudocode} \usepackage{multirow}
\usepackage{multicol}
\usepackage{makecell}
\usepackage{subcaption}
\usepackage[dvipsnames, table]{xcolor}
\usepackage{makecell}
\usepackage{tabularx}
\usepackage{babel}
\usepackage{balance}

\usepackage[hyphens, spaces]{url}

\usepackage{tikz}
\usetikzlibrary{arrows.meta, calc, fit, shapes, decorations.pathmorphing, backgrounds, shapes.geometric}

\usepackage{graphicx}
\usepackage{textcomp}
\usepackage{adjustbox}
\usepackage{pifont}
\usepackage[hidelinks,
    pdftitle={On the Robustness of Random Forest Against Untargeted Data Poisoning: An Ensemble-Based Approach},
    pdfauthor={Marco Anisetti, Claudio A. Ardagna, Alessandro Balestrucci, Nicola Bena, Ernesto Damiani, Chan Yeob Yeun}]{hyperref}
\usepackage[hyphenbreaks]{breakurl}
\usepackage[shortcuts]{extdash}
\usepackage[inline]{enumitem}
\setlist*[enumerate]{label=\emph{\roman*)}}

\def\BibTeX{{\rm B\kern-.05em{\sc i\kern-.025em b}\kern-.08em
    T\kern-.1667em\lower.7ex\hbox{E}\kern-.125emX}}

\makeatletter\begin{document}

\title{On the Robustness of Random Forest Against Untargeted Data Poisoning: An Ensemble-Based Approach}

\author{Marco Anisetti, Claudio A. Ardagna, Alessandro Balestrucci, Nicola Bena,\\Ernesto Damiani, Chan Yeob Yeun
\IEEEcompsocitemizethanks{
    \IEEEcompsocthanksitem Marco Anisetti, Claudio A. Ardagna, Nicola Bena, Ernesto Damiani are with the Department of Computer Science, Università degli Studi di Milano, Milan, Italy.\protect\\
    E-mail: \{marco.anisetti, claudio.ardagna, nicola.bena, ernesto.damiani\}@unimi.it
\IEEEcompsocthanksitem Alessandro Balestrucci is with Consorzio Interuniversitario per l'Informatica, Rome, Italy.\protect\\
    E-mail: alessandro.balestrucci@consorzio-cini.it
\IEEEcompsocthanksitem Ernesto Damiani, Chan Yeob Yeun are with Khalifa University of Science and Technology, Abu Dhabi, UAE.\protect\\
    E-mail:  \{ernesto.damiani, chan.yeun\}@ku.ac.ae
    }
}

\markboth{Accepted for publication in IEEE Transactions on Sustainable Computing; DOI: \href{https://doi.org/10.1109/TSUSC.2023.3293269}{10.1109/TSUSC.2023.3293269}}{DOI: \href{https://doi.org/10.1109/TSUSC.2023.3293269}{10.1109/TSUSC.2023.3293269}} 

\IEEEtitleabstractindextext{\begin{abstract}
    Machine learning is becoming ubiquitous. From finance to medicine, machine learning models are boosting decision\-/making processes and even outperforming humans in some tasks. This huge progress in terms of prediction quality does not however find a counterpart in the security of such models and corresponding predictions, where perturbations of fractions of the training set (poisoning) can seriously undermine the model accuracy. Research on poisoning attacks and defenses received increasing attention in the last decade, leading to several promising solutions aiming to increase the robustness of machine learning. Among them, ensemble-based defenses, where different models are trained on portions of the training set and their predictions are then aggregated, provide strong theoretical guarantees at the price of a linear overhead. Surprisingly, ensemble-based defenses, which do not pose any restrictions on the base model, have not been applied to increase the robustness of random forest models. The work in this paper aims to fill in this gap by designing and implementing a novel hash-based ensemble approach that protects random forest against untargeted, random poisoning attacks. An extensive experimental evaluation measures the performance of our approach against a variety of attacks, as well as its sustainability in terms of resource consumption and performance, and compares it with a traditional monolithic model based on random forest. A final discussion presents our main findings and compares our approach with existing poisoning defenses targeting random forests.
\end{abstract}

}

\maketitle
\IEEEpeerreviewmaketitle

\tikzset{
    every node/.style={
        font=\scriptsize
    },
    dataset/.style={
        draw,
        cylinder,
        shape border rotate=90,
align=center,
    },
on path/.style={
above
    },
    rf/.style={
        draw,
        rounded corners,
        double
    },
    test path/.style={
        dashed
    },
    connections/.style={
        very thin
    },
    partition/.style={
        draw,
        cylinder,
        shape border rotate=90,
        align=center,
        minimum height=.6cm,
        minimum width=.35cm,
    },
    link/.style={
        >=Latex
    },
    container/.style={
        very thin,
        draw
    },
    poisoning link/.style={
        link,
        decorate,
ultra thin,
    },
    testing link/.style={},
    rf container/.style={
        dashed,
        ultra thin,
        draw,
        rounded corners,
        inner sep=2pt
    },
    poisoning link/.style={},
    clean link/.style={},
    to be crossed link/.style={},
    arrow between rf/.style={
        double distance=3pt,
        dashed,
        arrows={-Latex[fill=white]}
    },
    accuracy link/.style={}
}

\tikzset{
pics/test set/.style n args={2}{
        code= {
            \node[dataset, minimum height=.7cm, minimum width=.4cm] (test set-#1) at (#2) {};

            \node[] at ([yshift=3]test set-#1.north) (test set text-#1) {\makecell{Test Set}};
        }
    }
}

\tikzset{
pics/our approach training/.style n args={3}{
        code={
            \node[dataset, minimum height=1cm, minimum width=.6cm] (dataset-#1) {};

            \node[] at ([yshift=5]dataset-#1.north) (dataset text-#1) {\makecell{Dataset}};
        
            \node[dataset, minimum height=.8cm, minimum width=.5cm] (training set-#1) at ([xshift=25, yshift=#2]dataset-#1.east) {};
        
            \node[] at ([yshift=-8]training set-#1.south) (training set text-#1) {\makecell{Training\\Set}};

            \pic{test set={#1}{[xshift=25, yshift=#3]dataset-#1.east}};
            
            \draw[->, link, name=from dataset-#1 to training set-#1] (dataset-#1) to (training set-#1.west);
        
            \draw[->, link, name=from dataset-#1 to test set-#1] (dataset-#1) to (test set-#1.west);
        }
    },
}

\tikzset{
pics/rf with majority voting/.style n args={4}{
        code={
            \foreach \rfnumber/\yshi/\content in {1/25/1, 2/0/2, 3/-30/\nbasemodel}{

                \node[rf] at ([xshift=#3, yshift=\yshi]#2.east) (rf\rfnumber-#1) {\makecell{$\text{RF}_{\content}$}};

            }

            \ifthenelse{\equal{#4}{true}}{

                \node[isosceles triangle, draw] at ([xshift=25]rf2-#1.east) (mj-#1) {};

                \node[]  at ([yshift=9, xshift=5]mj-#1.north) (mj text-#1) {\makecell{Majority\\Voting}};

                \foreach \rfnumber in {1, 2, 3} {
                        \draw[->, link, name=from rf\rfnumber-#1] (rf\rfnumber-#1) to (mj-#1);

                        \draw[->, link, name=from anchor to rf\rfnumber-#1] (#2.center) to (rf\rfnumber-#1);
                }
            }

            \node[] at ($(rf3-#1)!0.5!(rf2-#1)$) (additional rf-#1) {\ldots};

            \node[fit=(rf1-#1)(rf2-#1)(rf3-#1), rf container] (rf container-#1) {};
            
        }
    }
}

\tikzset{
pics/rf mono/.style n args={2}{
        code={
            \node[rf] at (#2) (rf-mono poisoned-#1) {\makecell{RF}};

            \node[rf] at ([yshift=-15]rf-mono poisoned-#1.south) (rf-mono clean-#1) {\makecell{RF}};

            \node[rf container, fit=(rf-mono poisoned-#1)(rf-mono clean-#1)] (rf-mono container-#1) {};
        }
    }
}

\tikzset{
pics/ensemble approach/.style n args={4}{
        code={
            \node[draw, rounded corners] at ([xshift=10]#2.east) (hash-#1) {hash\%\nbasemodel};

\foreach \pnumber/\yshi/\content in {1/25/1, 2/0/2, 3/-32.5/\nbasemodel} {
                \node[partition] at ([xshift=20, yshift=\yshi]hash-#1.east) (p\pnumber-#1) {};
            }

            \pic{rf with majority voting={#1}{p2-#1}{#3}{false}};

\foreach \x/\yshift in {1/0, 2/0, 3/7.5}{
                \draw[->, link] (hash-#1) to (p\x-#1);
                \draw[->, link] ([yshift=-\yshift]p\x-#1) to (rf\x-#1);
            }

            \begin{scope}[on background layer]
                \node[container, fit=(p1-#1)(p2-#1)(p3-#1)(rf1-#1)(rf2-#1)(rf3-#1)(#2)] (approach-#1) {};
            \end{scope}

            \node[] (approach text-#1) at ([yshift=3]approach-#1.north) {\makecell{#4}};

            \node[] at ($(p2-#1)!.4!(p3-#1)$) {\ldots};
        
        }  

    }
}

\DeclarePairedDelimiterX{\norm}[1]{\lVert}{\rVert}{#1}

\newcommand{\ok}{\ding{51}}
\newcommand{\notok}{\ding{55}}

\newcommand{\trainingset}{\ensuremath{D}}
\newcommand{\poisonedtrainingset}{\ensuremath{\widetilde{D}}}

\newcommand{\extentpoint}{\ensuremath{\epsilon_{\text{p}}}}
\newcommand{\extentfeature}{\ensuremath{\epsilon_{\text{f}}}}

\newcommand{\lines}[2]{\ensuremath{#1}--\ensuremath{#2}}

\newcommand{\nbasemodel}{\ensuremath{N}}

\newcommand{\numberofmodels}{\ensuremath{N}}

\newcommand{\poisoned}[1]{\ensuremath{\widetilde{#1}}}

\newcommand{\dataset}{\ensuremath{D}}
\newcommand{\datasetpoisoned}{\ensuremath{\poisoned{\dataset}}}

\newcommand{\point}{\ensuremath{p}}
\newcommand{\pointpoisoned}{\ensuremath{\widetilde{p}}}

\newcommand{\deltaf}{\ensuremath{\Delta}}

\newcommand{\accuracysymbol}{\ensuremath{\text{ACC}}}
\newcommand{\accuracyfn}[1]{\ensuremath{\accuracysymbol(#1)}}
\newcommand{\accuracyclean}{\accuracyfn{\dataset}}
\newcommand{\accuracypoisoned}{\accuracyfn{\datasetpoisoned}}

\newcommand{\perturbation}[1]{\emph{#1}}
\newcommand{\zeroing}{\perturbation{zeroing}}
\newcommand{\noising}{\perturbation{noising}}
\newcommand{\outofranging}{\perturbation{out\-/of\-/ranging}}
\newcommand{\labelflipping}{\perturbation{label flipping}}

\newcommand{\datasetabbrevandroid}{AM}
\newcommand{\datasetabbrevmusk}{M2}

\newcommand{\headerEnsemblePoisoned}{Poisoned Hash\-/Based Ensemble}
\newcommand{\headerEnsembleNonPoisoned}{Non\-/Poisoned Hash\-/Based Ensemble}
\newcommand{\headerEnsembleDefault}{Hash\-/Based Ensemble}

\newcommand{\features}{\ensuremath{f}}
\newcommand{\featuresCount}{\ensuremath{\vert\features\vert}}
\newcommand{\dataPointsCount}{\ensuremath{\vert\dataset\vert}}

\newcommand{\domainTypeStyle}[1]{#1}
\newcommand{\domainTypeMulti}{\domainTypeStyle{Multi}}
\newcommand{\domainTypeSingle}{\domainTypeStyle{Single}}

\newcommand{\strategyPositiveSymbol}{S}
\newcommand{\strategyRandom}{\ensuremath{\lnot}\strategyPositiveSymbol}
\newcommand{\strategyNonRandom}{\strategyPositiveSymbol}

\newcommand{\targetedPositiveSymbol}{T}
\newcommand{\typeTargeted}{\targetedPositiveSymbol}
\newcommand{\typeUnTargeted}{\ensuremath{\lnot}\targetedPositiveSymbol}

\newcommand{\mycite}[1]{[#1]} \IEEEraisesectionheading{\section{Introduction}\label{sec:intro}}

\IEEEPARstart{W}{ith} the introduction of deep neural networks in the last decade, machine learning (ML) is now leaving academia and powering an increasing number of applications, from finance~\cite{app9245574} to smart grid~\cite{8625421}, weather forecast~\cite{atmos11070676}, signal processing~\cite{mio2019signal}, and medicine~\cite{KOUROU20158,KatraAnisetti22}.
Machine learning models are reportedly performing  better than humans in some specific tasks~\cite{Richens2020}, with an increasing adoption even in safety-critical application scenarios.

In this context, it is of paramount importance to properly evaluate and protect the security of ML models. As such, one of the most relevant threat vectors are data, being ML models trained on (very) large datasets. In particular, \emph{poisoning attacks} include attacks carried out at training time by maliciously altering the training set, with the aim of decreasing the overall classification accuracy, or misclassifying some specific inputs when the model is deployed. Poisoning attacks have been reported in several application scenarios, from malware detection~\cite{sma2020rf} to biometrics~\cite{chen2017targeted}, healthcare~\cite{6868201}, and source code completion~\cite{263874} and against several types of machine learning models, from support vector machines~\cite{pmlr-v20-biggio11} to decision trees~\cite{6868201}, random forests~\cite{sma2020rf}, and neural networks~\cite{su12166434}, to name but a few.
Solutions counteracting poisoning are vary, and range from improving the poisoned dataset by removing or repairing (suspicious) data points~\cite{8489495, pmlr-v97-diakonikolas19a, 10.5555/3367471.3367701, pmlr-v119-rosenfeld20b, 10.1007/978-3-030-66415-2_4, Koh2021} to strengthening the ML model itself, to make it more resistant to poisoning~\cite{jia2021intrinsic, levine2021deep, wang2022improved}.
Among the model strengthening solutions, ensemble is mostly studied in deep learning and image recognition~\cite{jia2021intrinsic, levine2021deep, wang2022improved, pmlr-v162-chen22k}. It consists of training several ML models on different (possibly partially overlapped) partitions of the training set, and then aggregating their predictions in a single one. This solution stands out for its ability to provide a theoretical bound on the correctness of the prediction according to the extent of poisoning. Although simple in design and implementation, it often builds on a large number (hundreds or thousands) of base models (e.g.,~\cite{levine2021deep}), leaving open questions on their sustainability.

Surprisingly, although random forests are one of the most adopted models on tabular datasets~\cite{why} and have been deeply studied from several perspectives, such as explainability~\cite{10.1145/2939672.2939778}, fairness~\cite{10.1145/3540250.3549093}, and sustainability~\cite{9923840}, their robustness has been barely analyzed. Existing works focused on robustness against traditional poisoning attacks~\cite{su12166434,Zhang2021,VERDE20212624,9652959,10.1145/3468218.3469050,YERLIKAYA2022118101} and defenses that aim to repair poisoned datasets~\cite{Taheri2020, https://doi.org/10.48550/arxiv.2208.08433} in limited settings. 
No model strengthening solutions, including ensemble-based defenses, have been proposed.

Our paper aims to fill in these research gaps in a novel scenario that targets a sustainable and scalable robustness approach for random forest, assuming poisoning attacks that can be implemented by attackers with little to none knowledge and resources (Section~\ref{sec:attack}). In particular, we propose a novel hash\-/based ensemble approach and empirically evaluate its robustness and sustainability against untargeted, random data poisoning attacks to the accuracy of random forest.
Our hash-based ensemble is based on hash functions to route data points in the original training set in different partitions used to train different models in the ensemble. Our implementation extends the well\-/studied ensemble in~\cite{levine2021deep, pmlr-v162-chen22k} as follows: \emph{i)} each model in the ensemble is trained on a disjoint partition of the training set to which data points are assigned according to hashing, and \emph{ii)} the final prediction is retrieved according to majority voting.  
Contrary to state of the art, our paper evaluates ensembles of small to moderate size (i.e., up to $21$ random forests), targeting sustainability of defense. Throughout fine\-/grained experiments, we show that even the simplest label flipping attack carried out with no knowledge or strategy can significantly undermine plain random forests' performance, while consistently with results in literature~\cite{su12166434}, random forests are almost insensitive to other perturbations. In addition, we show that the usage of even the smallest ensemble does protect from label flipping, while providing a sustainable approach in terms of required resources (CPU and RAM) and performance (execution time). 

Our contribution is twofold. We first design and develop a sustainable hash-based ensemble approach extending~\cite{levine2021deep, pmlr-v162-chen22k} to increase the robustness of random forest against untargeted, random poisoning attacks; according to our knowledge, this is the first defense based on model strengthening that is applied on random forest. We then evaluate the robustness in terms of accuracy variation according to several untargeted poisoning perturbations, and corresponding sustainability comparing the performance and resource demands of our approach and a plain random forest.

The remainder of this paper is organized as follows.
Section~\ref{sec:background} discusses the background and state of the art in the context of poisoning attacks and defenses.
Section~\ref{sec:ensemble} presents an overview of our approach based on an ensemble of random forests, whose robustness and sustainability is evaluated according to the threat model in Section~\ref{sec:attack}.
Section~\ref{sec:exp-process} describes the evaluation process and target datasets, while Section~\ref{sec:exp-result} details the results of such a process.
Section~\ref{subsec:exp-resource} discusses the sustainability of our approach.
Section~\ref{sec:discussion} discusses our main findings, while Section~\ref{sec:comparison} presents a comparison with approaches in literature. Section~\ref{sec:conclusion} draws our concluding remarks. \section{Background and Related Work}\label{sec:background}
The research community has worked hard to strengthen the security of machine learning (ML) models~\cite{damiani2020certified, anisetti2020methodology, 10.1145/3446331, 10.1145/3585385} against different categories of attacks that can be classified according to the stage where they occur. On one side, \emph{adversarial attacks} occur at inference time and consist of specially\-/crafted data points that are routed to the ML model to cause a faulty or wrong inference. Their goal is the misclassification of such data points.
On the other side, \emph{poisoning attacks}, the focus of this paper, occur at training time and inject poisoned data points in the dataset. They aim to reduce the accuracy of the model or cause the misclassification of specific data points at inference time. 

\vspace{.5em}

\noindent \textbf{Poisoning attacks} alter the dataset with malicious data points. They are created by perturbing existing data points in terms of \begin{enumerate*}
    \item samples or values of the features~\cite{42503};
    \item labels, having the advantage of not creating anomalous, or at least suspicious, data points~\cite{pmlr-v20-biggio11, zhang2017understanding, paudice2018label}.
\end{enumerate*}
Perturbations can be crafted according to a specific goal such as \begin{enumerate*}
    \item misclassification of \emph{positive} data points, for instance in spam detection (\emph{targeted poisoning}), requiring sophisticated perturbations such as \emph{feature collision}~\cite{NEURIPS2018_22722a34}; or
    \item accuracy reduction (\emph{untargeted poisoning}).
\end{enumerate*}
The latter often corresponds to random perturbations~\cite{pmlr-v20-biggio11} and is the focus of this paper.

\vspace{.5em}

\noindent \textbf{Defenses against poisoning attacks} can be performed in two main ways: \emph{dataset strengthening} or \emph{model strengthening}; for other approaches, we refer the reader to~\cite{10.1145/3585385}.
Dataset strengthening aims to increase the quality of the dataset by removing or sanitizing poisoned data points, detected with some heuristics.
The latter are based on outlier identification~\cite{4531146, 8489495, paudice2018label, 10.1007/978-3-030-66415-2_4, Koh2021} and the evaluation of the impact of data points on the ML model~\cite{Barreno2010, pmlr-v97-diakonikolas19a, prasad2020robust}, to name but a few. 
Sanitization techniques include \emph{randomized smoothing} and \emph{differential privacy}. In randomized smoothing, each data point is \emph{smoothed} (i.e., its label is replaced) according to its neighbor data points. Smoothing has been initially proposed to counteract inference\-/time attacks~\cite{biggio2013evasion}, and then adapted to poisoning~\cite{pmlr-v119-rosenfeld20b}. Similarly, in differential privacy, noise is added during training such that the predictions done by a model trained on the original dataset are indistinguishable from those of a model trained on the corresponding poisoned dataset, up to a certain $\epsilon$~\cite{10.5555/3367471.3367701, hong2020effectiveness}.

Model strengthening aims to increase the robustness of the ML model by altering the model itself, such that the effect of poisoning is reduced. Among them, we focus on simple yet effective ensemble approaches, where the monolithic ML model is replaced by a (large) ensemble of the same model~\cite{jia2021intrinsic, levine2021deep, wang2022improved, pmlr-v162-chen22k}. This technique, evaluated mostly on neural network\-/based models, splits the training set in different partitions according to some strategies, and each partition is used as the training set of a model of the ensemble. Intuitively, this reduces the influence of poisoned data points, since each model is trained on a smaller fraction of poisoned data points.

Some of the above techniques, including ensemble~\cite{jia2021intrinsic, levine2021deep, wang2022improved, pmlr-v162-chen22k}, randomized smoothing~\cite{pmlr-v119-rosenfeld20b}, and differential privacy~\cite{10.5555/3367471.3367701}, can provide a certifiable guarantee such that the model prediction is correct up to a certain amount of poisoning on the dataset.

Poisoning defenses often add a not\-/negligible resources overhead over training and inference, two procedures that by themselves are significantly resource\-/intensive. For instance, dataset strengthening techniques require training of additional (un)supervised models~\cite{8489495} or nearest neighbor search~\cite{10.1007/978-3-030-66415-2_4}, while model strengthening techniques require very large ensembles of base models~\cite{jia2021intrinsic, levine2021deep, wang2022improved, pmlr-v162-chen22k}, often without a performance analysis.

\vspace{.5em}

\noindent \textbf{Poisoning attacks and defenses on random forest}, the target of this paper, have been only partially investigated.
In terms of attacks, existing work focuses on poisoning attacks that alter either labels~\cite{su12166434,Taheri2020,Zhang2021,YERLIKAYA2022118101, https://doi.org/10.48550/arxiv.2208.08433,10020528} or features~\cite{VERDE20212624, 9652959, 10.1145/3468218.3469050}.
The most relevant finding is that random forests are more resilient to poisoning than other types of ML models~\cite{Zhang2021, YERLIKAYA2022118101}.
In terms of defenses, there exist only two papers studying dataset strengthening solutions~\cite{Taheri2020,https://doi.org/10.48550/arxiv.2208.08433}, while no papers presented solutions based on model strengthening, including ensembles.
The approach in this paper departs from traditional solutions where the attacker and the defender can perform sophisticated and resource\-/intensive attacks and defenses (e.g.,~\cite{10.1007/978-3-030-66415-2_4,levine2021deep}); it rather aims to provide a robust, while sustainable, model strengthening defense for random forest against untargeted training\-/time poisoning of labels and features. To the best of our knowledge, this is the first model strengthening defense applied on random forest; a more detailed comparison with the state of the art can be found in Section~\ref{sec:comparison}.

 \section{Our Approach}\label{sec:ensemble}

\begin{figure}[t]
    \centering
    \begin{adjustbox}{max totalsize={.5\textwidth}{\textheight},center}
        \begin{tikzpicture}
    
    \pic{our approach training={1}{-42}{42}};

    \node[draw, rounded corners] at ([xshift=25]training set-1.east) (poisoning) {\makecell{Poisoning}};

    \node[ ] (anchor) at ([xshift=10]poisoning.east) {};
    \draw[-] (poisoning) to (anchor);

    \pic{ensemble approach={1-train}{anchor}{20}{\headerEnsembleDefault}};

    \draw[-, link] (training set-1) to (poisoning);

\node[] (anchor for test) at (test set-1.north east -| p2-1-train) {};

    \pic{rf with majority voting={1-test}{anchor for test}{20}{true}};
    \draw[-, link] (test set-1.north east) to (anchor for test.center);

    \draw[->, arrow between rf, bend right=60, name=from rf-train to rf-test] (rf container-1-train.east) to ([yshift=10]rf container-1-test.south east);

\end{tikzpicture}     \end{adjustbox}
    \caption{Overview of our hash-based ensemble approach.}
    \label{fig:ensembleapp}
\end{figure}
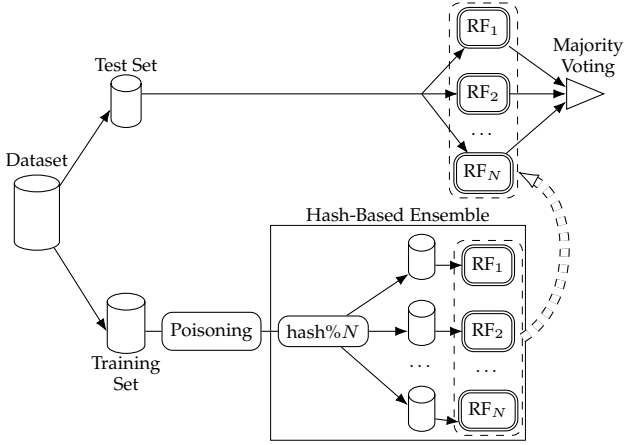

Figure~\ref{fig:ensembleapp} shows an overview of our hash-based ensemble approach that aims to increase the robustness of random forest against poisoning attacks. It first splits the tabular dataset in two parts forming the training set (\emph{Training Set} in Figure~\ref{fig:ensembleapp}) and the test set (\emph{Test Set} in Figure~\ref{fig:ensembleapp}). The training set is then poisoned according to our threat model in Section~\ref{sec:attack} (\emph{Poisoning} in Figure~\ref{fig:ensembleapp}). Based on the work in~\cite{levine2021deep, pmlr-v162-chen22k}, the training set is split in \numberofmodels\ disjoint partitions using a hash function (\emph{Hash-Based Ensemble} in Figure~\ref{fig:ensembleapp}), with \numberofmodels\ being the number of random forest in the ensemble, according to the following steps: \begin{enumerate*}
    \item the hash value of each data point of the training set is retrieved according to a given hash function;
    \item the modulo operator (modulo \numberofmodels) is applied on the corresponding hash value (\emph{hash \% \numberofmodels} in Figure~\ref{fig:ensembleapp});
    \item each data point is routed to a partition of the training set according to the modulo operator on the corresponding hash value (e.g., data points whose hash value modulo \numberofmodels\ is $0$ are assigned to partition $1$);
    \item the $i$\-/th training set of each random forest $RF_i$ is created by evenly taking data points from each partition, such that the training sets are disjoint, have the same cardinality, and balance the contribution from the different partitions in term of data points.
\end{enumerate*}
Each random $RF_i$ is then independently trained on the corresponding $i$\-/th training set.
At testing and inference time, data points in the test set are fed to each model $RF_1,$ $RF_2,$ \ldots$,$ $RF_n$ and the final prediction is retrieved according to majority voting.

Beyond being applied on random forest, our ensemble approach and its evaluation departs from existing works in the literature (Section~\ref{sec:background}) according to the following characteristics.

\begin{itemize}
    \item \emph{Additional round\-/robin training set creation}: most hash\-/based ensembles in literature (e.g.,~\cite{levine2021deep, pmlr-v162-chen22k}) consider one hash function plus a modulo operator only, except few special cases~\cite{wang2022improved}. We instead propose an additional phase, where data point assignment follows hash and modulo operations to increase diversity and ensure equally\-/sized training sets.
    \item \emph{Small number of base models}: existing ensemble\-/based defenses in neural networks require a large number of partitions and base models (e.g.,~\cite{levine2021deep}). We instead consider smaller numbers (up to $21$) to increase sustainability, while maintaining a good degree of protection.
    \item \emph{Tabular datasets for binary classification}: most of attacks and defenses are evaluated in image-based scenarios where image datasets are given as input to the models~\cite{levine2021deep}. We instead consider tabular datasets for binary classification, which are still a significant portion of ML.
    \item \emph{Untargeted poisoning}: most of defenses are evaluated against targeted poisoning (e.g.,~\cite{10.1007/978-3-030-66415-2_4}), where few specially\-/crafted data points are injected in the training set. We instead consider a threat model where an attacker with limited knowledge and resources randomly alters the dataset to reduce the accuracy of the resulting model (see Section~\ref{sec:attack}).
\end{itemize}

We note that our approach has been designed and developed to be sustainable, requiring a low amount of resources, since:
\begin{enumerate*}
    \item it is based on a limited number of base models in the ensemble;
    \item it does not involve any additional resource\-/intensive computations, such as the training of additional models other than the random forests in the ensemble; \item it uses a hash\-/based data point assignment, with hash functions being notoriously fast and lightweight; \item it trains \nbasemodel\ random forests independently on disjoint partitions, that is, the cardinality of the dataset is not increased, while training can be parallelized to reduce training time.
\end{enumerate*}

 \section{Threat Model}\label{sec:attack}
Our threat model considers a novel scenario where attackers need to cope with limited knowledge and resources. The attacker departs from targeted attacks and executes untargeted poisoning to reduce the accuracy of the ML model. To this aim, she randomly alters the dataset up to a given amount of manipulated features and labels. Specifically, the attacker implements different perturbations acting on features (\emph{zeroing, noising, out-of-ranging}) and labels (\emph{label flipping}), each implementing a specific poisoning attack that is tested independently.
Each perturbation takes as input a training set (denoted as \dataset), the percentage of data points and features to alter (denoted as \extentpoint\ and \extentfeature, respectively) according to the specific perturbation, and returns as output the poisoned training set (denoted as \datasetpoisoned). 
The poisoned training set is then partitioned in disjoint training sets, each used to train a model of the ensemble, according to the evaluation process in Section~\ref{subsec:exp-process-benchmark-process}.
We note that each perturbation randomly selects the data points and the corresponding features to poison according to \extentpoint\ and \extentfeature. In particular, the selected features are the same for every perturbation, to ensure proper comparison.

Let us consider as an example a binary classification task (classes $0$ and $1$), and a $5$\-/feature data point \point\ with value $\langle 0, 10, 15, 0, 1\rangle_{0}$, where the subscript indicates the label and the second feature (10) is the target of poisoning. 

Perturbation \emph{zeroing} produces a poisoned training set \datasetpoisoned, where the selected data points are perturbed by changing the values of the selected features to $0$. For instance, the poisoned data point of \point\ has value $\langle 0, \mathbf{0}, 15, 0, 1\rangle_{0}$.

Perturbation \emph{noising} produces a poisoned training set \datasetpoisoned, where the selected data points are perturbed by replacing the values of the selected features with a value within the distribution of the same feature in the opposite class. For instance, let us consider the second feature of data point \point. It takes value in $[0,  10]$ for class $0$, and $[20, 40]$ for class $1$. For instance, the poisoned data point of \point\ has value $\langle 0, \mathbf{37}, 15, 0, 1\rangle_{0}$.

Perturbation \emph{out\-/of\-/ranging} produces a poisoned training set \poisonedtrainingset, where the selected data points are perturbed by changing the value of the selected features with values outside their valid range. 
For instance, the poisoned data point of \point\ has value $\langle 0, \mathbf{-1}, 15, 0, 1\rangle_{0}$.

Perturbation \emph{label flipping} produces a poisoned training set \poisonedtrainingset, where the selected data points are perturbed by flipping their labels. For instance, the poisoned data point of \point\ has value $\langle 0, 10, 15, 0, 1\rangle_{\mathbf{1}}$.

\vspace{.5em}

We note that the effectiveness of these perturbations strongly depends on the data points \emph{actually} perturbed.
For instance, let us consider perturbation \emph{zeroing}. Assuming that the first feature of data point $\point$ (with value $0$) is selected for poisoning, the corresponding poisoned data point is not altered.
We also note that the threat model in this paper follows the general trend of cybersecurity attacks, where the danger mostly comes from unsophisticated yet impactful attacks~\cite{enisaetl2022}.

 \section{Evaluation Process}\label{sec:exp-process}

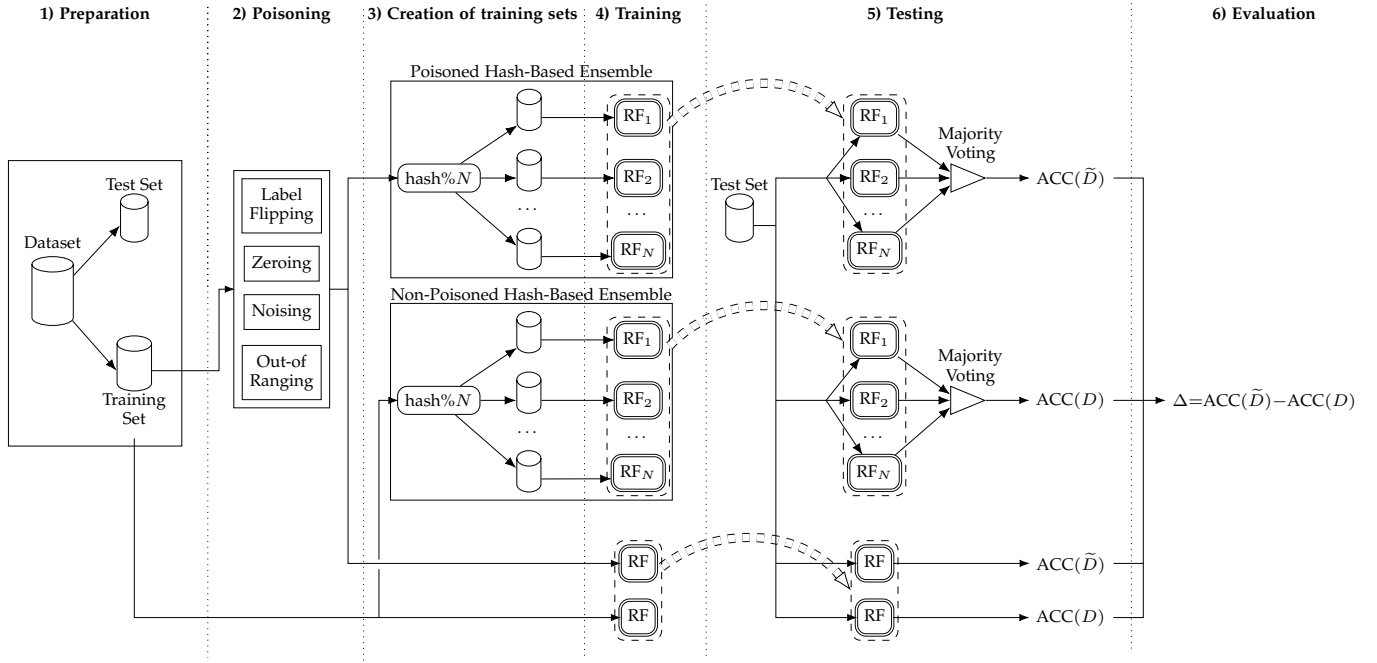
\begin{figure*}[!ht]
    \centering
    \begin{adjustbox}{max totalsize={.99\textwidth}{\textheight},center}
        \begin{tikzpicture}
    \tikzset{
        poisoning attack/.style={
            draw,
},
        super container/.style={
            draw,
            dashed
        },
        low link/.style={
            link,
            ultra thin
        },
        bar/.style={
            thin,
            dotted
        },
        from test/.style={
            ultra thin
        },
}

    \pic{our approach training={1}{-30}{30}};

    \node[container, fit=(dataset-1)(training set-1)(test set-1)(training set text-1)(test set text-1)(dataset text-1)] (container dataset) {};

    \node[poisoning attack] at ([xshift=55, yshift=2.25]test set-1.north east) (flipping) {\makecell{Label\\Flipping}};

    \foreach \offset/\content/\name in {12.5/Zeroing/zeroing, 32.5/Noising/noising, 57.5/{Out\-/of\\Ranging}/oof} {
        \node[poisoning attack] at ([yshift=-\offset]flipping.south) (\name) {\makecell{\content}};
    }

    \node[container, fit=(flipping)(oof)] (container poisoning) {};

    \node[] (anchor for train poisoned) at ([xshift=35, yshift=11.5]flipping.east) {};

    \pic{ensemble approach={poisoned}{anchor for train poisoned}{40}{\headerEnsemblePoisoned}};

    \node[] (anchor for train clean) at ([xshift=35, yshift=-11.5]oof.east) {};

    \pic{ensemble approach={clean}{anchor for train clean}{40}{\headerEnsembleNonPoisoned}};

    \node[] at ([yshift=60pt]container dataset.north) (step 1) {\makecell{\textbf{1) Preparation}}};

    \path (step 1) -|  node[] (step 2) {\makecell{\textbf{2) Poisoning}}} (container poisoning);

    \path (step 2) -| node[xshift=-20] (step 3) {\makecell{\textbf{3) Creation of training sets}}} ([xshift=-20]p2-poisoned);

    \path (step 3) -| node[] (step 4) {\makecell{\textbf{4) Training}}} (rf1-poisoned);

\path[] (container dataset) to node[midway] (midway between dataset and poisoning) {} (container poisoning);

    \draw[-, bar] (midway between dataset and poisoning) -- +(0, 120pt) -- +(0, -150pt);

\coordinate (midway of our approach) at ($(hash-poisoned.west)!.5!(hash-clean.west)$);

    \path[] (container poisoning) to node[midway] (midway between poisoning and training) {} (midway of our approach);

\draw[-, bar] let \p1 = (midway between poisoning and training),
        \p2 = (midway between dataset and poisoning) in
            (\x1, \y2) -- +(0, 120pt); 

    \draw[-, bar] let \p1 = (midway between poisoning and training),
         \p2 = ([yshift=-2]midway between dataset and poisoning) in
             (\x1, \y2) --  +(0, -150pt);

\path (p1-poisoned) to node[midway, xshift=2.5] (midway between partition and rf-1) {} (rf1-poisoned);

\draw[-, bar] let \p1 = (midway between partition and rf-1),
        \p2 = (midway between dataset and poisoning) in
            (\x1, \y2) -- +(0, 120pt);

    \draw[-, bar] let \p1 = (midway between partition and rf-1),
        \p2 = (midway between dataset and poisoning) in
            (\x1, \y2) -- +(0, -150pt);

\coordinate (rf mono training ancor) at ([yshift=-30]rf3-clean.south);
    \pic{rf mono={training}{rf mono training ancor}};

\draw[->, low link, clean link] (training set-1) -- +(35pt, 0) |- (container poisoning);

\draw[->, low link, poisoning link] (container poisoning.east) -- +(7.5pt, 0) |- (hash-poisoned);

\draw[->, low link, poisoning link, to be crossed link] ([xshift=7.5pt]container poisoning.east) |- (rf-mono poisoned-training);

\draw[->, low link, clean link, to be crossed link] (training set text-1) |- node[midway] (midway between training set and rf-mono clean){} (rf-mono clean-training);

\coordinate (midway between hash clean and rf-mono clean) at (midway between training set and rf-mono clean -| hash-clean.west);

\draw[-, low link, clean link]
        let \p1 = ([xshift=-7.5]midway between hash clean and rf-mono clean),
        \p2 = (midway between training set and rf-mono clean) in
            (\x1, \y2) -- +(0, 20pt);

    \draw[->, low link, clean link]
        let \p1 = ([xshift=-7.5]midway between hash clean and rf-mono clean),
        \p2 = ([yshift=24]midway between training set and rf-mono clean) in
            (\x1, \y2) |- (hash-clean);

\coordinate (second test set anchor) at ([xshift=250]test set-1);

    \pic{test set={testing}{second test set anchor}};

\coordinate (poisoned random forest testing anchor) at ([xshift=160]hash-poisoned);
    \coordinate (clean random forest testing anchor) at ([xshift=160]hash-clean);

    \pic{rf with majority voting={testing poisoned}{poisoned random forest testing anchor}{20}{true}};

    \pic{rf with majority voting={testing clean}{clean random forest testing anchor}{20}{true}};

\coordinate (rf mono testing anchor) at (rf-mono poisoned-training -| rf3-testing clean);
    \pic{rf mono={testing}{rf mono testing anchor}};

\coordinate (test set-testing shift) at ([xshift=15pt]test set-testing);
\draw[-, low link, testing link] (test set-testing) -- (test set-testing shift);

\foreach \dest in {poisoned random forest testing anchor, clean random forest testing anchor}{
        \draw[-, low link, testing link] (test set-testing shift) |- node[midway] (midway between test set-testing and \dest) {} (\dest);
    }

\draw[->, low link, testing link] (midway between test set-testing and clean random forest testing anchor.center) |- node[midway] (between test set-testing and and poisoned mono random forest) {} (rf-mono poisoned-testing);

    \draw[->, low link, testing link] (between test set-testing and and poisoned mono random forest.center) |- (rf-mono clean-testing);

\node[] at ([xshift=35]mj-testing poisoned.east) (accuracy ensemble poisoned) {\accuracypoisoned};

    \node[] at ([xshift=35]mj-testing clean.east) (accuracy ensemble clean)  {\accuracyclean};

    \coordinate (accuracy mono poisoned anchor) at (accuracy ensemble poisoned |- rf-mono poisoned-testing);
    \coordinate (accuracy mono clean anchor) at (accuracy ensemble clean |- rf-mono clean-testing);

    \node[] (accuracy mono poisoned) at (accuracy mono poisoned anchor) {\accuracypoisoned};
    \node[] (accuracy mono clean) at (accuracy mono clean anchor) {\accuracyclean};

    \coordinate (accuracy delta anchor) at ([xshift=80]accuracy ensemble clean);
    \node[] (accuracy delta) at (accuracy delta anchor) {\makecell{\deltaf$=$\accuracypoisoned$-$\accuracyclean}};

\coordinate (accuracy to delta meeting point) at ([xshift=30]accuracy ensemble clean);

    \draw[-, low link, accuracy link] (accuracy ensemble clean) to (accuracy to delta meeting point);

    \draw[-, low link, accuracy link] (accuracy ensemble poisoned) -| (accuracy to delta meeting point);
    \draw[-, low link, accuracy link] (accuracy mono poisoned) -| node[midway] (midway between accuracy mono poisoned and accuracy to delta meeting point) {} (accuracy to delta meeting point);
    \draw[-, low link, accuracy link] (accuracy mono clean) -| (midway between accuracy mono poisoned and accuracy to delta meeting point.center);

    \draw[-> , low link, accuracy link] (accuracy to delta meeting point) to (accuracy delta);

\foreach \src/\dest in {mj-testing clean/accuracy ensemble clean, mj-testing poisoned/accuracy ensemble poisoned, rf-mono poisoned-testing/accuracy mono poisoned,  rf-mono clean-testing/accuracy mono clean} {
        \draw[-> , low link, accuracy link] (\src) to (\dest);
    }

\coordinate (midway between test set-testing and accuracy ensemble poisoned) at (test set-testing -| accuracy ensemble poisoned);
    \coordinate (midway midway between test set-testing and accuracy ensemble poisoned) at ($(test set-testing)!.5!(midway between test set-testing and accuracy ensemble poisoned)$);

    \coordinate (title of phase 5 anchor) at (step 4 -| midway midway between test set-testing and accuracy ensemble poisoned);

    \node[] at (title of phase 5 anchor) (step 5) {\makecell{\textbf{5) Testing}}};

\coordinate (title of phase 6 anchor) at (step 5 -| accuracy delta);
    \node[] at (title of phase 6 anchor) (step 6) {\makecell{\textbf{6) Evaluation}}};

\foreach \src/\dest/\yshift in {rf container-clean/rf container-testing clean.north west/-10, rf container-poisoned/rf container-testing poisoned.north west/-10, rf-mono container-training/rf-mono container-testing.west/0}{
        \begin{scope}[on background layer]
            \draw[->, arrow between rf, bend left=45] (\src) to ([yshift=\yshift]\dest);
        \end{scope}
    }

\coordinate (midway between ensemble poisoned and test) at ($(approach-poisoned.east)!.5!(test set-testing)$);

    \coordinate (midway between step 4 and step 5) at ($(step 4.north east)!.5!(step 5.north west)$);

\draw[-, bar]
        let \p1 = (midway between ensemble poisoned and test),
            \p2 = ([yshift=-3]midway between step 4 and step 5) in
                (\x1, \y2) -- +(0, -270pt);

\coordinate (midway between accuracy ensemble poisoned and delta) at ($([xshift=-30]accuracy ensemble poisoned)!.5!(accuracy delta)$);

    \coordinate (midway between step 5 and step 6) at ($(step 5.north east)!.5!(step 6.north west)$);

    \draw[-, bar]
        let \p1 = (midway between accuracy ensemble poisoned and delta),
            \p2 = (midway between step 5 and step 6) in
                (\x1, \y2) -- +(0, -270pt);

\end{tikzpicture}     \end{adjustbox}
    \caption{Evaluation process based on the ensemble approach in Figure~\ref{fig:ensembleapp}.}
    \label{fig:exps}
\end{figure*}

We present the evaluation process and the target datasets at the basis of the experimental results on accuracy degradation in Section~\ref{sec:exp-result}.

\subsection{Evaluation Process in a Nutshell}\label{subsec:exp-process-benchmark-process}
The evaluation process in Figure~\ref{fig:exps} validates the robustness of our ensemble approach in Section~\ref{sec:ensemble} against poisoning attacks in Section~\ref{sec:attack}. For each attack, we calculate the accuracy loss between the plain (monolithic) model and our ensemble approach trained on both original and poisoned datasets. Figure~\ref{fig:exps} shows 4 different paths: \begin{enumerate*}
    \item poisoned hash-based ensemble, considering base models in the ensemble trained on partitions of the poisoned dataset,
    \item non-poisoned hash-based ensemble, considering base models in the ensemble trained on partitions of the original dataset,
    \item monolithic model \emph{RF} trained on the entire poisoned dataset,
    \item monolithic model \emph{RF} trained on the entire original dataset.
\end{enumerate*}

Our process takes as input: \begin{enumerate*}
    \item the original dataset; \item the number of random forests \nbasemodel\ composing the ensemble;
    \item the perturbation type;
    \item the percentage of data points \extentpoint\ and 
    \item features \extentfeature\ to poison.
\end{enumerate*}

The evaluation process consists of two activities, namely, training, and testing and evaluation. Activity training is composed of 4 steps as follows.

\vspace{0.5em}

\noindent \textbf{Step 1: Preparation.} It splits the dataset into training set (denoted as \dataset) and test set, that is, \emph{held out}. Test set is left untouched for the rest of process.

\vspace{.5em}

\noindent \textbf{Step 2: Poisoning.} It applies the selected perturbation to the training set \trainingset\ producing a poisoned training set \datasetpoisoned, according to the percentages of poisoning received as input.

\vspace{.5em}

\noindent \textbf{Step 3: Creation of training sets.} It builds the training sets for the monolithic and ensemble models. It first creates $\nbasemodel$ empty sets (\emph{partitions}).
Second, given a (original or poisoned) training set, each data point is converted to a string by concatenating the value of each feature. For instance, data point \point\ with value $\langle 0, 10, 15, 0, 1\rangle_0$ becomes \texttt{0101501}. Third, the hash of such string is retrieved according to a specific hash algorithm. For instance, the hash value of \texttt{0101501} according to MD5 is \texttt{adf5c364bc3a61133eb2360f7dd0b8f2} (in hexadecimal). Fourth, the modulo operator (modulo \nbasemodel) is applied to the hash value converted back to a number. The result of this operation indicates to which partition the data point belongs, that is, the result $n$ of modulo corresponds to the $n$$+$$1$ partition.
Then, this step produces \nbasemodel\ training sets (subsets of the original training set) by selecting the data points assigned to each partition in a round\-/robin fashion, such that each training set contains roughly the same amount of data points from each partition. This additional assignment distinguishes our ensemble\-/based approach from existing hash\-/based ensemble approaches (e.g.,~\cite{levine2021deep,wang2022improved,pmlr-v162-chen22k}) and guarantees that each training set has approximately the same size and diversity. The $i$-th training set is then used to train random forest $RF_i$ in the ensemble. 
We note that the training set of the monolithic model is the entire input training set.
We also note that this step is repeated both on the original and poisoned datasets.
 
\vspace{.5em}

\noindent \textbf{Step 4: Training.} It trains both the ensemble and monolithic models separately on the original and poisoned training sets created at Step~3. We note that random forests in the ensemble are trained independently one to another.

\vspace{.5em}

Activity testing and evaluation is composed of 2 steps as follows.

\vspace{0.5em}

\noindent \textbf{Step 5: Testing.} It evaluates the accuracy of both the ensemble and monolithic models on the test set isolated at at Step~1. 

\vspace{0.5em}

\noindent \textbf{Step 6: Evaluation.} It compares poisoned and original models using evaluation metric \emph{delta}, denoted as \deltaf, as follows. 

\begin{equation}\label{eq:1}
    \deltaf = \accuracypoisoned - \accuracyclean,
\end{equation}

where \accuracyclean\ is the accuracy retrieved on the original training set and \accuracypoisoned\ is the accuracy retrieved on the poisoned training set. 

We note that \deltaf\ measures the accuracy variation in a model trained on a poisoned training set with regards to the same model trained on the original training set. A negative value of \deltaf\ indicates that the model trained on a poisoned dataset decreases in accuracy, a positive value indicates that the model trained on a poisoned dataset increases in accuracy, a value equals to $0$ indicates the same accuracy.

\subsection{Target Datasets}\label{subsec:exp-process-datasets-settings}

\begin{table*}[htp!]
    \centering
    \caption{Datasets details}
    \label{tab:datasets}
    \renewcommand{\arraystretch}{1.2}
    \begin{tabular}[!t]{lccccccc}
        \hline 
        \multirow{2}{*}{\textbf{Name}} &
        \textbf{N° data points} &
        \multirow{2}{*}{\textbf{N° features}} & 
        \textbf{N° data points} &
        \multirow{2}{*}{\textbf{Sparsity (\%)}} &
        \textbf{Training set size} &
        \textbf{Test set size} \\
& \textbf{(N° per class)} &
        & \textbf{(Preproc.)} &
        & \textbf{(N° per class)} 
        & \textbf{(N° per class)}\\
        \hline

        \multirow{2}{*}{Musk (M2)} & 
        6,598 &                     \multirow{2}{*}{166} &      \multirow{2}{*}{2,034} &    \multirow{2}{*}{0.28} &     1,628 &                     406 \\                      & (1,017/5,581) &           & & &
        (810/818) &                 (207/199)                   

        \\ \hline

        Android & 
        18,733 &                     \multirow{2}{*}{1,000} &     \multirow{2}{*}{14,508} &    \multirow{2}{*}{92.37} &     11,607 &                     2,901 \\                     Malware (AM) 
        & (7,254/11,479) &           & & &
        (5,831/5,776) &              (1,423/1,478)                

        \\ \hline

        \multirow{2}{*}{Spambase (SB)} & 
        4,061 &                     \multirow{2}{*}{57} &       \multirow{2}{*}{3,626} &    \multirow{2}{*}{77.44} &    2,901 &                     725 \\                      & (2,788/1,813) &           & & &
        (1,458/1,443) &              (355/370)                    

        \\ \hline

        Diabetic & 
        \multirow{2}{*}{1,151} &    \multirow{3}{*}{19} &       \multirow{3}{*}{1,080} &    \multirow{3}{*}{10.41} &    \multirow{2}{*}{864} &      \multirow{2}{*}{216} \\     Retinopathy &
        \multirow{2}{*}{(611/540)} &   & & &
        \multirow{2}{*}{(431/443)} &   \multirow{2}{*}{(109/107)} \\  Debrecen (DR)
        & & & & & & \\
        \hline

    \end{tabular}
\end{table*}

We experimentally evaluated the approach in this paper using the datasets in Table~\ref{tab:datasets}, which significantly differ in cardinality (\emph{N° of data points (N° per class)}), number of features (\emph{N° of features}), and sparsity (\emph{Sparsity (\%)}). Table~\ref{tab:datasets} also describes the details of the datasets after a class balancing preprocessing in terms of cardinality (\emph{N° of data points (Preproc.)}), number of data points randomly selected in the training set and number of data points per class (\emph{Training set size (N° per class)}), and the number of data points randomly selected in the test set and the number of data points per class (\emph{Test set size (N° per class)}). We note that sparsity is retrieved from the dataset after preprocessing, and columns describing the cardinality of each class report the cardinality of the positive class first.\footnote{Datasets are publicly available at \url{https://github.com/SESARLab/ensemble-random-forest-robustness-against-poisoning}}

\vspace{.5em}

\noindent 
\textbf{Musk2 (M2)} is an open dataset for the identification of musk molecules~\cite{Dua:2019}, divided in two classes \emph{musk} and \emph{non\-/musk}. The dataset is collected by including different conformations (shapes) of musk and non\-/musk molecules. In particular, all the low\-/energy conformations of $141$ initial molecules have been generated and manually annotated.

The dataset consists of 6,598 data points (1,017 musk and 5,581 non musk), organized in 166 features. We built a balanced dataset of 2,034 points by randomly subsampling data points in class non musk. After preprocessing, the dataset exhibits a low sparsity $\approx$$0.28\%$. We finally split the dataset in a training set of 1,628 data points (810 musks and 818 non musks) and in a test set of 406 data points (207 musks and 199 non musks).

\vspace{.5em}

\noindent \textbf{Android malware (AM)} is a proprietary dataset for the detection of malware on Android devices. 
The dataset is collected on Android devices with benign and malign apps installed, by capturing the system calls performed by the apps. Any sequence of three consecutive system calls is a feature, whose value is the number of times such sequence has been called.

The dataset consists of 18,733 data points (7,254 malware and 11,479 non malware), organized in 25,802 features. We then built a balanced dataset of 14,508 points, by randomly subsampling data points in class non malware. We further split the dataset in a training set of 11,607 data points (5,831 malware and 5,776 non malware) and in a test set of 2,901 data points (1,423 malware and 1,478 non malware). Finally, being a dataset with high dimensionality and more features than data points, to avoid overfitting, we reduced the number of features according to \emph{InfoGain}~\cite{Quinlan1986}, a feature ranking method selecting those features that reduce the \emph{entropy} in the dataset (i.e., the most informative features with regards to the dataset). With this method, we reduced the number of features to 1,000. After preprocessing, the dataset exhibits a high sparsity ($\approx$$92.37\%$).

\vspace{.5em}

\noindent \textbf{Spambase (SB)} is an open and well-know dataset for spam detection in email body messages~\cite{DBLP:journals/ml/HushSS07,DBLP:conf/icml/WangW02,DBLP:conf/esann/ChristosS04}.
It contains features counting the occurrence of particular words and the length of sequences of consecutive capital letters.

The dataset consists of 4,061 data points (2,788 spam and 1,813 non spam), organized in 57 features. We built a balanced dataset of 3,626 points, by randomly subsampling data points in class spam. After preprocessing, the dataset exhibits a medium-high sparsity but lower than AM ($\approx$$77.44\%$). We finally split the dataset in a training set of 2,901 data points (1,458 spam and 1,433 non spam) and in a test set of 725 instances (355 spam and 370 non spam).

\vspace{.5em}

\noindent \textbf{Diabetic Retinopaty Debrecen (DR)} is an open dataset for the detection of symptoms of diabetic retinopathy~\cite{DBLP:journals/kbs/AntalH14}. It contains features extracted from the Messidor image set~\cite{decenciere2014feedback}. All features are numeric and represent either a detected lesion, a descriptive feature of a anatomical part, or an image-level descriptor. The label indicates if an image contains signs of diabetic retinopathy or not.

The dataset consists of 1,151 data points (611 signs of disease and 540 no signs of disease), organized in 19 features. We built a balanced dataset of 1,080 points, by randomly subsampling data points in class signs of disease. After preprocessing, the dataset exhibits a low sparsity but higher than M2 ($\approx$$10.41\%$). We finally split the dataset in a training set of 864 data points (431 signs of disease and 433 no signs of disease) and in a test set of 216 data points (109 signs of disease and 107 no signs of disease). 
 \section{Experimental Results}\label{sec:exp-result}
We present the accuracy degradation retrieved by our ensemble approach against \labelflipping\ (Section \ref{subsubsec:exp-result-flip}), and \zeroing, \noising, and \outofranging\ (Section~\ref{subsubsec:exp-result-other}) for datasets M2, AM, SB, and DR in Section \ref{subsec:exp-process-datasets-settings}. 
Label flipping is in fact the most effective perturbation substantially affecting the behavior of monolithic model, while \zeroing, \noising, \outofranging\ do not produce substantial accuracy degradation on it.
For readability, we omit the percentage symbol (\%) when presenting values of \deltaf, accuracy, as well as percentage of data points \extentpoint\ and features \extentfeature\ to be poisoned. 

\subsection{Experimental Settings}
Our experiments have been built on the ML library \emph{Weka}~\cite{DBLP:journals/sigkdd/HallFHPRW09} version 3.8 running on Java version 8.
We executed our process in Section~\ref{subsec:exp-process-benchmark-process} on a VM equipped with 16 vCPUs Intel Core Processor (Broadwell, no TSX) 2.00 GHz and 48 GBs of RAM.
The entire process has been executed 5 times averaging accuracy and \deltaf. 

The settings of our experiments varied
\begin{enumerate*}
    \item the number of random forests \nbasemodel\ in the ensemble in $\{3,$ $5,$ $7,$ $9,$ $11,$ $13,$ $15,$ $17,$ $19,$ $21\}$;
    \item the perturbations in \zeroing, \noising, \outofranging\ and \labelflipping\ (Section~\ref{sec:attack});
    \item the percentage of poisoned data points \extentpoint\ in $[10, 35]$, step $5$;
    \item the percentage of poisoned features \extentfeature on each data point in $[10, 35]$, step $5$. We note that \extentfeature\ is not applicable to perturbation \labelflipping. 
\end{enumerate*}
Each combination of these parameters represented an instance of the evaluation process in Section~\ref{subsec:exp-process-benchmark-process}.
In addition, we put ourselves in a worst-case scenario using the outdated hash function MD5 to assign data points to partitions, to determine whether our ensemble approach can still provide some protection against poisoning attacks.
Finally, we configured random forests according to the well\-/known practices in the state of the art.\footnote{\url{https://weka.sourceforge.io/doc.dev/weka/classifiers/trees/RandomForest.html}}

\subsection{Label Flipping}\label{subsubsec:exp-result-flip}

\begin{table*}[htbp!]
    \centering
    \caption{ Results of \labelflipping\ varying number of random forests \nbasemodel\ and percentage of poisoned data points \extentpoint.}\vspace{-0.5em}
    \label{tab:deltaFlip}
    \begin{subtable}[c]{1\textwidth}
        \centering
        \begin{tabular}{rlcccccccccccc}
\hline\\[-1.5ex]
& & \multicolumn{12}{c}{Number of random forests \nbasemodel\ of the ensemble }\\
& & 1 & 3 & 5 & 7 & 9 & 11 & 13 & 15 & 17 & 19 & 21 &\\
\cline{3-13}\\
[-1.5ex]\multirow{13}{*}{\rotatebox[origin=c]{90}{\textbf{Poison. data points \extentpoint\ (\%)}}} 
& \multicolumn{1}{r}{\cellcolor{lightgray}0} &
\cellcolor{lightgray}\makecell[r]{0.000 \\ 91.872} &
\cellcolor{lightgray}\makecell[r]{0.000 \\ 91.133} &
\cellcolor{lightgray}\makecell[r]{0.000 \\ 90.805} &
\cellcolor{lightgray}\makecell[r]{0.000 \\ 89.901} &
\cellcolor{lightgray}\makecell[r]{0.000 \\ 90.230} &
\cellcolor{lightgray}\makecell[r]{0.000 \\ 90.066} &
\cellcolor{lightgray}\makecell[r]{0.000 \\ 89.984} &
\cellcolor{lightgray}\makecell[r]{0.000 \\ 89.737} &
\cellcolor{lightgray}\makecell[r]{0.000 \\ 88.752} &
\cellcolor{lightgray}\makecell[r]{0.000 \\ 88.423} &
\cellcolor{lightgray}\makecell[r]{0.000 \\ 88.177} &
\vspace{.2em}\\
&10 &
\makecell[r]{-3.449\\88.423} &
\makecell[r]{-1.560\\89.573} &
\makecell[r]{-0.083 \\ 90.722} &
\makecell[r]{-0.328 \\ 89.573} &
\makecell[r]{0.493 \\ 90.723} &
\makecell[r]{0.000 \\ 90.066} &
\makecell[r]{-0.493 \\ 89.491} &
\makecell[r]{-0.410 \\ 89.327} &
\makecell[r]{0.246 \\ 88.998} &
\makecell[r]{-0.246 \\ 88.177} &
\makecell[r]{0.821 \\ 88.998} &
\vspace{.2em}\\
& 15 &
\makecell[r]{-6.650 \\ 85.222} &
\makecell[r]{-2.874 \\ 88.259} &
\makecell[r]{-1.314 \\ 89.491} &
\makecell[r]{-1.149 \\ 88.752} &
\makecell[r]{-0.247 \\ 89.983} &
\makecell[r]{-1.232 \\ 88.834} &
\makecell[r]{-1.560 \\ 88.424} &
\makecell[r]{0.327 \\ 90.066} &
\makecell[r]{-3.202 \\ 85.550} &
\makecell[r]{-0.575 \\ 87.849} &
\makecell[r]{0.411 \\ 88.588} &
\vspace{.2em}\\
& 20 &
\makecell[r]{-7.553 \\ 84.319} &
\makecell[r]{-1.970 \\ 89.163} &
\makecell[r]{-0.657 \\ 90.148} &
\makecell[r]{-1.149 \\ 88.752} &
\makecell[r]{-0.985 \\ 89.245} &
\makecell[r]{-1.478 \\ 88.588} &
\makecell[r]{-1.971 \\ 88.013} &
\makecell[r]{-2.052 \\ 87.685} &
\makecell[r]{-1.806 \\ 86.946} &
\makecell[r]{-0.657 \\ 87.767} &
\makecell[r]{-0.657 \\ 87.520} &
\vspace{.2em}\\
& 25 &
\makecell[r]{-11.330 \\ 80.542} &
\makecell[r]{-4.187 \\ 86.946} &
\makecell[r]{-4.269 \\ 86.535} &
\makecell[r]{-3.448 \\ 86.453} &
\makecell[r]{-2.463 \\ 87.767} &
\makecell[r]{-2.135 \\ 87.931} &
\makecell[r]{-3.449 \\ 86.535} &
\makecell[r]{-3.612 \\ 86.125} &
\makecell[r]{-2.052 \\ 86.700} &
\makecell[r]{-0.903 \\ 87.521} &
\makecell[r]{-2.709 \\ 85.468} &
\vspace{.2em}\\
& 30 &
\makecell[r]{-16.092 \\ 75.780} &
\makecell[r]{-10.920 \\ 80.213} &
\makecell[r]{-6.158 \\ 84.647} &
\makecell[r]{-4.597 \\ 85.304} &
\makecell[r]{-5.665 \\ 84.565} &
\makecell[r]{-3.941 \\ 86.125} &
\makecell[r]{-5.173 \\ 84.811} &
\makecell[r]{-5.254 \\ 84.483} &
\makecell[r]{-3.695 \\ 85.057} &
\makecell[r]{-5.255 \\ 83.169} &
\makecell[r]{-3.694 \\ 84.483} &
\vspace{.2em}\\
& 35 &
\makecell[r]{-22.414 \\ 69.458} &
\makecell[r]{-15.435 \\ 75.698} &
\makecell[r]{-12.562 \\ 78.243} &
\makecell[r]{-8.949 \\ 80.952} &
\makecell[r]{-9.688 \\ 80.542} &
\makecell[r]{-11.330 \\ 78.736} &
\makecell[r]{-9.442 \\ 80.542} &
\makecell[r]{-9.113 \\ 80.624} &
\makecell[r]{-7.636 \\ 81.116} &
\makecell[r]{-10.181 \\ 78.243} &
\makecell[r]{-6.814 \\ 81.363} &
\\\hline
\end{tabular}  \\
        \subcaption{Dataset M2}
\label{subtab:musk}
    \end{subtable}
\quad \begin{subtable}[c]{1\textwidth}
        \centering
        \begin{tabular}{rlcccccccccccc}
\hline\\[-1.5ex]
& & \multicolumn{12}{c}{Number of random forests \nbasemodel\ of the ensemble }\\
& & 1 & 3 & 5 & 7 & 9 & 11 & 13 & 15 & 17 & 19 & 21 &\\
\cline{3-13}\\
[-1.5ex]\multirow{13}{*}{\rotatebox[origin=c]{90}{\textbf{Poison. data points \extentpoint\ (\%)}}} 
& \multicolumn{1}{r}{\cellcolor{lightgray}0} &
\cellcolor{lightgray}\makecell[r]{0.000 \\ 98.828} &
\cellcolor{lightgray}\makecell[r]{0.000 \\ 98.541} &
\cellcolor{lightgray}\makecell[r]{0.000 \\ 98.449} &
\cellcolor{lightgray}\makecell[r]{0.000 \\ 98.047} &
\cellcolor{lightgray}\makecell[r]{0.000 \\ 98.070} &
\cellcolor{lightgray}\makecell[r]{0.000 \\ 97.886} &
\cellcolor{lightgray}\makecell[r]{0.000 \\ 97.828} &
\cellcolor{lightgray}\makecell[r]{0.000 \\ 97.679} &
\cellcolor{lightgray}\makecell[r]{0.000 \\ 97.483} &
\cellcolor{lightgray}\makecell[r]{0.000 \\ 97.311} &
\cellcolor{lightgray}\makecell[r]{0.000 \\ 97.104} &
\vspace{.2em}\\
& 10 &
\makecell[r]{-2.344 \\ 96.484} &
\makecell[r]{-0.402 \\ 98.139} &
\makecell[r]{-0.184 \\ 98.265} &
\makecell[r]{-0.023 \\ 98.024} &
\makecell[r]{0.023 \\ 98.093} &
\makecell[r]{-0.046 \\ 97.840} &
\makecell[r]{-0.068 \\ 97.760} &
\makecell[r]{0.000 \\ 97.679} &
\makecell[r]{-0.034 \\ 97.449} &
\makecell[r]{-0.069 \\ 97.242} &
\makecell[r]{0.081 \\ 97.185} &
\vspace{.2em}\\
& 15 &
\makecell[r]{-4.447 \\ 94.381} &
\makecell[r]{-0.943 \\ 97.598} &
\makecell[r]{-0.333 \\ 98.116} &
\makecell[r]{-0.150 \\ 97.897} &
\makecell[r]{-0.161 \\ 97.909} &
\makecell[r]{-0.069 \\ 97.817} &
\makecell[r]{-0.321 \\ 97.507} &
\makecell[r]{-0.230 \\ 97.449} &
\makecell[r]{-0.367 \\ 97.116} &
\makecell[r]{0.046 \\ 97.357} &
\makecell[r]{0.092 \\ 97.196} &
\vspace{.2em}\\
& 20 &
\makecell[r]{-6.641 \\ 92.187} &
\makecell[r]{-1.862 \\ 96.679} &
\makecell[r]{-1.092 \\ 97.357} &
\makecell[r]{-0.356 \\ 97.691} &
\makecell[r]{-0.219 \\ 97.851} &
\makecell[r]{-0.265 \\ 97.621} &
\makecell[r]{-0.310 \\ 97.518} &
\makecell[r]{-0.138 \\ 97.541} &
\makecell[r]{-0.137 \\ 97.346} &
\makecell[r]{-0.241 \\ 97.070} &
\makecell[r]{0.161 \\ 97.265} &
\vspace{.2em}\\
& 25 &
\makecell[r]{-11.100 \\ 87.728} &
\makecell[r]{-4.332 \\ 94.209} &
\makecell[r]{-1.999 \\ 96.450} &
\makecell[r]{-1.494 \\ 96.553} &
\makecell[r]{-1.230 \\ 96.840} &
\makecell[r]{-0.644 \\ 97.242} &
\makecell[r]{-0.505 \\ 97.323} &
\makecell[r]{-0.138 \\ 97.541} &
\makecell[r]{-0.471 \\ 97.012} &
\makecell[r]{-0.195 \\ 97.116} &
\makecell[r]{-0.252 \\ 96.852} &
\vspace{.2em}\\
& 30 &
\makecell[r]{-16.052 \\ 82.776} &
\makecell[r]{-7.136 \\ 91.405} &
\makecell[r]{-4.148 \\ 94.301} &
\makecell[r]{-2.620 \\ 95.427} &
\makecell[r]{-1.873 \\ 96.197} &
\makecell[r]{-1.218 \\ 96.668} &
\makecell[r]{-1.597 \\ 96.231} &
\makecell[r]{-0.655 \\ 97.024} &
\makecell[r]{-0.723 \\ 96.760} &
\makecell[r]{-0.655 \\ 96.656} &
\makecell[r]{-0.574 \\ 96.530} &
\vspace{.2em}\\
& 35 &
\makecell[r]{-22.578 \\ 76.250} &
\makecell[r]{-14.536 \\ 84.005} &
\makecell[r]{-8.595 \\ 89.854} &
\makecell[r]{-6.067 \\ 91.980} &
\makecell[r]{-4.355 \\ 93.715} &
\makecell[r]{-3.562 \\ 94.324} &
\makecell[r]{-3.630 \\ 94.198} &
\makecell[r]{-2.298 \\ 95.381} &
\makecell[r]{-1.964 \\ 95.519} &
\makecell[r]{-1.884 \\ 95.427} &
\makecell[r]{-0.942 \\ 96.162} &
\\\hline
\end{tabular}  \\
        \subcaption{Dataset AM}
\label{subtab:AM}
    \end{subtable}
\quad \begin{subtable}[c]{1\textwidth}
        \centering
        \begin{tabular}{rlcccccccccccc}
\hline\\[-1.5ex]
& & \multicolumn{12}{c}{Number of random forests \nbasemodel\ of the ensemble }\\
& & 1 & 3 & 5 & 7 & 9 & 11 & 13 & 15 & 17 & 19 & 21 &\\
\cline{3-13}\\
[-1.5ex]\multirow{13}{*}{\rotatebox[origin=c]{90}{\textbf{Poison. data points \extentpoint\ (\%)}}} 
& \multicolumn{1}{r}{\cellcolor{lightgray}0} &
\cellcolor{lightgray}\makecell[r]{0.000 \\ 94.897} &
\cellcolor{lightgray}\makecell[r]{0.000 \\ 93.425} &
\cellcolor{lightgray}\makecell[r]{0.000 \\ 92.184} &
\cellcolor{lightgray}\makecell[r]{0.000 \\ 91.632} &
\cellcolor{lightgray}\makecell[r]{0.000 \\ 91.402} &
\cellcolor{lightgray}\makecell[r]{0.000 \\ 91.264} &
\cellcolor{lightgray}\makecell[r]{0.000 \\ 90.942} &
\cellcolor{lightgray}\makecell[r]{0.000 \\ 90.575} &
\cellcolor{lightgray}\makecell[r]{0.000 \\ 90.391} &
\cellcolor{lightgray}\makecell[r]{0.000 \\ 90.483} &
\cellcolor{lightgray}\makecell[r]{0.000 \\ 90.115} &
\vspace{.2em}\\
&10 &
\makecell[r]{-3.173 \\ 91.724} &
\makecell[r]{-0.322 \\ 93.103} &
\makecell[r]{0.873 \\ 93.057} &
\makecell[r]{0.138 \\ 91.770} &
\makecell[r]{0.138 \\ 91.540} &
\makecell[r]{0.138 \\ 91.402} &
\makecell[r]{0.644 \\ 91.586} &
\makecell[r]{0.184 \\ 90.759} &
\makecell[r]{0.046 \\ 90.437} &
\makecell[r]{0.276 \\ 90.759} &
\makecell[r]{0.414 \\ 90.529} &
\vspace{.2em}\\
& 15 &
\makecell[r]{-4.828 \\ 90.069} &
\makecell[r]{-0.506 \\ 92.919} &
\makecell[r]{0.644 \\ 92.828} &
\makecell[r]{0.736 \\ 92.368} &
\makecell[r]{0.460 \\ 91.862} &
\makecell[r]{0.046 \\ 91.310} &
\makecell[r]{0.598 \\ 91.540} &
\makecell[r]{0.873 \\ 91.448} &
\makecell[r]{0.781 \\ 91.172} &
\makecell[r]{0.322 \\ 90.805} &
\makecell[r]{0.552 \\ 90.667} &
\vspace{.2em}\\
& 20 &
\makecell[r]{-6.069 \\ 88.828} &
\makecell[r]{-1.655 \\ 91.770} &
\makecell[r]{0.184 \\ 92.368} &
\makecell[r]{0.552 \\ 92.184} &
\makecell[r]{0.506 \\ 91.908} &
\makecell[r]{0.782 \\ 92.046} &
\makecell[r]{0.414 \\ 91.356} &
\makecell[r]{1.103 \\ 91.678} &
\makecell[r]{0.414 \\ 90.805} &
\makecell[r]{1.011 \\ 91.494} &
\makecell[r]{0.828 \\ 90.943} &
\vspace{.2em}\\
& 25 &
\makecell[r]{-8.138 \\ 86.759} &
\makecell[r]{-3.448 \\ 89.977} &
\makecell[r]{-0.828 \\ 91.356} &
\makecell[r]{-0.643 \\ 90.989} &
\makecell[r]{-0.184 \\ 91.218} &
\makecell[r]{-0.092 \\ 91.172} &
\makecell[r]{0.368 \\ 91.310} &
\makecell[r]{0.413 \\ 90.988} &
\makecell[r]{0.827 \\ 91.218} &
\makecell[r]{-0.506 \\ 89.977} &
\makecell[r]{1.195 \\ 91.310} &
\vspace{.2em}\\
& 30 &
\makecell[r]{-10.759 \\ 84.138} &
\makecell[r]{-5.839 \\ 87.586} &
\makecell[r]{-1.931 \\ 90.253} &
\makecell[r]{-1.241 \\ 90.391} &
\makecell[r]{-0.597 \\ 90.805} &
\makecell[r]{-0.414 \\ 90.850} &
\makecell[r]{0.414 \\ 91.356} &
\makecell[r]{-0.138 \\ 90.437} &
\makecell[r]{-0.184 \\ 90.207} &
\makecell[r]{0.414 \\ 90.897} &
\makecell[r]{-0.046 \\ 90.069} &
\vspace{.2em}\\
& 35 &
\makecell[r]{-17.380 \\ 77.517} &
\makecell[r]{-9.609 \\ 83.816} &
\makecell[r]{-5.839 \\ 86.345} &
\makecell[r]{-3.494 \\ 88.138} &
\makecell[r]{-2.115 \\ 89.287} &
\makecell[r]{-2.115 \\ 89.151} &
\makecell[r]{-1.379 \\ 89.563} &
\makecell[r]{-0.414 \\ 90.161} &
\makecell[r]{0.184 \\ 90.575} &
\makecell[r]{-0.552 \\ 89.931} &
\makecell[r]{0.046 \\ 90.161} &
\\\hline
\end{tabular}  \\
        \subcaption{Dataset SB}
        \label{subtab:spam}
\end{subtable}
\quad \begin{subtable}[c]{1\textwidth}
        \centering
        \begin{tabular}{rlcccccccccccc}
\hline\\[-1.5ex]
& & \multicolumn{12}{c}{Number of random forests \nbasemodel\ of the ensemble }\\
& & 1 & 3 & 5 & 7 & 9 & 11 & 13 & 15 & 17 & 19 & 21 &\\
\cline{3-13}\\
[-1.5ex]\multirow{13}{*}{\rotatebox[origin=c]{90}{\textbf{Poison. data points \extentpoint\ (\%)}}} 
& \multicolumn{1}{r}{\cellcolor{lightgray}0} &
\cellcolor{lightgray}\makecell[r]{0.000 \\ 69.908} &
\cellcolor{lightgray}\makecell[r]{0.000 \\ 68.827} &
\cellcolor{lightgray}\makecell[r]{0.000 \\ 67.901} &
\cellcolor{lightgray}\makecell[r]{0.000 \\ 68.519} &
\cellcolor{lightgray}\makecell[r]{0.000 \\ 66.821} &
\cellcolor{lightgray}\makecell[r]{0.000 \\ 66.050} &
\cellcolor{lightgray}\makecell[r]{0.000 \\ 65.587} &
\cellcolor{lightgray}\makecell[r]{0.000 \\ 68.210} &
\cellcolor{lightgray}\makecell[r]{0.000 \\ 65.741} &
\cellcolor{lightgray}\makecell[r]{0.000 \\ 67.439} &
\cellcolor{lightgray}\makecell[r]{0.000 \\ 67.593} &
\vspace{.2em}\\
&10 &
\makecell[r]{-1.852\\68.056} &
\makecell[r]{-0.309\\68.518} &
\makecell[r]{1.389 \\ 69.290} &
\makecell[r]{-0.926 \\ 67.593} &
\makecell[r]{-1.389\\ 65.432} &
\makecell[r]{0.000\\ 66.050} &
\makecell[r]{0.463 \\ 66.050} &
\makecell[r]{-1.852 \\ 66.358} &
\makecell[r]{2.161\\ 67.902} &
\makecell[r]{-0.926 \\ 66.513} &
\makecell[r]{-1.389\\ 66.204} &
\vspace{.2em}\\
& 15 &
\makecell[r]{-3.241 \\ 66.667} &
\makecell[r]{-1.697 \\ 67.130} &
\makecell[r]{-2.469 \\ 65.432} &
\makecell[r]{-3.241 \\ 65.278} &
\makecell[r]{0.154 \\ 66.975} &
\makecell[r]{-0.309 \\ 65.741} &
\makecell[r]{0.154 \\ 65.741} &
\makecell[r]{-3.704 \\ 64.506} &
\makecell[r]{-0.463 \\ 65.278} &
\makecell[r]{-0.309 \\ 67.130} &
\makecell[r]{-2.161 \\ 65.432} &
\vspace{.2em}\\
& 20 &
\makecell[r]{-6.019 \\ 63.889} &
\makecell[r]{-6.481 \\ 62.346} &
\makecell[r]{-4.167 \\ 63.736} &
\makecell[r]{-3.395 \\ 65.124} &
\makecell[r]{1.543 \\ 68.364} &
\makecell[r]{0.000 \\ 66.050} &
\makecell[r]{-1.544 \\ 64.043} &
\makecell[r]{-2.315 \\ 65.895} &
\makecell[r]{1.389 \\ 67.130} &
\makecell[r]{-4.013 \\ 63.426} &
\makecell[r]{-2.315 \\ 65.278} &
\vspace{.2em}\\
& 25 &
\makecell[r]{-9.259 \\ 60.648} &
\makecell[r]{-5.864 \\ 62.963} &
\makecell[r]{-2.773 \\ 65.124} &
\makecell[r]{-5.402 \\ 63.117} &
\makecell[r]{-2.469 \\ 64.352} &
\makecell[r]{-1.544 \\ 64.506} &
\makecell[r]{-2.933 \\ 62.654} &
\makecell[r]{-3.241 \\ 64.969} &
\makecell[r]{0.309 \\ 66.050} &
\makecell[r]{-1.389 \\ 66.050} &
\makecell[r]{-3.550 \\ 64.043} &
\vspace{.2em}\\
& 30 &
\makecell[r]{-15.741 \\ 54.167} &
\makecell[r]{-7.870 \\ 60.957} &
\makecell[r]{-4.012 \\ 63.889} &
\makecell[r]{-7.562 \\ 60.957} &
\makecell[r]{-3.549 \\ 63.272} &
\makecell[r]{-2.624 \\ 63.426} &
\makecell[r]{-2.624 \\ 62.963} &
\makecell[r]{-3.858 \\ 64.352} &
\makecell[r]{-2.932 \\ 62.809} &
\makecell[r]{-4.630 \\ 62.809} &
\makecell[r]{-2.624 \\ 64.969} &
\vspace{.2em}\\
& 35 &
\makecell[r]{-17.593 \\ 52.315} &
\makecell[r]{-10.185 \\ 58.642} &
\makecell[r]{-9.414 \\ 58.487} &
\makecell[r]{-9.723 \\ 58.796} &
\makecell[r]{-6.482 \\ 60.339} &
\makecell[r]{-5.093 \\ 60.957} &
\makecell[r]{-6.019 \\ 59.568} &
\makecell[r]{-6.173 \\ 62.037} &
\makecell[r]{-6.173 \\ 59.568} &
\makecell[r]{-6.328 \\ 61.111} &
\makecell[r]{-9.260 \\ 58.333} &
\\\hline
\end{tabular}  \\
        \subcaption{Dataset DR}
        \label{subtab:DRD}
\end{subtable}

\end{table*}

Tables~\ref{subtab:musk}--\ref{subtab:DRD} show the results retrieved by executing perturbation \labelflipping\ against datasets M2 (Table~\ref{subtab:musk}), AM (Table~\ref{subtab:AM}),  SB (Table~\ref{subtab:spam}), and DR (Table~\ref{subtab:DRD}), varying the percentage of poisoned data points \extentpoint\ and the number \nbasemodel\ of random forests in the ensemble. The column with $\nbasemodel$$=$$1$ indicates the monolithic model, while the row with \extentpoint$=$$0$ (rows with gray background in Tables~\ref{subtab:musk}--\ref{subtab:DRD}) indicates the accuracy \accuracyclean\ retrieved from the model trained on the original dataset, that is, the dataset with no poisoned data points.
Each cell is divided in two parts. The top\-/most part reports the \deltaf\ in Equation~(\ref{eq:1}), retrieved according to the accuracy of the model trained on the poisoned training set and the one on the original training set. The bottom\-/most part reports the accuracy \accuracypoisoned\ retrieved by the model trained on the poisoned training set.

Our results first show that the accuracy retrieved using the 4 datasets varies significantly. In particular, the monolithic model shows \accuracyclean\ of $91.872$ for M2, $98.828$ for AM, $94.897$ for SB, and $69.908$ for DR. These variations in accuracy depends on the diversity of the selected dataset and can be observed in all configurations, reaching the peak with \extentpoint$=$$35$.

Our results additionally show two clear trends.
First, as the percentage of poisoned data points increases, the corresponding \deltaf\ decreases, that is, the more label flips, the higher the accuracy decrease. This trend can be observed downward column by column.
For instance, considering the smallest ensemble $\nbasemodel$$=$$3$. 
$\deltaf$ decreases from $-1.560$ with $\extentpoint$$=$$10$ to $-15.435$ with $\extentpoint$$=$$35$ for M2, from $-0.402$ to $-14.536$ for AM, from $-0.322$ to $-9.609$ for SP, and from $-0.309$ to $-10.185$ for DR.
Second, as the number of random forests in the ensemble increases, the corresponding \deltaf\ increases, that is, the larger the ensemble, the lesser the accuracy decrease. This trend can be observed rightward row by row. For instance, considering the worst perturbation ($\extentpoint$$=$$35$), we can see that $\deltaf$ improves of $\approx$$70\%$ for M2 (increasing from $-22.414$ with $\nbasemodel$$=$$1$ to $-6.814$ with $\nbasemodel$$=$$21$), 
$\approx$$96\%$ for AM (increasing from $-22.578$ to $-0.942$), 
$100\%$ for SB (increasing from $-17.380$ to $0.046$, and $\approx$$53\%$ on DR (increasing from $-17.593$ to $-9.260$).
It is important to note that these two trends are less pronounced in dataset DR, due to the limited amount of data points in the training set and to the low classification performance of the random forest.

\begin{figure}[t]
  \begin{adjustbox}{max totalsize={.49\textwidth}{\textheight},center}
  \centering
 \begingroup
  \makeatletter
  \providecommand\color[2][]{\GenericError{(gnuplot) \space\space\space\@spaces}{Package color not loaded in conjunction with
      terminal option `colourtext'}{See the gnuplot documentation for explanation.}{Either use 'blacktext' in gnuplot or load the package
      color.sty in LaTeX.}\renewcommand\color[2][]{}}\providecommand\includegraphics[2][]{\GenericError{(gnuplot) \space\space\space\@spaces}{Package graphicx or graphics not loaded}{See the gnuplot documentation for explanation.}{The gnuplot epslatex terminal needs graphicx.sty or graphics.sty.}\renewcommand\includegraphics[2][]{}}\providecommand\rotatebox[2]{#2}\@ifundefined{ifGPcolor}{\newif\ifGPcolor
    \GPcolorfalse
  }{}\@ifundefined{ifGPblacktext}{\newif\ifGPblacktext
    \GPblacktexttrue
  }{}\let\gplgaddtomacro\g@addto@macro
\gdef\gplbacktext{}\gdef\gplfronttext{}\makeatother
  \ifGPblacktext
\def\colorrgb#1{}\def\colorgray#1{}\else
\ifGPcolor
      \def\colorrgb#1{\color[rgb]{#1}}\def\colorgray#1{\color[gray]{#1}}\expandafter\def\csname LTw\endcsname{\color{white}}\expandafter\def\csname LTb\endcsname{\color{black}}\expandafter\def\csname LTa\endcsname{\color{black}}\expandafter\def\csname LT0\endcsname{\color[rgb]{1,0,0}}\expandafter\def\csname LT1\endcsname{\color[rgb]{0,1,0}}\expandafter\def\csname LT2\endcsname{\color[rgb]{0,0,1}}\expandafter\def\csname LT3\endcsname{\color[rgb]{1,0,1}}\expandafter\def\csname LT4\endcsname{\color[rgb]{0,1,1}}\expandafter\def\csname LT5\endcsname{\color[rgb]{1,1,0}}\expandafter\def\csname LT6\endcsname{\color[rgb]{0,0,0}}\expandafter\def\csname LT7\endcsname{\color[rgb]{1,0.3,0}}\expandafter\def\csname LT8\endcsname{\color[rgb]{0.5,0.5,0.5}}\else
\def\colorrgb#1{\color{black}}\def\colorgray#1{\color[gray]{#1}}\expandafter\def\csname LTw\endcsname{\color{white}}\expandafter\def\csname LTb\endcsname{\color{black}}\expandafter\def\csname LTa\endcsname{\color{black}}\expandafter\def\csname LT0\endcsname{\color{black}}\expandafter\def\csname LT1\endcsname{\color{black}}\expandafter\def\csname LT2\endcsname{\color{black}}\expandafter\def\csname LT3\endcsname{\color{black}}\expandafter\def\csname LT4\endcsname{\color{black}}\expandafter\def\csname LT5\endcsname{\color{black}}\expandafter\def\csname LT6\endcsname{\color{black}}\expandafter\def\csname LT7\endcsname{\color{black}}\expandafter\def\csname LT8\endcsname{\color{black}}\fi
  \fi
    \setlength{\unitlength}{0.0500bp}\ifx\gptboxheight\undefined \newlength{\gptboxheight}\newlength{\gptboxwidth}\newsavebox{\gptboxtext}\fi \setlength{\fboxrule}{0.5pt}\setlength{\fboxsep}{1pt}\definecolor{tbcol}{rgb}{1,1,1}\begin{picture}(7200.00,5040.00)\gplgaddtomacro\gplbacktext{\csname LTb\endcsname \put(814,4159){\makebox(0,0)[r]{\strut{}$-25$}}\put(814,3583){\makebox(0,0)[r]{\strut{}$-20$}}\put(814,3007){\makebox(0,0)[r]{\strut{}$-15$}}\put(814,2432){\makebox(0,0)[r]{\strut{}$-10$}}\put(814,1856){\makebox(0,0)[r]{\strut{}$-5$}}\put(814,1280){\makebox(0,0)[r]{\strut{}$0$}}\put(814,704){\makebox(0,0)[r]{\strut{}$5$}}\put(946,484){\makebox(0,0){\strut{}$10$}}\put(2117,484){\makebox(0,0){\strut{}$15$}}\put(3289,484){\makebox(0,0){\strut{}$20$}}\put(4460,484){\makebox(0,0){\strut{}$25$}}\put(5632,484){\makebox(0,0){\strut{}$30$}}\put(6803,484){\makebox(0,0){\strut{}$35$}}}\gplgaddtomacro\gplfronttext{\csname LTb\endcsname \put(209,2431){\rotatebox{-270}{\makebox(0,0){\strut{}\Large \deltaf}}}\put(3874,154){\makebox(0,0){\strut{}\Large \extentpoint (\%)}}\put(1636,4867){\makebox(0,0)[r]{\strut{}M2 N=1}}\put(1636,4647){\makebox(0,0)[r]{\strut{}M2 N=3}}\put(1636,4427){\makebox(0,0)[r]{\strut{}M2 N=21}}\put(3085,4867){\makebox(0,0)[r]{\strut{}AM N=1}}\put(3085,4647){\makebox(0,0)[r]{\strut{}AM N=3}}\put(3085,4427){\makebox(0,0)[r]{\strut{}AM N=21}}\put(4534,4867){\makebox(0,0)[r]{\strut{}SB N=1}}\put(4534,4647){\makebox(0,0)[r]{\strut{}SB N=3}}\put(4534,4427){\makebox(0,0)[r]{\strut{}SB N=21}}\put(5983,4867){\makebox(0,0)[r]{\strut{}DR N=1}}\put(5983,4647){\makebox(0,0)[r]{\strut{}DR N=3}}\put(5983,4427){\makebox(0,0)[r]{\strut{}DR N=21}}}\gplbacktext
    \put(0,0){\includegraphics[width={360.00bp},height={252.00bp}]{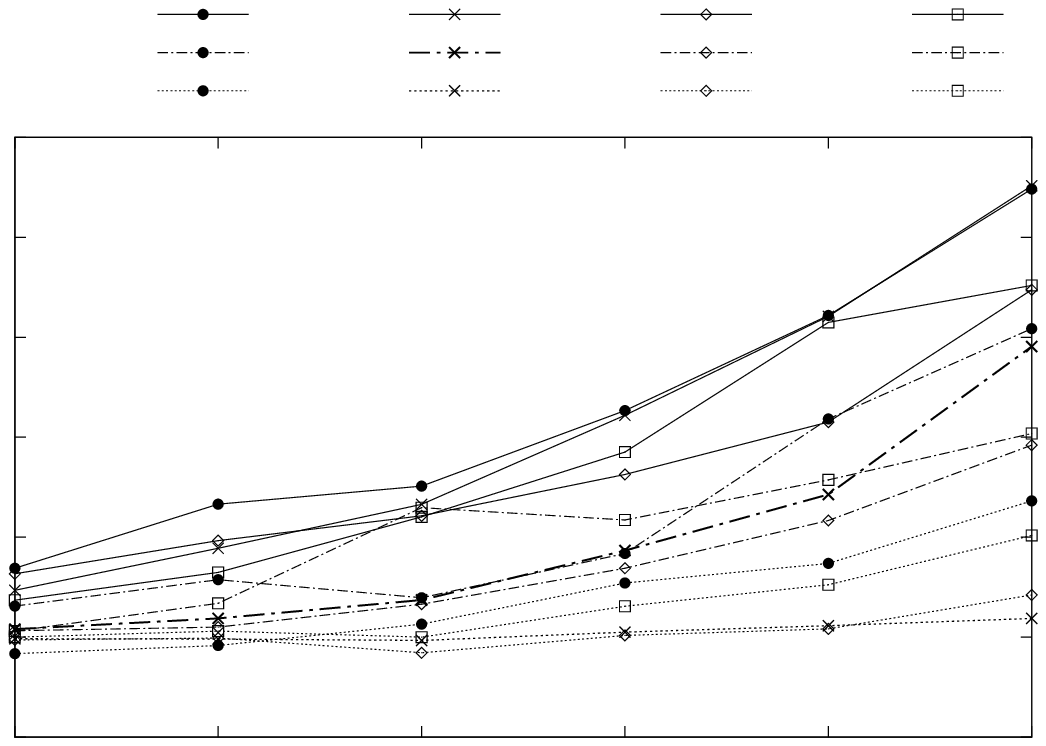}}\gplfronttext
  \end{picture}\endgroup
 \end{adjustbox}
          \caption{Results for \labelflipping\ with monolithic ($\nbasemodel$$=$$1$) and the smallest ($\nbasemodel$$=$$3$) and largest ($\nbasemodel$$=$$21$) ensemble models for datasets M2, AM, SB, and DR.}
          \label{fig:flipping-overall}
\end{figure}

Figure~\ref{fig:flipping-overall} shows $\deltaf$ for the monolithic model ($\nbasemodel$$=$$1$) and the smallest ($\nbasemodel$$=$$3$) and largest ensembles ($\nbasemodel$$=$$21$) varying the datasets and the percentage of poisoned data points $\extentpoint$. As expected, the monolithic model always experiences the largest accuracy decrease. The decrease ranges from $-3.449$ (with \extentpoint$=$$10$) to $-22.414$ (with \extentpoint$=$$35$) with minimum accuracy of $69.458$ for M2; from $-2.344$ to $-22.578$ with minimum accuracy $76.520$ for AM; from $-3.173$ to $-17.380$ with minimum accuracy $77.517$ for SB; from $-1.852$ to $-17.593$ with minimum accuracy $52.315$ for DR.
Instead, our ensemble approach shows higher robustness and keeps the accuracy drop under control.
This can be noticed even with the smallest ensemble $\nbasemodel$$=$$3$, 
where \deltaf=$-9.225$ with $\nbasemodel$$=$$1$ and \deltaf=$-4.740$ with $\nbasemodel$$=$$3$ on average on the 4 datasets. 
More in detail, when considering dataset M2, \deltaf\ increases from $-3.449$ to $-1.560$ with $\extentpoint$$=$$10$, and from $-22.414$ to $-15.436$ with $\extentpoint$$=$$35$ ($-6.158$ on average with \nbasemodel$=$$3$).
When considering dataset AM, \deltaf\ increases from $-2.344$ to $-0.402$ with $\extentpoint$$=$$10$ and from $-22.578$ to $-14.535$ with $\extentpoint$$=$$35$ ($-4.868$ on average with \nbasemodel$=$$3$).
When considering dataset SB, \deltaf\ increases from $-3.173$ to $-0.322$ with $\extentpoint$$=$$10$ and from $-17.380$ to $-9.609$ with $\extentpoint$$=$$35$ ($-3.563$ on average with \nbasemodel$=$$3$).
When considering dataset DR, 
\deltaf\ increases from $-1.852$ to $-0.309$ with $\extentpoint$$=$$10$ and from $-17.593$ to $-10.185$ with $\extentpoint$$=$$35$ ($-5.401$ on average with \nbasemodel$=$$3$).
We can therefore observe that, when the number of random forests $\nbasemodel$ increases, $\deltaf$ increases too, as Figure~\ref{fig:flipping-overall} shows.

Finally, when the number of random forests $\nbasemodel$ is greater than $9$, \deltaf\ improves significantly on all the datasets. This is clear with datasets AM and SB, where \deltaf\ are higher than $-1$ in almost all configurations.
A similar trend can be observed in dataset M2 though the impact of poisoning (\extentpoint) on the model accuracy is higher: \deltaf=$-3.202$ in the worst case with \extentpoint$\le$$20$, \deltaf=$-11.330$ with \extentpoint$>$$20$. 

Overall, our results show that plain random forests are sensitive to \labelflipping, but its effect can be easily counteracted using our ensemble approach. 
In particular, \accuracypoisoned$\le$\accuracyclean$\pm$$1.438$ on average with our ensemble approach,  \accuracypoisoned$\le$\accuracyclean$\pm$$9.225$ with the monolithic model.

\subsection{Other Attacks}\label{subsubsec:exp-result-other}
\begin{figure}[t]
  \begin{adjustbox}{max totalsize={.49\textwidth}{\textheight},center}
  \centering
 \begingroup
  \makeatletter
  \providecommand\color[2][]{\GenericError{(gnuplot) \space\space\space\@spaces}{Package color not loaded in conjunction with
      terminal option `colourtext'}{See the gnuplot documentation for explanation.}{Either use 'blacktext' in gnuplot or load the package
      color.sty in LaTeX.}\renewcommand\color[2][]{}}\providecommand\includegraphics[2][]{\GenericError{(gnuplot) \space\space\space\@spaces}{Package graphicx or graphics not loaded}{See the gnuplot documentation for explanation.}{The gnuplot epslatex terminal needs graphicx.sty or graphics.sty.}\renewcommand\includegraphics[2][]{}}\providecommand\rotatebox[2]{#2}\@ifundefined{ifGPcolor}{\newif\ifGPcolor
    \GPcolorfalse
  }{}\@ifundefined{ifGPblacktext}{\newif\ifGPblacktext
    \GPblacktexttrue
  }{}\let\gplgaddtomacro\g@addto@macro
\gdef\gplbacktext{}\gdef\gplfronttext{}\makeatother
  \ifGPblacktext
\def\colorrgb#1{}\def\colorgray#1{}\else
\ifGPcolor
      \def\colorrgb#1{\color[rgb]{#1}}\def\colorgray#1{\color[gray]{#1}}\expandafter\def\csname LTw\endcsname{\color{white}}\expandafter\def\csname LTb\endcsname{\color{black}}\expandafter\def\csname LTa\endcsname{\color{black}}\expandafter\def\csname LT0\endcsname{\color[rgb]{1,0,0}}\expandafter\def\csname LT1\endcsname{\color[rgb]{0,1,0}}\expandafter\def\csname LT2\endcsname{\color[rgb]{0,0,1}}\expandafter\def\csname LT3\endcsname{\color[rgb]{1,0,1}}\expandafter\def\csname LT4\endcsname{\color[rgb]{0,1,1}}\expandafter\def\csname LT5\endcsname{\color[rgb]{1,1,0}}\expandafter\def\csname LT6\endcsname{\color[rgb]{0,0,0}}\expandafter\def\csname LT7\endcsname{\color[rgb]{1,0.3,0}}\expandafter\def\csname LT8\endcsname{\color[rgb]{0.5,0.5,0.5}}\else
\def\colorrgb#1{\color{black}}\def\colorgray#1{\color[gray]{#1}}\expandafter\def\csname LTw\endcsname{\color{white}}\expandafter\def\csname LTb\endcsname{\color{black}}\expandafter\def\csname LTa\endcsname{\color{black}}\expandafter\def\csname LT0\endcsname{\color{black}}\expandafter\def\csname LT1\endcsname{\color{black}}\expandafter\def\csname LT2\endcsname{\color{black}}\expandafter\def\csname LT3\endcsname{\color{black}}\expandafter\def\csname LT4\endcsname{\color{black}}\expandafter\def\csname LT5\endcsname{\color{black}}\expandafter\def\csname LT6\endcsname{\color{black}}\expandafter\def\csname LT7\endcsname{\color{black}}\expandafter\def\csname LT8\endcsname{\color{black}}\fi
  \fi
    \setlength{\unitlength}{0.0500bp}\ifx\gptboxheight\undefined \newlength{\gptboxheight}\newlength{\gptboxwidth}\newsavebox{\gptboxtext}\fi \setlength{\fboxrule}{0.5pt}\setlength{\fboxsep}{1pt}\definecolor{tbcol}{rgb}{1,1,1}\begin{picture}(7200.00,5040.00)\gplgaddtomacro\gplbacktext{\csname LTb\endcsname \put(814,704){\makebox(0,0)[r]{\strut{}$60$}}\put(814,1136){\makebox(0,0)[r]{\strut{}$65$}}\put(814,1568){\makebox(0,0)[r]{\strut{}$70$}}\put(814,2000){\makebox(0,0)[r]{\strut{}$75$}}\put(814,2432){\makebox(0,0)[r]{\strut{}$80$}}\put(814,2863){\makebox(0,0)[r]{\strut{}$85$}}\put(814,3295){\makebox(0,0)[r]{\strut{}$90$}}\put(814,3727){\makebox(0,0)[r]{\strut{}$95$}}\put(814,4159){\makebox(0,0)[r]{\strut{}$100$}}\put(946,484){\makebox(0,0){\strut{}$1$}}\put(1504,484){\makebox(0,0){\strut{}$3$}}\put(2062,484){\makebox(0,0){\strut{}$5$}}\put(2619,484){\makebox(0,0){\strut{}$7$}}\put(3177,484){\makebox(0,0){\strut{}$9$}}\put(3735,484){\makebox(0,0){\strut{}$11$}}\put(4293,484){\makebox(0,0){\strut{}$13$}}\put(4851,484){\makebox(0,0){\strut{}$15$}}\put(5408,484){\makebox(0,0){\strut{}$17$}}\put(5966,484){\makebox(0,0){\strut{}$19$}}\put(6524,484){\makebox(0,0){\strut{}$21$}}}\gplgaddtomacro\gplfronttext{\csname LTb\endcsname \put(209,2431){\rotatebox{-270}{\makebox(0,0){\strut{}\large \accuracypoisoned}}}\put(3874,154){\makebox(0,0){\strut{}\large \nbasemodel}}\put(1504,4867){\makebox(0,0)[r]{\strut{}M2 - zero}}\put(1504,4647){\makebox(0,0)[r]{\strut{}M2 - noise}}\put(1504,4427){\makebox(0,0)[r]{\strut{}M2 - OoR}}\put(3151,4867){\makebox(0,0)[r]{\strut{}AM - zero}}\put(3151,4647){\makebox(0,0)[r]{\strut{}AM - noise}}\put(3151,4427){\makebox(0,0)[r]{\strut{}AM - OoR}}\put(4798,4867){\makebox(0,0)[r]{\strut{}SB - zero}}\put(4798,4647){\makebox(0,0)[r]{\strut{}SB - noise}}\put(4798,4427){\makebox(0,0)[r]{\strut{}SB - OoR}}\put(6445,4867){\makebox(0,0)[r]{\strut{}DR - zero}}\put(6445,4647){\makebox(0,0)[r]{\strut{}DR - noise}}\put(6445,4427){\makebox(0,0)[r]{\strut{}DR - OoR}}}\gplbacktext
    \put(0,0){\includegraphics[width={360.00bp},height={252.00bp}]{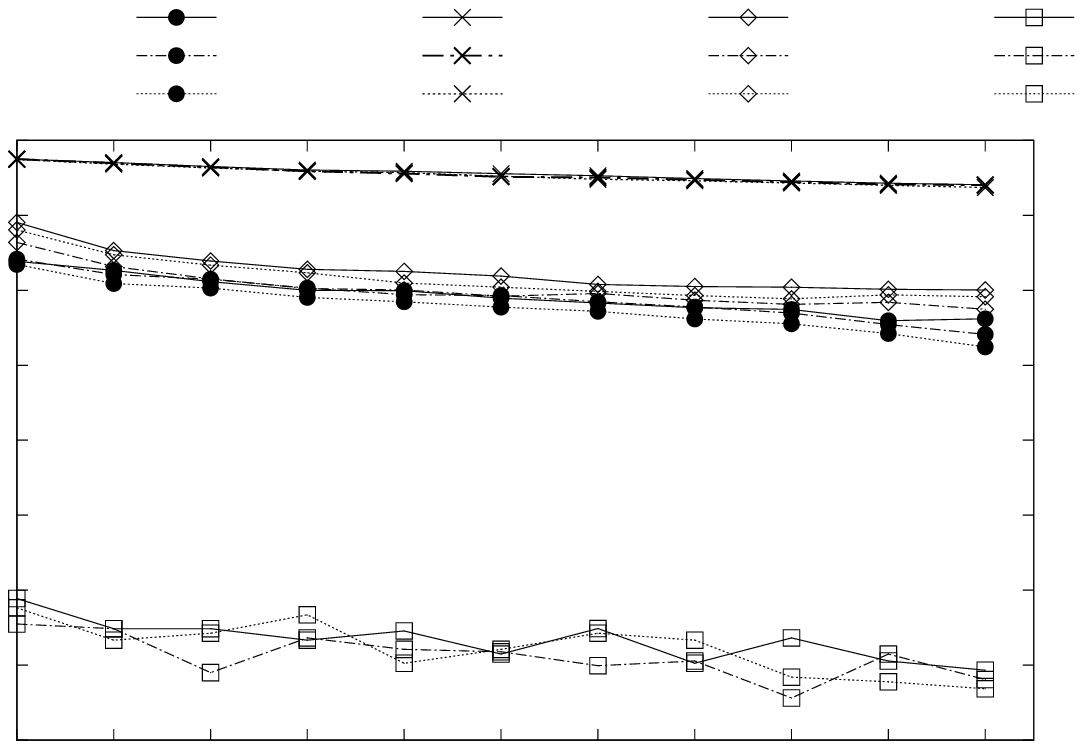}}\gplfronttext
  \end{picture}\endgroup
 \end{adjustbox}
          \caption{Results for other attacks averaged over \extentpoint\ with perturbations \zeroing, \noising, and \outofranging\ abbreviated as \emph{zero}, \emph{noise}, and \emph{OoR}, respectively.}
          \label{fig:other-overall}
\end{figure}

Figure~\ref{fig:other-overall} shows the accuracy degradation for perturbations \zeroing, \noising, and \outofranging\ in Section~\ref{sec:attack} varying the  datasets.
\deltaf\ does not show any major trends and is not presented in Figure~\ref{fig:other-overall}, being $-0.503$ on average (with $\sigma$$\approx$$0.925$).

Our results show two clear trends opposed to the trends retrieved with label flipping.
First, \zeroing, \noising, and \outofranging\ marginally affect the monolithic model, with \deltaf$=$$-0.401$ on average (\deltaf$=$$-0.158$ for \zeroing, \deltaf$=$$-0.507$ for \noising, and \deltaf$=$$-0.539$ for \outofranging). 
As a consequence, there are no major improvements in \deltaf\ when using our ensemble approach, with \deltaf$=$$-0.514$ on average (\deltaf$=$$-0.219$ for \zeroing, \deltaf$=$$-0.562$ for \noising, and \deltaf$=$$-0.761$ for \outofranging).

Second, as depicted in Figure~\ref{fig:other-overall}, the accuracy decreases as the number of random forests \nbasemodel\ increases, but with a relatively small average difference of $2.755$ between \nbasemodel$=$$3$ and \nbasemodel$=$$21$. 
This decrease is higher for M2 in all perturbations with a decrease of $3.818$ (from $90.941$ with \nbasemodel$=$$3$ to $87.123$ with \nbasemodel$=$$21$, on average), followed by DR with a decrease of $2.898$ (from $68.038$ with \nbasemodel$=$$3$ to $65.140$ with \nbasemodel$=$$21$, on average), SB with a decrease of $2.802$ (from $92.631$ with \nbasemodel$=$$3$ to $89.829$ with \nbasemodel$=$$21$, on average), and finally AM with a decrease of $1.501$ (from $98.469$ with \nbasemodel$=$$3$ to $96.968$ with \nbasemodel$=$$21$, on average).

Overall, our results show that monolithic models are significantly less sensitive to perturbations \zeroing, \noising, and \outofranging\ than label flipping, with \deltaf\ always larger than $-0.492$. In particular, \accuracypoisoned$\le$\accuracyclean$\pm$$0.503$ on average. 

 \section{Sustainability of Our Approach}\label{subsec:exp-resource}
We evaluated the sustainability of our ensemble approach measuring the execution time (Section~\ref{subsubsec:exp-resource-time}) and resource consumption (Section~\ref{subsubsec:exp-resource-cpu}).

\subsection{Settings}\label{subsec:exp-resource-setting}
We recall that our experiments have been executed on a VM equipped with 16 vCPUs Intel Core Processor (Broadwell, no TSX) 2.00 GHz and 48 GBs of RAM. Our experiments varied the number of models in the ensemble in \nbasemodel$\in$$\{1,$ $3,$ $5,$ $7,$ $9,$ $11,$ $13,$ $15,$ $17,$ $19,$ $21\}$, the percentage of used data points (i.e., dataset cardinality) in \dataPointsCount$\in$$\{10,$ $25,$ $50,$ $75,$ $100\}$, and the percentage of features in \featuresCount$\in$$\{10,$ $25,$ $50,$ $75,$ $100\}$. 

We note that, as shown in Figure~\ref{fig:exp-resource-time-varying-n}, the execution time for model training (step 4 in Section~\ref{subsec:exp-process-benchmark-process}) is negligible (less than 0.2s) when datasets  M2, SB, and DR are used, while it is over 2s in the worst case when dataset AM is used. The execution time of training sets creation (step 3 in Section~\ref{subsec:exp-process-benchmark-process}) is constant for each dataset and does not depend on the number of models \nbasemodel\ (e.g., 2.46s, with $\sigma$$=$$\pm10$ms, in the worst case for dataset AM).
For these reasons, we evaluated the sustainability of our approach using dataset AM in Table~\ref{tab:datasets} (worst case scenario). Execution time was evaluated considering only step 4 in Section~\ref{subsec:exp-process-benchmark-process}, while resource consumption considered steps 3 and 4 in Section~\ref{subsec:exp-process-benchmark-process}.

The experiments have been repeated 5 times and the reported results correspond to the average over repetitions. 

\subsection{Execution Time}\label{subsubsec:exp-resource-time}
We measured the impact of the training process (step 4 in Section~\ref{subsec:exp-process-benchmark-process}) on the execution time of our approach by varying the number of models \nbasemodel\ in the ensemble, the dataset cardinality (\dataPointsCount), and the feature cardinality (\featuresCount). 

\vspace{.5em}

\begin{figure}[!t]
  \centering
  \begin{adjustbox}{max totalsize={\linewidth}{\textheight},center}
\begingroup
  \makeatletter
  \providecommand\color[2][]{\GenericError{(gnuplot) \space\space\space\@spaces}{Package color not loaded in conjunction with
      terminal option `colourtext'}{See the gnuplot documentation for explanation.}{Either use 'blacktext' in gnuplot or load the package
      color.sty in LaTeX.}\renewcommand\color[2][]{}}\providecommand\includegraphics[2][]{\GenericError{(gnuplot) \space\space\space\@spaces}{Package graphicx or graphics not loaded}{See the gnuplot documentation for explanation.}{The gnuplot epslatex terminal needs graphicx.sty or graphics.sty.}\renewcommand\includegraphics[2][]{}}\providecommand\rotatebox[2]{#2}\@ifundefined{ifGPcolor}{\newif\ifGPcolor
    \GPcolorfalse
  }{}\@ifundefined{ifGPblacktext}{\newif\ifGPblacktext
    \GPblacktexttrue
  }{}\let\gplgaddtomacro\g@addto@macro
\gdef\gplbacktext{}\gdef\gplfronttext{}\makeatother
  \ifGPblacktext
\def\colorrgb#1{}\def\colorgray#1{}\else
\ifGPcolor
      \def\colorrgb#1{\color[rgb]{#1}}\def\colorgray#1{\color[gray]{#1}}\expandafter\def\csname LTw\endcsname{\color{white}}\expandafter\def\csname LTb\endcsname{\color{black}}\expandafter\def\csname LTa\endcsname{\color{black}}\expandafter\def\csname LT0\endcsname{\color[rgb]{1,0,0}}\expandafter\def\csname LT1\endcsname{\color[rgb]{0,1,0}}\expandafter\def\csname LT2\endcsname{\color[rgb]{0,0,1}}\expandafter\def\csname LT3\endcsname{\color[rgb]{1,0,1}}\expandafter\def\csname LT4\endcsname{\color[rgb]{0,1,1}}\expandafter\def\csname LT5\endcsname{\color[rgb]{1,1,0}}\expandafter\def\csname LT6\endcsname{\color[rgb]{0,0,0}}\expandafter\def\csname LT7\endcsname{\color[rgb]{1,0.3,0}}\expandafter\def\csname LT8\endcsname{\color[rgb]{0.5,0.5,0.5}}\else
\def\colorrgb#1{\color{black}}\def\colorgray#1{\color[gray]{#1}}\expandafter\def\csname LTw\endcsname{\color{white}}\expandafter\def\csname LTb\endcsname{\color{black}}\expandafter\def\csname LTa\endcsname{\color{black}}\expandafter\def\csname LT0\endcsname{\color{black}}\expandafter\def\csname LT1\endcsname{\color{black}}\expandafter\def\csname LT2\endcsname{\color{black}}\expandafter\def\csname LT3\endcsname{\color{black}}\expandafter\def\csname LT4\endcsname{\color{black}}\expandafter\def\csname LT5\endcsname{\color{black}}\expandafter\def\csname LT6\endcsname{\color{black}}\expandafter\def\csname LT7\endcsname{\color{black}}\expandafter\def\csname LT8\endcsname{\color{black}}\fi
  \fi
    \setlength{\unitlength}{0.0500bp}\ifx\gptboxheight\undefined \newlength{\gptboxheight}\newlength{\gptboxwidth}\newsavebox{\gptboxtext}\fi \setlength{\fboxrule}{0.5pt}\setlength{\fboxsep}{1pt}\definecolor{tbcol}{rgb}{1,1,1}\begin{picture}(7200.00,5040.00)\gplgaddtomacro\gplbacktext{\csname LTb\endcsname \put(814,704){\makebox(0,0)[r]{\strut{}$0$}}\put(814,1292){\makebox(0,0)[r]{\strut{}$0.5$}}\put(814,1880){\makebox(0,0)[r]{\strut{}$1$}}\put(814,2468){\makebox(0,0)[r]{\strut{}$1.5$}}\put(814,3055){\makebox(0,0)[r]{\strut{}$2$}}\put(814,3643){\makebox(0,0)[r]{\strut{}$2.5$}}\put(814,4231){\makebox(0,0)[r]{\strut{}$3$}}\put(814,4819){\makebox(0,0)[r]{\strut{}$3.5$}}\put(946,484){\makebox(0,0){\strut{}$1$}}\put(1504,484){\makebox(0,0){\strut{}$3$}}\put(2062,484){\makebox(0,0){\strut{}$5$}}\put(2619,484){\makebox(0,0){\strut{}$7$}}\put(3177,484){\makebox(0,0){\strut{}$9$}}\put(3735,484){\makebox(0,0){\strut{}$11$}}\put(4293,484){\makebox(0,0){\strut{}$13$}}\put(4851,484){\makebox(0,0){\strut{}$15$}}\put(5408,484){\makebox(0,0){\strut{}$17$}}\put(5966,484){\makebox(0,0){\strut{}$19$}}\put(6524,484){\makebox(0,0){\strut{}$21$}}}\gplgaddtomacro\gplfronttext{\csname LTb\endcsname \put(209,2761){\rotatebox{-270}{\makebox(0,0){\strut{}\large Execution time (s)}}}\put(3874,154){\makebox(0,0){\strut{}\large Number of models \nbasemodel}}\put(5816,4646){\makebox(0,0)[r]{\strut{}M2}}\put(5816,4426){\makebox(0,0)[r]{\strut{}AM}}\put(5816,4206){\makebox(0,0)[r]{\strut{}SB}}\put(5816,3986){\makebox(0,0)[r]{\strut{}DR}}}\gplbacktext
    \put(0,0){\includegraphics[width={360.00bp},height={252.00bp}]{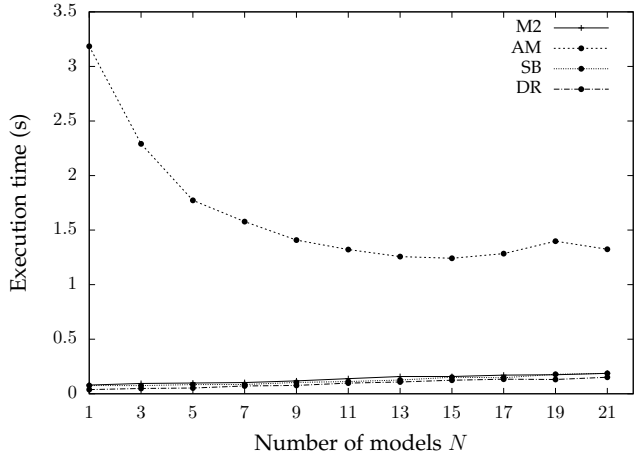}}\gplfronttext
  \end{picture}\endgroup
 \end{adjustbox}
  \caption{Execution time for model training varying \nbasemodel\ and datasets M2, AM, SB, and DR.}
  \label{fig:exp-resource-time-varying-n}
\end{figure}

\noindent \textbf{Execution time varying \nbasemodel.} Figure~\ref{fig:exp-resource-time-varying-n} shows the training execution time according to the number of models \nbasemodel\ of the ensemble. 
The execution time is affected by two main dimensions: \emph{i)} the size of the datasets, \emph{ii)} the parallelization approach used to create the training sets and train the ensemble. We recall that \nbasemodel=1 refers to the monolithic model, while 3$\leq$\nbasemodel$\leq$21 to our ensemble approach. 

Figure~\ref{fig:exp-resource-time-varying-n} shows that our approach is sustainable regardless the value of \nbasemodel, and its execution time is 2.29s in the worst case for \nbasemodel$=$$3$ and dataset AM. Such dataset shows a decreasing trend, where the training time is inversely proportional to the number of models \nbasemodel. As such, the monolithic model requires the highest training time (3.18s). These results are due to: \emph{i)} the capability of our implementation to train the base models in the enemble in parallel, thus exploiting all cores (16) of the VMs used for the experiments; \emph{ii)} the cardinality of each single training set, which decreases (\dataPointsCount/\nbasemodel) as the number of models \nbasemodel\ in the ensemble increases. The decreasing trend is more evident for \nbasemodel$<$$9$ in Figure~\ref{fig:exp-resource-time-varying-n}, reaching a time-stability with \nbasemodel$=$$9$. With \nbasemodel$\ge$$17$, the trend slightly grows because all 16 cores are used, and the thread scheduling affects the performance.

Figure~\ref{fig:exp-resource-time-varying-n} also shows that the trend followed by datasets M2, SB, and DR is significantly different from the one followed by dataset AM. This is mostly due to the cardinality of the datasets, which is particularly low for dataset M2 (2,034 data points), SB (3,626) and DR (1,080). As a consequence, training cannot fully benefit from parallelism, and training time increases as the number of models \nbasemodel\ increases, being 0.187s in the worst case for \nbasemodel$=$$21$ on dataset SB.

\vspace{.5em}

\begin{figure}[!t]
  \centering
  \begin{adjustbox}{max totalsize={\linewidth}{\textheight},center}
    \begingroup
  \makeatletter
  \providecommand\color[2][]{\GenericError{(gnuplot) \space\space\space\@spaces}{Package color not loaded in conjunction with
      terminal option `colourtext'}{See the gnuplot documentation for explanation.}{Either use 'blacktext' in gnuplot or load the package
      color.sty in LaTeX.}\renewcommand\color[2][]{}}\providecommand\includegraphics[2][]{\GenericError{(gnuplot) \space\space\space\@spaces}{Package graphicx or graphics not loaded}{See the gnuplot documentation for explanation.}{The gnuplot epslatex terminal needs graphicx.sty or graphics.sty.}\renewcommand\includegraphics[2][]{}}\providecommand\rotatebox[2]{#2}\@ifundefined{ifGPcolor}{\newif\ifGPcolor
    \GPcolorfalse
  }{}\@ifundefined{ifGPblacktext}{\newif\ifGPblacktext
    \GPblacktexttrue
  }{}\let\gplgaddtomacro\g@addto@macro
\gdef\gplbacktext{}\gdef\gplfronttext{}\makeatother
  \ifGPblacktext
\def\colorrgb#1{}\def\colorgray#1{}\else
\ifGPcolor
      \def\colorrgb#1{\color[rgb]{#1}}\def\colorgray#1{\color[gray]{#1}}\expandafter\def\csname LTw\endcsname{\color{white}}\expandafter\def\csname LTb\endcsname{\color{black}}\expandafter\def\csname LTa\endcsname{\color{black}}\expandafter\def\csname LT0\endcsname{\color[rgb]{1,0,0}}\expandafter\def\csname LT1\endcsname{\color[rgb]{0,1,0}}\expandafter\def\csname LT2\endcsname{\color[rgb]{0,0,1}}\expandafter\def\csname LT3\endcsname{\color[rgb]{1,0,1}}\expandafter\def\csname LT4\endcsname{\color[rgb]{0,1,1}}\expandafter\def\csname LT5\endcsname{\color[rgb]{1,1,0}}\expandafter\def\csname LT6\endcsname{\color[rgb]{0,0,0}}\expandafter\def\csname LT7\endcsname{\color[rgb]{1,0.3,0}}\expandafter\def\csname LT8\endcsname{\color[rgb]{0.5,0.5,0.5}}\else
\def\colorrgb#1{\color{black}}\def\colorgray#1{\color[gray]{#1}}\expandafter\def\csname LTw\endcsname{\color{white}}\expandafter\def\csname LTb\endcsname{\color{black}}\expandafter\def\csname LTa\endcsname{\color{black}}\expandafter\def\csname LT0\endcsname{\color{black}}\expandafter\def\csname LT1\endcsname{\color{black}}\expandafter\def\csname LT2\endcsname{\color{black}}\expandafter\def\csname LT3\endcsname{\color{black}}\expandafter\def\csname LT4\endcsname{\color{black}}\expandafter\def\csname LT5\endcsname{\color{black}}\expandafter\def\csname LT6\endcsname{\color{black}}\expandafter\def\csname LT7\endcsname{\color{black}}\expandafter\def\csname LT8\endcsname{\color{black}}\fi
  \fi
    \setlength{\unitlength}{0.0500bp}\ifx\gptboxheight\undefined \newlength{\gptboxheight}\newlength{\gptboxwidth}\newsavebox{\gptboxtext}\fi \setlength{\fboxrule}{0.5pt}\setlength{\fboxsep}{1pt}\definecolor{tbcol}{rgb}{1,1,1}\begin{picture}(7200.00,5040.00)\gplgaddtomacro\gplbacktext{\csname LTb\endcsname \put(814,704){\makebox(0,0)[r]{\strut{}$0$}}\put(814,1292){\makebox(0,0)[r]{\strut{}$0.5$}}\put(814,1880){\makebox(0,0)[r]{\strut{}$1$}}\put(814,2468){\makebox(0,0)[r]{\strut{}$1.5$}}\put(814,3055){\makebox(0,0)[r]{\strut{}$2$}}\put(814,3643){\makebox(0,0)[r]{\strut{}$2.5$}}\put(814,4231){\makebox(0,0)[r]{\strut{}$3$}}\put(814,4819){\makebox(0,0)[r]{\strut{}$3.5$}}\put(946,484){\makebox(0,0){\strut{}$10$}}\put(1597,484){\makebox(0,0){\strut{}$20$}}\put(2248,484){\makebox(0,0){\strut{}$30$}}\put(2898,484){\makebox(0,0){\strut{}$40$}}\put(3549,484){\makebox(0,0){\strut{}$50$}}\put(4200,484){\makebox(0,0){\strut{}$60$}}\put(4851,484){\makebox(0,0){\strut{}$70$}}\put(5501,484){\makebox(0,0){\strut{}$80$}}\put(6152,484){\makebox(0,0){\strut{}$90$}}\put(6803,484){\makebox(0,0){\strut{}$100$}}}\gplgaddtomacro\gplfronttext{\csname LTb\endcsname \put(209,2761){\rotatebox{-270}{\makebox(0,0){\strut{}\large Execution time (s)}}}\put(3874,154){\makebox(0,0){\strut{}\large Percentage of data points \dataPointsCount}}\put(1078,4646){\makebox(0,0)[l]{\strut{}\nbasemodel$=$$1$}}\put(1078,4426){\makebox(0,0)[l]{\strut{}\nbasemodel$=$$3$}}\put(1078,4206){\makebox(0,0)[l]{\strut{}\nbasemodel$=$$11$}}\put(1078,3986){\makebox(0,0)[l]{\strut{}\nbasemodel$=$$21$}}}\gplbacktext
    \put(0,0){\includegraphics[width={360.00bp},height={252.00bp}]{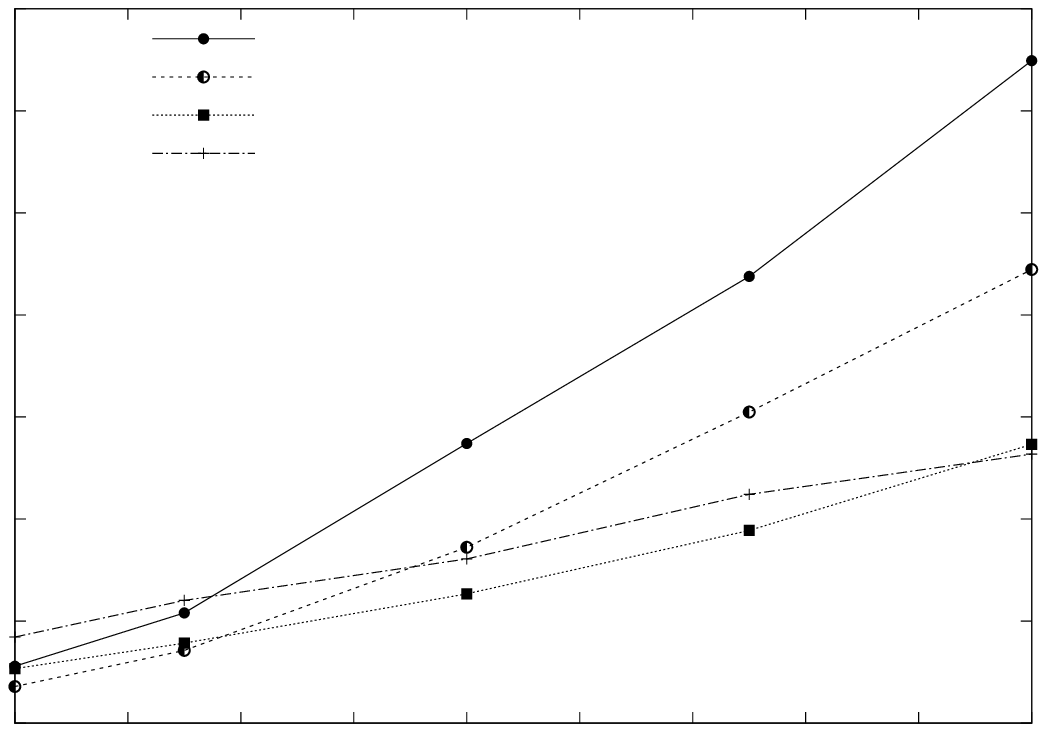}}\gplfronttext
  \end{picture}\endgroup
   \end{adjustbox}
  \caption{Execution time varying dataset cardinality.}
  \label{fig:exp-resource-time-varying-cardinality}
\end{figure}

\noindent \textbf{Execution time varying \dataPointsCount.} Figure~\ref{fig:exp-resource-time-varying-cardinality} shows the training execution time varying the cardinality of dataset Android malware. Specifically, we randomly subsampled the dataset to achieve a percentage of data points in $\{10,$ $25,$ $50,$ $75,$ $100\}$
with respect to the original dataset, and measured the execution time with \nbasemodel$\in$$\{1,$ $3,$ $11,$ $21\}$.
Our results show a sustainable trend that linearly increases for all \nbasemodel\ with a worst case of 3.246s obtained with the monolithic model. We note that, up to 25\% of the dataset, the training time is almost similar for all \nbasemodel. For higher percentages, the execution time for \nbasemodel$=$$1$ and \nbasemodel$=$$3$ starts raising more steeply. Comparing \nbasemodel$=$$1$ and \nbasemodel$=$$3$, we can already appreciate a good gain when our ensemble approach is used, becoming more evident with larger values of \nbasemodel.

\vspace{.5em}

\begin{figure}[!t]
  \centering
  \begin{adjustbox}{max totalsize={\linewidth}{\textheight},center}
    \begingroup
  \makeatletter
  \providecommand\color[2][]{\GenericError{(gnuplot) \space\space\space\@spaces}{Package color not loaded in conjunction with
      terminal option `colourtext'}{See the gnuplot documentation for explanation.}{Either use 'blacktext' in gnuplot or load the package
      color.sty in LaTeX.}\renewcommand\color[2][]{}}\providecommand\includegraphics[2][]{\GenericError{(gnuplot) \space\space\space\@spaces}{Package graphicx or graphics not loaded}{See the gnuplot documentation for explanation.}{The gnuplot epslatex terminal needs graphicx.sty or graphics.sty.}\renewcommand\includegraphics[2][]{}}\providecommand\rotatebox[2]{#2}\@ifundefined{ifGPcolor}{\newif\ifGPcolor
    \GPcolorfalse
  }{}\@ifundefined{ifGPblacktext}{\newif\ifGPblacktext
    \GPblacktexttrue
  }{}\let\gplgaddtomacro\g@addto@macro
\gdef\gplbacktext{}\gdef\gplfronttext{}\makeatother
  \ifGPblacktext
\def\colorrgb#1{}\def\colorgray#1{}\else
\ifGPcolor
      \def\colorrgb#1{\color[rgb]{#1}}\def\colorgray#1{\color[gray]{#1}}\expandafter\def\csname LTw\endcsname{\color{white}}\expandafter\def\csname LTb\endcsname{\color{black}}\expandafter\def\csname LTa\endcsname{\color{black}}\expandafter\def\csname LT0\endcsname{\color[rgb]{1,0,0}}\expandafter\def\csname LT1\endcsname{\color[rgb]{0,1,0}}\expandafter\def\csname LT2\endcsname{\color[rgb]{0,0,1}}\expandafter\def\csname LT3\endcsname{\color[rgb]{1,0,1}}\expandafter\def\csname LT4\endcsname{\color[rgb]{0,1,1}}\expandafter\def\csname LT5\endcsname{\color[rgb]{1,1,0}}\expandafter\def\csname LT6\endcsname{\color[rgb]{0,0,0}}\expandafter\def\csname LT7\endcsname{\color[rgb]{1,0.3,0}}\expandafter\def\csname LT8\endcsname{\color[rgb]{0.5,0.5,0.5}}\else
\def\colorrgb#1{\color{black}}\def\colorgray#1{\color[gray]{#1}}\expandafter\def\csname LTw\endcsname{\color{white}}\expandafter\def\csname LTb\endcsname{\color{black}}\expandafter\def\csname LTa\endcsname{\color{black}}\expandafter\def\csname LT0\endcsname{\color{black}}\expandafter\def\csname LT1\endcsname{\color{black}}\expandafter\def\csname LT2\endcsname{\color{black}}\expandafter\def\csname LT3\endcsname{\color{black}}\expandafter\def\csname LT4\endcsname{\color{black}}\expandafter\def\csname LT5\endcsname{\color{black}}\expandafter\def\csname LT6\endcsname{\color{black}}\expandafter\def\csname LT7\endcsname{\color{black}}\expandafter\def\csname LT8\endcsname{\color{black}}\fi
  \fi
    \setlength{\unitlength}{0.0500bp}\ifx\gptboxheight\undefined \newlength{\gptboxheight}\newlength{\gptboxwidth}\newsavebox{\gptboxtext}\fi \setlength{\fboxrule}{0.5pt}\setlength{\fboxsep}{1pt}\definecolor{tbcol}{rgb}{1,1,1}\begin{picture}(7200.00,5040.00)\gplgaddtomacro\gplbacktext{\csname LTb\endcsname \put(814,704){\makebox(0,0)[r]{\strut{}$0$}}\put(814,1292){\makebox(0,0)[r]{\strut{}$0.5$}}\put(814,1880){\makebox(0,0)[r]{\strut{}$1$}}\put(814,2468){\makebox(0,0)[r]{\strut{}$1.5$}}\put(814,3055){\makebox(0,0)[r]{\strut{}$2$}}\put(814,3643){\makebox(0,0)[r]{\strut{}$2.5$}}\put(814,4231){\makebox(0,0)[r]{\strut{}$3$}}\put(814,4819){\makebox(0,0)[r]{\strut{}$3.5$}}\put(946,484){\makebox(0,0){\strut{}$10$}}\put(1597,484){\makebox(0,0){\strut{}$20$}}\put(2248,484){\makebox(0,0){\strut{}$30$}}\put(2898,484){\makebox(0,0){\strut{}$40$}}\put(3549,484){\makebox(0,0){\strut{}$50$}}\put(4200,484){\makebox(0,0){\strut{}$60$}}\put(4851,484){\makebox(0,0){\strut{}$70$}}\put(5501,484){\makebox(0,0){\strut{}$80$}}\put(6152,484){\makebox(0,0){\strut{}$90$}}\put(6803,484){\makebox(0,0){\strut{}$100$}}}\gplgaddtomacro\gplfronttext{\csname LTb\endcsname \put(209,2761){\rotatebox{-270}{\makebox(0,0){\strut{}\large Execution time (s)}}}\put(3874,154){\makebox(0,0){\strut{}\large Percentage of features \featuresCount}}\put(1078,4646){\makebox(0,0)[l]{\strut{}\nbasemodel$=$$1$}}\put(1078,4426){\makebox(0,0)[l]{\strut{}\nbasemodel$=$$3$}}\put(1078,4206){\makebox(0,0)[l]{\strut{}\nbasemodel$=$$11$}}\put(1078,3986){\makebox(0,0)[l]{\strut{}\nbasemodel$=$$21$}}}\gplbacktext
    \put(0,0){\includegraphics[width={360.00bp},height={252.00bp}]{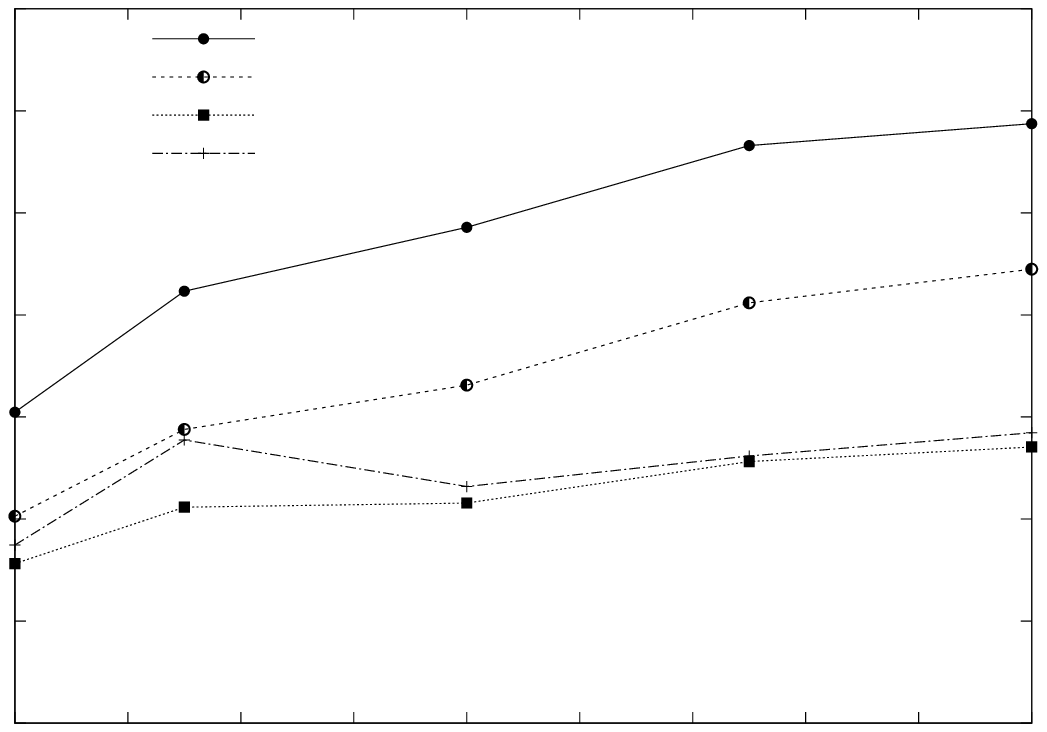}}\gplfronttext
  \end{picture}\endgroup
   \end{adjustbox}
  \caption{Execution time varying number of features.}
  \label{fig:exp-resource-time-varying-features}
\end{figure}

\noindent \textbf{Execution time varying \featuresCount.} Figure~\ref{fig:exp-resource-time-varying-features} shows the training execution time varying the number of features in dataset AM. Specifically, we randomly subsampled the dataset to achieve a percentage of features in $\{10,$ $25,$ $50,$ $75,$ $100\}$
with respect to the original dataset, and measured the execution time with \nbasemodel$\in$$\{1$, $3,$ $11,$ $21\}$.
Our results show a sustainable trend that increases for all \nbasemodel\ following the equation of a parabola with a worst case of 2.937s obtained, also in this case, with the monolithic model. As expected, the execution time grows proportionally to the number of features regardless \nbasemodel, while the slope of the corresponding curves are gradually lower as \nbasemodel\ increases.

\subsection{Resource Consumption}\label{subsubsec:exp-resource-cpu}

\begin{figure*}[!t]
    \centering
    \linespread{0.6}\selectfont\centering
    \begin{adjustbox}{max totalsize={.99\linewidth}{\textheight},center}
    \begin{tabular}{cc}
        \resizebox*{.49\textwidth}{!}{
            \begingroup
  \makeatletter
  \providecommand\color[2][]{\GenericError{(gnuplot) \space\space\space\@spaces}{Package color not loaded in conjunction with
      terminal option `colourtext'}{See the gnuplot documentation for explanation.}{Either use 'blacktext' in gnuplot or load the package
      color.sty in LaTeX.}\renewcommand\color[2][]{}}\providecommand\includegraphics[2][]{\GenericError{(gnuplot) \space\space\space\@spaces}{Package graphicx or graphics not loaded}{See the gnuplot documentation for explanation.}{The gnuplot epslatex terminal needs graphicx.sty or graphics.sty.}\renewcommand\includegraphics[2][]{}}\providecommand\rotatebox[2]{#2}\@ifundefined{ifGPcolor}{\newif\ifGPcolor
    \GPcolorfalse
  }{}\@ifundefined{ifGPblacktext}{\newif\ifGPblacktext
    \GPblacktexttrue
  }{}\let\gplgaddtomacro\g@addto@macro
\gdef\gplbacktext{}\gdef\gplfronttext{}\makeatother
  \ifGPblacktext
\def\colorrgb#1{}\def\colorgray#1{}\else
\ifGPcolor
      \def\colorrgb#1{\color[rgb]{#1}}\def\colorgray#1{\color[gray]{#1}}\expandafter\def\csname LTw\endcsname{\color{white}}\expandafter\def\csname LTb\endcsname{\color{black}}\expandafter\def\csname LTa\endcsname{\color{black}}\expandafter\def\csname LT0\endcsname{\color[rgb]{1,0,0}}\expandafter\def\csname LT1\endcsname{\color[rgb]{0,1,0}}\expandafter\def\csname LT2\endcsname{\color[rgb]{0,0,1}}\expandafter\def\csname LT3\endcsname{\color[rgb]{1,0,1}}\expandafter\def\csname LT4\endcsname{\color[rgb]{0,1,1}}\expandafter\def\csname LT5\endcsname{\color[rgb]{1,1,0}}\expandafter\def\csname LT6\endcsname{\color[rgb]{0,0,0}}\expandafter\def\csname LT7\endcsname{\color[rgb]{1,0.3,0}}\expandafter\def\csname LT8\endcsname{\color[rgb]{0.5,0.5,0.5}}\else
\def\colorrgb#1{\color{black}}\def\colorgray#1{\color[gray]{#1}}\expandafter\def\csname LTw\endcsname{\color{white}}\expandafter\def\csname LTb\endcsname{\color{black}}\expandafter\def\csname LTa\endcsname{\color{black}}\expandafter\def\csname LT0\endcsname{\color{black}}\expandafter\def\csname LT1\endcsname{\color{black}}\expandafter\def\csname LT2\endcsname{\color{black}}\expandafter\def\csname LT3\endcsname{\color{black}}\expandafter\def\csname LT4\endcsname{\color{black}}\expandafter\def\csname LT5\endcsname{\color{black}}\expandafter\def\csname LT6\endcsname{\color{black}}\expandafter\def\csname LT7\endcsname{\color{black}}\expandafter\def\csname LT8\endcsname{\color{black}}\fi
  \fi
    \setlength{\unitlength}{0.0500bp}\ifx\gptboxheight\undefined \newlength{\gptboxheight}\newlength{\gptboxwidth}\newsavebox{\gptboxtext}\fi \setlength{\fboxrule}{0.5pt}\setlength{\fboxsep}{1pt}\definecolor{tbcol}{rgb}{1,1,1}\begin{picture}(7200.00,5040.00)\gplgaddtomacro\gplbacktext{\csname LTb\endcsname \put(682,704){\makebox(0,0)[r]{\strut{}$10$}}\put(682,1260){\makebox(0,0)[r]{\strut{}$20$}}\put(682,1817){\makebox(0,0)[r]{\strut{}$30$}}\put(682,2373){\makebox(0,0)[r]{\strut{}$40$}}\put(682,2930){\makebox(0,0)[r]{\strut{}$50$}}\put(682,3486){\makebox(0,0)[r]{\strut{}$60$}}\put(682,4043){\makebox(0,0)[r]{\strut{}$70$}}\put(682,4599){\makebox(0,0)[r]{\strut{}$80$}}\put(814,484){\makebox(0,0){\strut{}$10$}}\put(1479,484){\makebox(0,0){\strut{}$20$}}\put(2145,484){\makebox(0,0){\strut{}$30$}}\put(2810,484){\makebox(0,0){\strut{}$40$}}\put(3476,484){\makebox(0,0){\strut{}$50$}}\put(4141,484){\makebox(0,0){\strut{}$60$}}\put(4807,484){\makebox(0,0){\strut{}$70$}}\put(5472,484){\makebox(0,0){\strut{}$80$}}\put(6138,484){\makebox(0,0){\strut{}$90$}}\put(6803,484){\makebox(0,0){\strut{}$100$}}}\gplgaddtomacro\gplfronttext{\csname LTb\endcsname \put(209,2651){\rotatebox{-270}{\makebox(0,0){\strut{}\large CPU user time (s)}}}\put(3808,154){\makebox(0,0){\strut{}\large Number of data points \dataPointsCount}}\put(1570,4867){\makebox(0,0)[r]{\strut{}$\nbasemodel$$=$$1$}}\put(2953,4867){\makebox(0,0)[r]{\strut{}$\nbasemodel$$=$$3$}}\put(4336,4867){\makebox(0,0)[r]{\strut{}$\nbasemodel$$=$$11$}}\put(5719,4867){\makebox(0,0)[r]{\strut{}$\nbasemodel$$=$$21$}}}\gplbacktext
    \put(0,0){\includegraphics[width={360.00bp},height={252.00bp}]{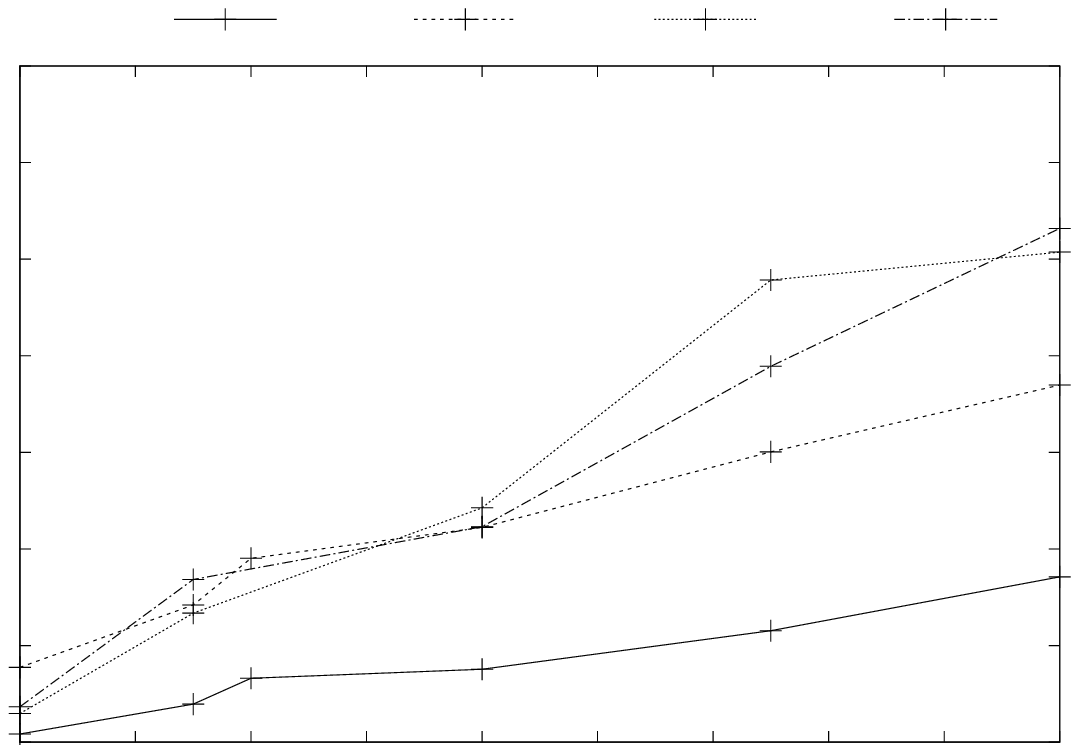}}\gplfronttext
  \end{picture}\endgroup
         }
        &
        \resizebox*{.49\textwidth}{!}{
            \begingroup
  \makeatletter
  \providecommand\color[2][]{\GenericError{(gnuplot) \space\space\space\@spaces}{Package color not loaded in conjunction with
      terminal option `colourtext'}{See the gnuplot documentation for explanation.}{Either use 'blacktext' in gnuplot or load the package
      color.sty in LaTeX.}\renewcommand\color[2][]{}}\providecommand\includegraphics[2][]{\GenericError{(gnuplot) \space\space\space\@spaces}{Package graphicx or graphics not loaded}{See the gnuplot documentation for explanation.}{The gnuplot epslatex terminal needs graphicx.sty or graphics.sty.}\renewcommand\includegraphics[2][]{}}\providecommand\rotatebox[2]{#2}\@ifundefined{ifGPcolor}{\newif\ifGPcolor
    \GPcolorfalse
  }{}\@ifundefined{ifGPblacktext}{\newif\ifGPblacktext
    \GPblacktexttrue
  }{}\let\gplgaddtomacro\g@addto@macro
\gdef\gplbacktext{}\gdef\gplfronttext{}\makeatother
  \ifGPblacktext
\def\colorrgb#1{}\def\colorgray#1{}\else
\ifGPcolor
      \def\colorrgb#1{\color[rgb]{#1}}\def\colorgray#1{\color[gray]{#1}}\expandafter\def\csname LTw\endcsname{\color{white}}\expandafter\def\csname LTb\endcsname{\color{black}}\expandafter\def\csname LTa\endcsname{\color{black}}\expandafter\def\csname LT0\endcsname{\color[rgb]{1,0,0}}\expandafter\def\csname LT1\endcsname{\color[rgb]{0,1,0}}\expandafter\def\csname LT2\endcsname{\color[rgb]{0,0,1}}\expandafter\def\csname LT3\endcsname{\color[rgb]{1,0,1}}\expandafter\def\csname LT4\endcsname{\color[rgb]{0,1,1}}\expandafter\def\csname LT5\endcsname{\color[rgb]{1,1,0}}\expandafter\def\csname LT6\endcsname{\color[rgb]{0,0,0}}\expandafter\def\csname LT7\endcsname{\color[rgb]{1,0.3,0}}\expandafter\def\csname LT8\endcsname{\color[rgb]{0.5,0.5,0.5}}\else
\def\colorrgb#1{\color{black}}\def\colorgray#1{\color[gray]{#1}}\expandafter\def\csname LTw\endcsname{\color{white}}\expandafter\def\csname LTb\endcsname{\color{black}}\expandafter\def\csname LTa\endcsname{\color{black}}\expandafter\def\csname LT0\endcsname{\color{black}}\expandafter\def\csname LT1\endcsname{\color{black}}\expandafter\def\csname LT2\endcsname{\color{black}}\expandafter\def\csname LT3\endcsname{\color{black}}\expandafter\def\csname LT4\endcsname{\color{black}}\expandafter\def\csname LT5\endcsname{\color{black}}\expandafter\def\csname LT6\endcsname{\color{black}}\expandafter\def\csname LT7\endcsname{\color{black}}\expandafter\def\csname LT8\endcsname{\color{black}}\fi
  \fi
    \setlength{\unitlength}{0.0500bp}\ifx\gptboxheight\undefined \newlength{\gptboxheight}\newlength{\gptboxwidth}\newsavebox{\gptboxtext}\fi \setlength{\fboxrule}{0.5pt}\setlength{\fboxsep}{1pt}\definecolor{tbcol}{rgb}{1,1,1}\begin{picture}(7200.00,5040.00)\gplgaddtomacro\gplbacktext{\csname LTb\endcsname \put(946,704){\makebox(0,0)[r]{\strut{}$0$}}\put(946,1137){\makebox(0,0)[r]{\strut{}$500$}}\put(946,1570){\makebox(0,0)[r]{\strut{}$1000$}}\put(946,2002){\makebox(0,0)[r]{\strut{}$1500$}}\put(946,2435){\makebox(0,0)[r]{\strut{}$2000$}}\put(946,2868){\makebox(0,0)[r]{\strut{}$2500$}}\put(946,3301){\makebox(0,0)[r]{\strut{}$3000$}}\put(946,3733){\makebox(0,0)[r]{\strut{}$3500$}}\put(946,4166){\makebox(0,0)[r]{\strut{}$4000$}}\put(946,4599){\makebox(0,0)[r]{\strut{}$4500$}}\put(1078,484){\makebox(0,0){\strut{}$10$}}\put(1714,484){\makebox(0,0){\strut{}$20$}}\put(2350,484){\makebox(0,0){\strut{}$30$}}\put(2986,484){\makebox(0,0){\strut{}$40$}}\put(3622,484){\makebox(0,0){\strut{}$50$}}\put(4259,484){\makebox(0,0){\strut{}$60$}}\put(4895,484){\makebox(0,0){\strut{}$70$}}\put(5531,484){\makebox(0,0){\strut{}$80$}}\put(6167,484){\makebox(0,0){\strut{}$90$}}\put(6803,484){\makebox(0,0){\strut{}$100$}}}\gplgaddtomacro\gplfronttext{\csname LTb\endcsname \put(209,2651){\rotatebox{-270}{\makebox(0,0){\strut{}\large Max memory (MB)}}}\put(3940,154){\makebox(0,0){\strut{}\large Number of data points \dataPointsCount}}\put(1438,4867){\makebox(0,0)[r]{\strut{}$\nbasemodel$$=$$1$}}\put(3085,4867){\makebox(0,0)[r]{\strut{}$\nbasemodel$$=$$3$}}\put(4732,4867){\makebox(0,0)[r]{\strut{}$\nbasemodel$$=$$11$}}\put(6379,4867){\makebox(0,0)[r]{\strut{}$\nbasemodel$$=$$21$}}}\gplbacktext
    \put(0,0){\includegraphics[width={360.00bp},height={252.00bp}]{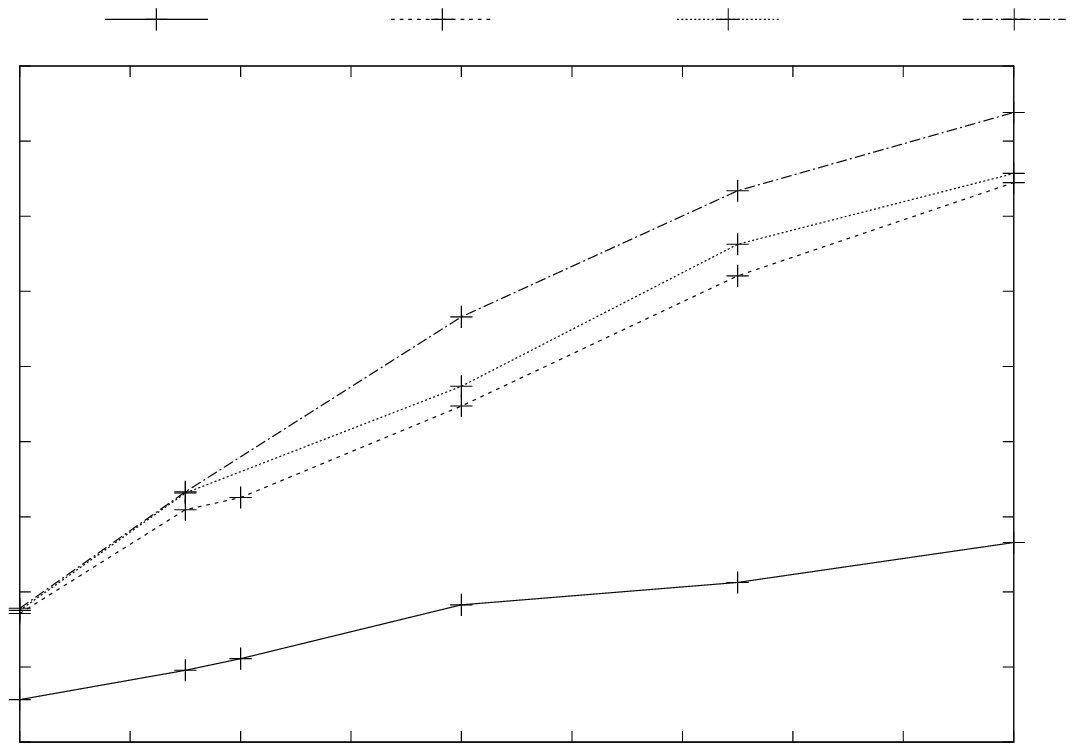}}\gplfronttext
  \end{picture}\endgroup
         }\vspace{.2em}\\
\makecell{\footnotesize (a) CPU user time varying \dataPointsCount, with \nbasemodel$\in$$\{1,$ $3,$ $11,$ $21\}$and  \featuresCount=$100$} & \makecell{\footnotesize(b) Max memory usage varying \dataPointsCount, with \nbasemodel$\in$$\{1,$ $3,$ $11,$ $21\}$ and \featuresCount=$100$}\\
        &\\
\resizebox*{.49\textwidth}{!}{
            \begingroup
  \makeatletter
  \providecommand\color[2][]{\GenericError{(gnuplot) \space\space\space\@spaces}{Package color not loaded in conjunction with
      terminal option `colourtext'}{See the gnuplot documentation for explanation.}{Either use 'blacktext' in gnuplot or load the package
      color.sty in LaTeX.}\renewcommand\color[2][]{}}\providecommand\includegraphics[2][]{\GenericError{(gnuplot) \space\space\space\@spaces}{Package graphicx or graphics not loaded}{See the gnuplot documentation for explanation.}{The gnuplot epslatex terminal needs graphicx.sty or graphics.sty.}\renewcommand\includegraphics[2][]{}}\providecommand\rotatebox[2]{#2}\@ifundefined{ifGPcolor}{\newif\ifGPcolor
    \GPcolorfalse
  }{}\@ifundefined{ifGPblacktext}{\newif\ifGPblacktext
    \GPblacktexttrue
  }{}\let\gplgaddtomacro\g@addto@macro
\gdef\gplbacktext{}\gdef\gplfronttext{}\makeatother
  \ifGPblacktext
\def\colorrgb#1{}\def\colorgray#1{}\else
\ifGPcolor
      \def\colorrgb#1{\color[rgb]{#1}}\def\colorgray#1{\color[gray]{#1}}\expandafter\def\csname LTw\endcsname{\color{white}}\expandafter\def\csname LTb\endcsname{\color{black}}\expandafter\def\csname LTa\endcsname{\color{black}}\expandafter\def\csname LT0\endcsname{\color[rgb]{1,0,0}}\expandafter\def\csname LT1\endcsname{\color[rgb]{0,1,0}}\expandafter\def\csname LT2\endcsname{\color[rgb]{0,0,1}}\expandafter\def\csname LT3\endcsname{\color[rgb]{1,0,1}}\expandafter\def\csname LT4\endcsname{\color[rgb]{0,1,1}}\expandafter\def\csname LT5\endcsname{\color[rgb]{1,1,0}}\expandafter\def\csname LT6\endcsname{\color[rgb]{0,0,0}}\expandafter\def\csname LT7\endcsname{\color[rgb]{1,0.3,0}}\expandafter\def\csname LT8\endcsname{\color[rgb]{0.5,0.5,0.5}}\else
\def\colorrgb#1{\color{black}}\def\colorgray#1{\color[gray]{#1}}\expandafter\def\csname LTw\endcsname{\color{white}}\expandafter\def\csname LTb\endcsname{\color{black}}\expandafter\def\csname LTa\endcsname{\color{black}}\expandafter\def\csname LT0\endcsname{\color{black}}\expandafter\def\csname LT1\endcsname{\color{black}}\expandafter\def\csname LT2\endcsname{\color{black}}\expandafter\def\csname LT3\endcsname{\color{black}}\expandafter\def\csname LT4\endcsname{\color{black}}\expandafter\def\csname LT5\endcsname{\color{black}}\expandafter\def\csname LT6\endcsname{\color{black}}\expandafter\def\csname LT7\endcsname{\color{black}}\expandafter\def\csname LT8\endcsname{\color{black}}\fi
  \fi
    \setlength{\unitlength}{0.0500bp}\ifx\gptboxheight\undefined \newlength{\gptboxheight}\newlength{\gptboxwidth}\newsavebox{\gptboxtext}\fi \setlength{\fboxrule}{0.5pt}\setlength{\fboxsep}{1pt}\definecolor{tbcol}{rgb}{1,1,1}\begin{picture}(7200.00,5040.00)\gplgaddtomacro\gplbacktext{\csname LTb\endcsname \put(682,704){\makebox(0,0)[r]{\strut{}$10$}}\put(682,1260){\makebox(0,0)[r]{\strut{}$20$}}\put(682,1817){\makebox(0,0)[r]{\strut{}$30$}}\put(682,2373){\makebox(0,0)[r]{\strut{}$40$}}\put(682,2930){\makebox(0,0)[r]{\strut{}$50$}}\put(682,3486){\makebox(0,0)[r]{\strut{}$60$}}\put(682,4043){\makebox(0,0)[r]{\strut{}$70$}}\put(682,4599){\makebox(0,0)[r]{\strut{}$80$}}\put(814,484){\makebox(0,0){\strut{}$10$}}\put(1479,484){\makebox(0,0){\strut{}$20$}}\put(2145,484){\makebox(0,0){\strut{}$30$}}\put(2810,484){\makebox(0,0){\strut{}$40$}}\put(3476,484){\makebox(0,0){\strut{}$50$}}\put(4141,484){\makebox(0,0){\strut{}$60$}}\put(4807,484){\makebox(0,0){\strut{}$70$}}\put(5472,484){\makebox(0,0){\strut{}$80$}}\put(6138,484){\makebox(0,0){\strut{}$90$}}\put(6803,484){\makebox(0,0){\strut{}$100$}}}\gplgaddtomacro\gplfronttext{\csname LTb\endcsname \put(209,2651){\rotatebox{-270}{\makebox(0,0){\strut{}\large CPU user time (s)}}}\put(3808,154){\makebox(0,0){\strut{}\large Number of features \featuresCount}}\put(1570,4867){\makebox(0,0)[r]{\strut{}$\nbasemodel$$=$$1$}}\put(2953,4867){\makebox(0,0)[r]{\strut{}$\nbasemodel$$=$$3$}}\put(4336,4867){\makebox(0,0)[r]{\strut{}$\nbasemodel$$=$$11$}}\put(5719,4867){\makebox(0,0)[r]{\strut{}$\nbasemodel$$=$$21$}}}\gplbacktext
    \put(0,0){\includegraphics[width={360.00bp},height={252.00bp}]{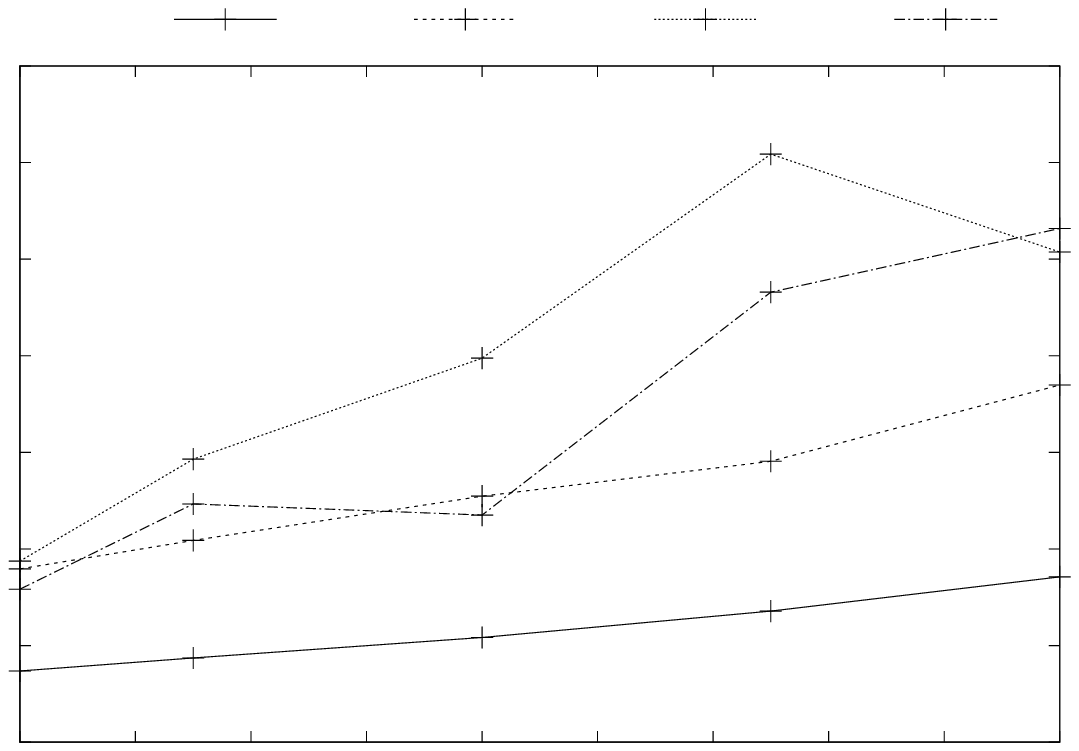}}\gplfronttext
  \end{picture}\endgroup
         }
        &
        \resizebox*{.49\textwidth}{!}{
            \begingroup
  \makeatletter
  \providecommand\color[2][]{\GenericError{(gnuplot) \space\space\space\@spaces}{Package color not loaded in conjunction with
      terminal option `colourtext'}{See the gnuplot documentation for explanation.}{Either use 'blacktext' in gnuplot or load the package
      color.sty in LaTeX.}\renewcommand\color[2][]{}}\providecommand\includegraphics[2][]{\GenericError{(gnuplot) \space\space\space\@spaces}{Package graphicx or graphics not loaded}{See the gnuplot documentation for explanation.}{The gnuplot epslatex terminal needs graphicx.sty or graphics.sty.}\renewcommand\includegraphics[2][]{}}\providecommand\rotatebox[2]{#2}\@ifundefined{ifGPcolor}{\newif\ifGPcolor
    \GPcolorfalse
  }{}\@ifundefined{ifGPblacktext}{\newif\ifGPblacktext
    \GPblacktexttrue
  }{}\let\gplgaddtomacro\g@addto@macro
\gdef\gplbacktext{}\gdef\gplfronttext{}\makeatother
  \ifGPblacktext
\def\colorrgb#1{}\def\colorgray#1{}\else
\ifGPcolor
      \def\colorrgb#1{\color[rgb]{#1}}\def\colorgray#1{\color[gray]{#1}}\expandafter\def\csname LTw\endcsname{\color{white}}\expandafter\def\csname LTb\endcsname{\color{black}}\expandafter\def\csname LTa\endcsname{\color{black}}\expandafter\def\csname LT0\endcsname{\color[rgb]{1,0,0}}\expandafter\def\csname LT1\endcsname{\color[rgb]{0,1,0}}\expandafter\def\csname LT2\endcsname{\color[rgb]{0,0,1}}\expandafter\def\csname LT3\endcsname{\color[rgb]{1,0,1}}\expandafter\def\csname LT4\endcsname{\color[rgb]{0,1,1}}\expandafter\def\csname LT5\endcsname{\color[rgb]{1,1,0}}\expandafter\def\csname LT6\endcsname{\color[rgb]{0,0,0}}\expandafter\def\csname LT7\endcsname{\color[rgb]{1,0.3,0}}\expandafter\def\csname LT8\endcsname{\color[rgb]{0.5,0.5,0.5}}\else
\def\colorrgb#1{\color{black}}\def\colorgray#1{\color[gray]{#1}}\expandafter\def\csname LTw\endcsname{\color{white}}\expandafter\def\csname LTb\endcsname{\color{black}}\expandafter\def\csname LTa\endcsname{\color{black}}\expandafter\def\csname LT0\endcsname{\color{black}}\expandafter\def\csname LT1\endcsname{\color{black}}\expandafter\def\csname LT2\endcsname{\color{black}}\expandafter\def\csname LT3\endcsname{\color{black}}\expandafter\def\csname LT4\endcsname{\color{black}}\expandafter\def\csname LT5\endcsname{\color{black}}\expandafter\def\csname LT6\endcsname{\color{black}}\expandafter\def\csname LT7\endcsname{\color{black}}\expandafter\def\csname LT8\endcsname{\color{black}}\fi
  \fi
    \setlength{\unitlength}{0.0500bp}\ifx\gptboxheight\undefined \newlength{\gptboxheight}\newlength{\gptboxwidth}\newsavebox{\gptboxtext}\fi \setlength{\fboxrule}{0.5pt}\setlength{\fboxsep}{1pt}\definecolor{tbcol}{rgb}{1,1,1}\begin{picture}(7200.00,5040.00)\gplgaddtomacro\gplbacktext{\csname LTb\endcsname \put(946,704){\makebox(0,0)[r]{\strut{}$0$}}\put(946,1137){\makebox(0,0)[r]{\strut{}$500$}}\put(946,1570){\makebox(0,0)[r]{\strut{}$1000$}}\put(946,2002){\makebox(0,0)[r]{\strut{}$1500$}}\put(946,2435){\makebox(0,0)[r]{\strut{}$2000$}}\put(946,2868){\makebox(0,0)[r]{\strut{}$2500$}}\put(946,3301){\makebox(0,0)[r]{\strut{}$3000$}}\put(946,3733){\makebox(0,0)[r]{\strut{}$3500$}}\put(946,4166){\makebox(0,0)[r]{\strut{}$4000$}}\put(946,4599){\makebox(0,0)[r]{\strut{}$4500$}}\put(1078,484){\makebox(0,0){\strut{}$10$}}\put(1714,484){\makebox(0,0){\strut{}$20$}}\put(2350,484){\makebox(0,0){\strut{}$30$}}\put(2986,484){\makebox(0,0){\strut{}$40$}}\put(3622,484){\makebox(0,0){\strut{}$50$}}\put(4259,484){\makebox(0,0){\strut{}$60$}}\put(4895,484){\makebox(0,0){\strut{}$70$}}\put(5531,484){\makebox(0,0){\strut{}$80$}}\put(6167,484){\makebox(0,0){\strut{}$90$}}\put(6803,484){\makebox(0,0){\strut{}$100$}}}\gplgaddtomacro\gplfronttext{\csname LTb\endcsname \put(209,2651){\rotatebox{-270}{\makebox(0,0){\strut{}\large Max memory (MB)}}}\put(3940,154){\makebox(0,0){\strut{}\large Number of features \featuresCount}}\put(1702,4867){\makebox(0,0)[r]{\strut{}$\nbasemodel$$=$$1$}}\put(3085,4867){\makebox(0,0)[r]{\strut{}$\nbasemodel$$=$$3$}}\put(4468,4867){\makebox(0,0)[r]{\strut{}$\nbasemodel$$=$$11$}}\put(5851,4867){\makebox(0,0)[r]{\strut{}$\nbasemodel$$=$$21$}}}\gplbacktext
    \put(0,0){\includegraphics[width={360.00bp},height={252.00bp}]{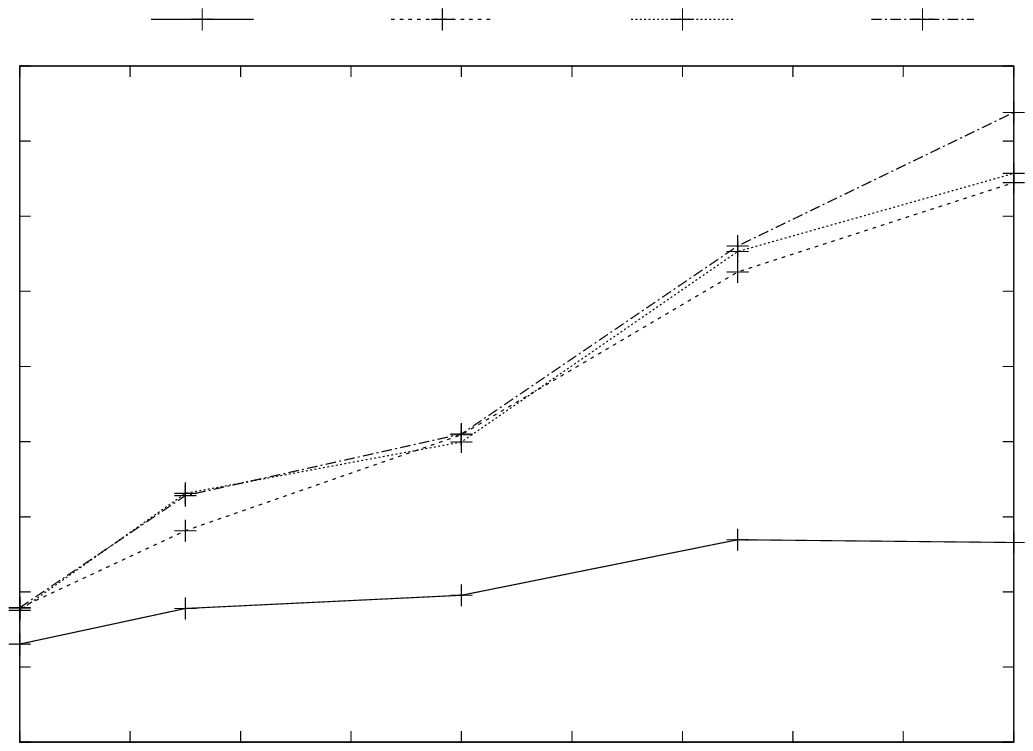}}\gplfronttext
  \end{picture}\endgroup
         }\vspace{.2em}\\
\makecell{\footnotesize (c) CPU user time varying \featuresCount, with \nbasemodel$\in$$\{1,$ $3,$ $11,$ $21\}$ and \dataPointsCount=$100$} & \makecell{\footnotesize (d) Max memory usage varying \featuresCount, with \nbasemodel$\in$$\{1,$ $3,$ $11,$ $21\}$ and \dataPointsCount=$100$}\vspace{1em}\\
        &\\
\resizebox*{.49\textwidth}{!}{
            \begingroup
  \makeatletter
  \providecommand\color[2][]{\GenericError{(gnuplot) \space\space\space\@spaces}{Package color not loaded in conjunction with
      terminal option `colourtext'}{See the gnuplot documentation for explanation.}{Either use 'blacktext' in gnuplot or load the package
      color.sty in LaTeX.}\renewcommand\color[2][]{}}\providecommand\includegraphics[2][]{\GenericError{(gnuplot) \space\space\space\@spaces}{Package graphicx or graphics not loaded}{See the gnuplot documentation for explanation.}{The gnuplot epslatex terminal needs graphicx.sty or graphics.sty.}\renewcommand\includegraphics[2][]{}}\providecommand\rotatebox[2]{#2}\@ifundefined{ifGPcolor}{\newif\ifGPcolor
    \GPcolorfalse
  }{}\@ifundefined{ifGPblacktext}{\newif\ifGPblacktext
    \GPblacktexttrue
  }{}\let\gplgaddtomacro\g@addto@macro
\gdef\gplbacktext{}\gdef\gplfronttext{}\makeatother
  \ifGPblacktext
\def\colorrgb#1{}\def\colorgray#1{}\else
\ifGPcolor
      \def\colorrgb#1{\color[rgb]{#1}}\def\colorgray#1{\color[gray]{#1}}\expandafter\def\csname LTw\endcsname{\color{white}}\expandafter\def\csname LTb\endcsname{\color{black}}\expandafter\def\csname LTa\endcsname{\color{black}}\expandafter\def\csname LT0\endcsname{\color[rgb]{1,0,0}}\expandafter\def\csname LT1\endcsname{\color[rgb]{0,1,0}}\expandafter\def\csname LT2\endcsname{\color[rgb]{0,0,1}}\expandafter\def\csname LT3\endcsname{\color[rgb]{1,0,1}}\expandafter\def\csname LT4\endcsname{\color[rgb]{0,1,1}}\expandafter\def\csname LT5\endcsname{\color[rgb]{1,1,0}}\expandafter\def\csname LT6\endcsname{\color[rgb]{0,0,0}}\expandafter\def\csname LT7\endcsname{\color[rgb]{1,0.3,0}}\expandafter\def\csname LT8\endcsname{\color[rgb]{0.5,0.5,0.5}}\else
\def\colorrgb#1{\color{black}}\def\colorgray#1{\color[gray]{#1}}\expandafter\def\csname LTw\endcsname{\color{white}}\expandafter\def\csname LTb\endcsname{\color{black}}\expandafter\def\csname LTa\endcsname{\color{black}}\expandafter\def\csname LT0\endcsname{\color{black}}\expandafter\def\csname LT1\endcsname{\color{black}}\expandafter\def\csname LT2\endcsname{\color{black}}\expandafter\def\csname LT3\endcsname{\color{black}}\expandafter\def\csname LT4\endcsname{\color{black}}\expandafter\def\csname LT5\endcsname{\color{black}}\expandafter\def\csname LT6\endcsname{\color{black}}\expandafter\def\csname LT7\endcsname{\color{black}}\expandafter\def\csname LT8\endcsname{\color{black}}\fi
  \fi
    \setlength{\unitlength}{0.0500bp}\ifx\gptboxheight\undefined \newlength{\gptboxheight}\newlength{\gptboxwidth}\newsavebox{\gptboxtext}\fi \setlength{\fboxrule}{0.5pt}\setlength{\fboxsep}{1pt}\definecolor{tbcol}{rgb}{1,1,1}\begin{picture}(7200.00,5040.00)\gplgaddtomacro\gplbacktext{\csname LTb\endcsname \put(682,704){\makebox(0,0)[r]{\strut{}$10$}}\put(682,1229){\makebox(0,0)[r]{\strut{}$20$}}\put(682,1754){\makebox(0,0)[r]{\strut{}$30$}}\put(682,2279){\makebox(0,0)[r]{\strut{}$40$}}\put(682,2804){\makebox(0,0)[r]{\strut{}$50$}}\put(682,3329){\makebox(0,0)[r]{\strut{}$60$}}\put(682,3854){\makebox(0,0)[r]{\strut{}$70$}}\put(682,4379){\makebox(0,0)[r]{\strut{}$80$}}\put(814,484){\makebox(0,0){\strut{}$1$}}\put(1384,484){\makebox(0,0){\strut{}$3$}}\put(1955,484){\makebox(0,0){\strut{}$5$}}\put(2525,484){\makebox(0,0){\strut{}$7$}}\put(3096,484){\makebox(0,0){\strut{}$9$}}\put(3666,484){\makebox(0,0){\strut{}$11$}}\put(4236,484){\makebox(0,0){\strut{}$13$}}\put(4807,484){\makebox(0,0){\strut{}$15$}}\put(5377,484){\makebox(0,0){\strut{}$17$}}\put(5947,484){\makebox(0,0){\strut{}$19$}}\put(6518,484){\makebox(0,0){\strut{}$21$}}}\gplgaddtomacro\gplfronttext{\csname LTb\endcsname \put(209,2541){\rotatebox{-270}{\makebox(0,0){\strut{}\large CPU user time (s)}}}\put(3808,154){\makebox(0,0){\strut{}\large Number of models \nbasemodel}}\put(1932,4867){\makebox(0,0)[r]{\strut{}$\vert\dataset\vert$$=$$10\%$}}\put(1932,4647){\makebox(0,0)[r]{\strut{}$\vert\dataset\vert$$=$$25\%$}}\put(3975,4867){\makebox(0,0)[r]{\strut{}$\vert\dataset\vert$$=$$50\%$}}\put(3975,4647){\makebox(0,0)[r]{\strut{}$\vert\dataset\vert$$=$$75\%$}}\put(6018,4867){\makebox(0,0)[r]{\strut{}$\vert\dataset\vert$$=$$100\%$}}}\gplbacktext
    \put(0,0){\includegraphics[width={360.00bp},height={252.00bp}]{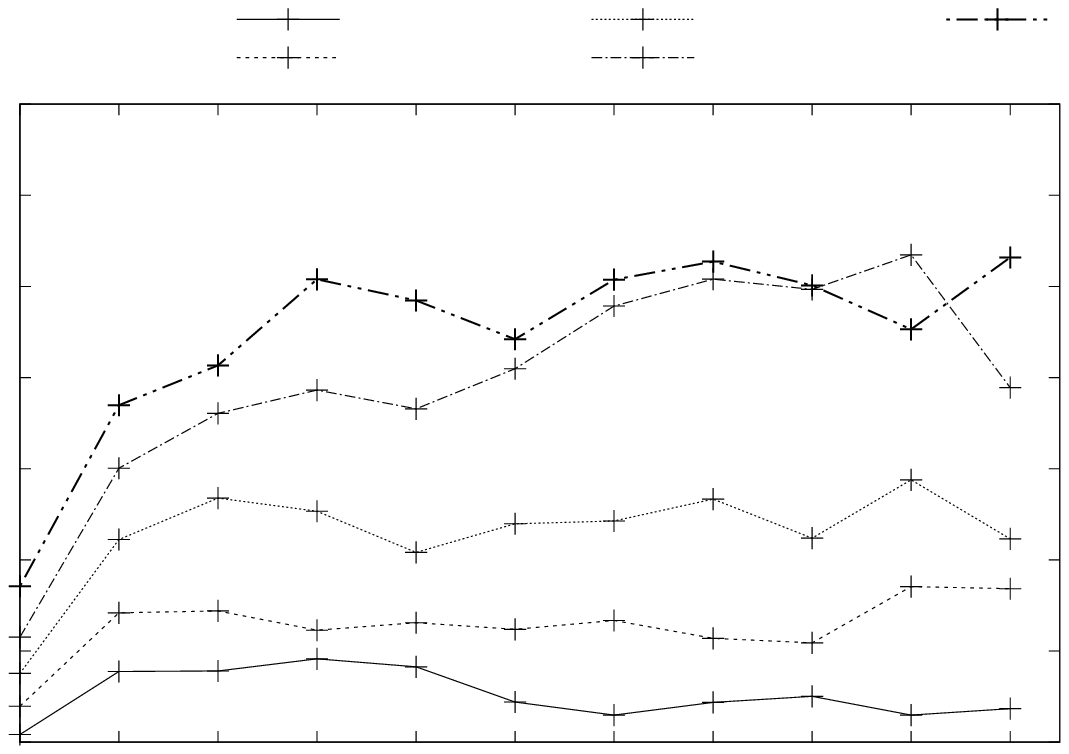}}\gplfronttext
  \end{picture}\endgroup
         }
        &
        \resizebox*{.49\textwidth}{!}{
            \begingroup
  \makeatletter
  \providecommand\color[2][]{\GenericError{(gnuplot) \space\space\space\@spaces}{Package color not loaded in conjunction with
      terminal option `colourtext'}{See the gnuplot documentation for explanation.}{Either use 'blacktext' in gnuplot or load the package
      color.sty in LaTeX.}\renewcommand\color[2][]{}}\providecommand\includegraphics[2][]{\GenericError{(gnuplot) \space\space\space\@spaces}{Package graphicx or graphics not loaded}{See the gnuplot documentation for explanation.}{The gnuplot epslatex terminal needs graphicx.sty or graphics.sty.}\renewcommand\includegraphics[2][]{}}\providecommand\rotatebox[2]{#2}\@ifundefined{ifGPcolor}{\newif\ifGPcolor
    \GPcolorfalse
  }{}\@ifundefined{ifGPblacktext}{\newif\ifGPblacktext
    \GPblacktexttrue
  }{}\let\gplgaddtomacro\g@addto@macro
\gdef\gplbacktext{}\gdef\gplfronttext{}\makeatother
  \ifGPblacktext
\def\colorrgb#1{}\def\colorgray#1{}\else
\ifGPcolor
      \def\colorrgb#1{\color[rgb]{#1}}\def\colorgray#1{\color[gray]{#1}}\expandafter\def\csname LTw\endcsname{\color{white}}\expandafter\def\csname LTb\endcsname{\color{black}}\expandafter\def\csname LTa\endcsname{\color{black}}\expandafter\def\csname LT0\endcsname{\color[rgb]{1,0,0}}\expandafter\def\csname LT1\endcsname{\color[rgb]{0,1,0}}\expandafter\def\csname LT2\endcsname{\color[rgb]{0,0,1}}\expandafter\def\csname LT3\endcsname{\color[rgb]{1,0,1}}\expandafter\def\csname LT4\endcsname{\color[rgb]{0,1,1}}\expandafter\def\csname LT5\endcsname{\color[rgb]{1,1,0}}\expandafter\def\csname LT6\endcsname{\color[rgb]{0,0,0}}\expandafter\def\csname LT7\endcsname{\color[rgb]{1,0.3,0}}\expandafter\def\csname LT8\endcsname{\color[rgb]{0.5,0.5,0.5}}\else
\def\colorrgb#1{\color{black}}\def\colorgray#1{\color[gray]{#1}}\expandafter\def\csname LTw\endcsname{\color{white}}\expandafter\def\csname LTb\endcsname{\color{black}}\expandafter\def\csname LTa\endcsname{\color{black}}\expandafter\def\csname LT0\endcsname{\color{black}}\expandafter\def\csname LT1\endcsname{\color{black}}\expandafter\def\csname LT2\endcsname{\color{black}}\expandafter\def\csname LT3\endcsname{\color{black}}\expandafter\def\csname LT4\endcsname{\color{black}}\expandafter\def\csname LT5\endcsname{\color{black}}\expandafter\def\csname LT6\endcsname{\color{black}}\expandafter\def\csname LT7\endcsname{\color{black}}\expandafter\def\csname LT8\endcsname{\color{black}}\fi
  \fi
    \setlength{\unitlength}{0.0500bp}\ifx\gptboxheight\undefined \newlength{\gptboxheight}\newlength{\gptboxwidth}\newsavebox{\gptboxtext}\fi \setlength{\fboxrule}{0.5pt}\setlength{\fboxsep}{1pt}\definecolor{tbcol}{rgb}{1,1,1}\begin{picture}(7200.00,5040.00)\gplgaddtomacro\gplbacktext{\csname LTb\endcsname \put(946,704){\makebox(0,0)[r]{\strut{}$0$}}\put(946,1112){\makebox(0,0)[r]{\strut{}$500$}}\put(946,1521){\makebox(0,0)[r]{\strut{}$1000$}}\put(946,1929){\makebox(0,0)[r]{\strut{}$1500$}}\put(946,2337){\makebox(0,0)[r]{\strut{}$2000$}}\put(946,2746){\makebox(0,0)[r]{\strut{}$2500$}}\put(946,3154){\makebox(0,0)[r]{\strut{}$3000$}}\put(946,3562){\makebox(0,0)[r]{\strut{}$3500$}}\put(946,3971){\makebox(0,0)[r]{\strut{}$4000$}}\put(946,4379){\makebox(0,0)[r]{\strut{}$4500$}}\put(1078,484){\makebox(0,0){\strut{}$1$}}\put(1623,484){\makebox(0,0){\strut{}$3$}}\put(2168,484){\makebox(0,0){\strut{}$5$}}\put(2714,484){\makebox(0,0){\strut{}$7$}}\put(3259,484){\makebox(0,0){\strut{}$9$}}\put(3804,484){\makebox(0,0){\strut{}$11$}}\put(4349,484){\makebox(0,0){\strut{}$13$}}\put(4895,484){\makebox(0,0){\strut{}$15$}}\put(5440,484){\makebox(0,0){\strut{}$17$}}\put(5985,484){\makebox(0,0){\strut{}$19$}}\put(6530,484){\makebox(0,0){\strut{}$21$}}}\gplgaddtomacro\gplfronttext{\csname LTb\endcsname \put(209,2541){\rotatebox{-270}{\makebox(0,0){\strut{}\large Max memory (MB)}}}\put(3940,154){\makebox(0,0){\strut{}\large Number of models \nbasemodel}}\put(2196,4867){\makebox(0,0)[r]{\strut{}$\vert\dataset\vert$$=$$10$}}\put(2196,4647){\makebox(0,0)[r]{\strut{}$\vert\dataset\vert$$=$$25$}}\put(3975,4867){\makebox(0,0)[r]{\strut{}$\vert\dataset\vert$$=$$50$}}\put(3975,4647){\makebox(0,0)[r]{\strut{}$\vert\dataset\vert$$=$$75$}}\put(5754,4867){\makebox(0,0)[r]{\strut{}$\vert\dataset\vert$$=$$100$}}}\gplbacktext
    \put(0,0){\includegraphics[width={360.00bp},height={252.00bp}]{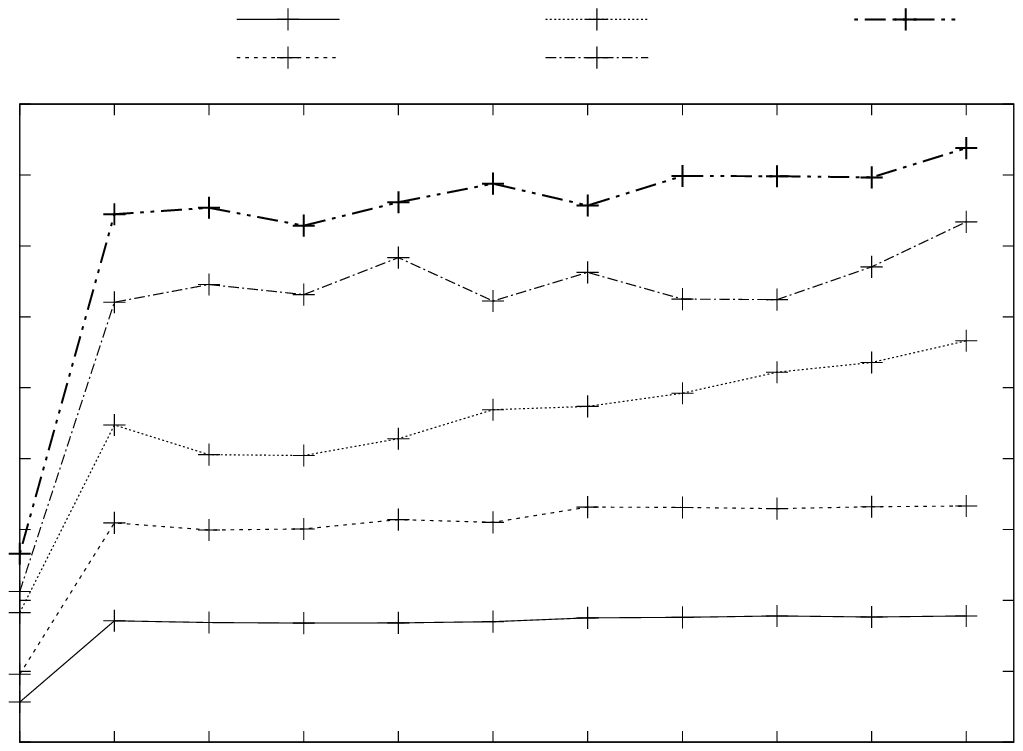}}\gplfronttext
  \end{picture}\endgroup
         }\vspace{.2em}\\
\makecell{\footnotesize (e) CPU user time varying \nbasemodel, with \dataPointsCount$\in$$\{10,$ $25,$ $50,$ $75,$ $100\}$\\and \featuresCount=$100$} & \makecell{\footnotesize (f) Max memory usage varying \nbasemodel, with \dataPointsCount$\in$$\{10,$ $25,$ $50,$ $75,$ $100\}$\\and \featuresCount=$100$}\\
\end{tabular}
\end{adjustbox}
    \caption{\label{fig:sustainability-all} CPU user time and maximum memory on dataset AM, with 14,508 data points and 1,000 features, varying the number of models (\nbasemodel), percentage of data points (\dataPointsCount), and percentage of features (\featuresCount).}
\end{figure*}
 
We measured the resource consumption in terms of CPU and RAM usage during activity training (i.e., steps 3--4 in Section~\ref{subsec:exp-process-benchmark-process}), by wrapping the execution of each individual experimental setting in Section~\ref{subsec:exp-resource-setting} with Linux tool \texttt{time}.\footnote{\url{https://www.man7.org/linux/man-pages/man1/time.1.html}} We measured the CPU user time and the maximum amount of allocated memory; we executed each individual setting $5$ times averaging its results. We note that the resource consumption for activity testing and evaluation (i.e., steps 5--6 in Section~\ref{subsec:exp-process-benchmark-process}), including the execution of our approach on a single data point in the test set and the calculation of $\deltaf$, is negligible.

\vspace{.5em}

\noindent \textbf{CPU usage.} Figures~\ref{fig:sustainability-all}(a), (c), and (e) show the CPU user time. We note that CPU user time is the sum of the time each CPU core spent within the process in the user space. Having implemented a parallel approach, CPU user time is higher than real execution time in Section~\ref{subsubsec:exp-resource-time}.
CPU user time is affected mostly by the dimensions (data points and features) of the dataset, growing at worst linearly as the percentage of data points and features increases, varying between $\approx$$9$s and $\approx$$71$s.
Figure~\ref{fig:sustainability-all}(e) shows that our ensemble approach increases the CPU user time with respect to the monolithic model, from $\approx$$25$s to $\approx$$45$s in the worst case with \nbasemodel$=$$3$. CPU user time remains however stable when \nbasemodel\ increases, showing that the size of the ensemble and therefore its protection does not substantially affect its sustainability.

\vspace{.5em}

\noindent \textbf{Memory usage.} Figures~\ref{fig:sustainability-all}(b), (d), and (f) show the maximum amount of allocated memory for the process. They exhibit similar patterns to the ones of CPU user time, being affected mostly by the dimensions of the dataset and growing at worst linearly with them. It varies between $\approx$$204$MB and $\approx$$4.2$GB.
The same is true also for memory consumption varying \nbasemodel, with a significant increase from monolithic model to our ensemble approach, from $\approx$$1$GB to $\approx$$3.75$GB in the worst case. Memory consumption remains however stable when \nbasemodel\ increases.

\vspace{.5em}

In summary, consumed resources are well below the available resources. Despite our ensemble approach introduces an additional overhead with respect to the monolithic model, it can be easily optimized thanks to its vertical and horizontal scalability.

 \section{Discussion}\label{sec:discussion}

Our main research question focused on investigating the behavior of random forests against poisoning attacks in a scenario where resources are limited on both sides. We evaluated the monolithic model (i.e., ensemble of decision trees) and our ensemble approach (i.e., ensemble of random forests) varying the type (i.e., perturbations) and impact (i.e., \extentpoint\ and \extentfeature) of poisoning. Our main findings are as follows.

\begin{description}
    \item[F1:]\textbf{Monolithic model is highly sensitive to label flipping only.}
    Finding F1 can be observed in Figure~\ref{fig:flipping-overall}, where the monolithic models (\nbasemodel$=$1) are significantly worse than our ensemble approach.
We also note that \deltaf\ of the monolithic models is $-9.225$ on average under flipping, while it is $-0.503$ 
    on average under \zeroing, \noising\ and \outofranging, with a difference of $94.56\%$.
In addition, the accuracy decrease caused by label flipping is proportional with the percentage \extentpoint\ of poisoned data points. This can be noted by comparing \accuracypoisoned\ and \deltaf\ on the datasets in Tables~\ref{tab:deltaFlip}(a)--(d), where \accuracypoisoned\ and \deltaf\ progressively worsen as \extentpoint\ increases.

    \item[F2:]\textbf{The effectiveness of perturbations depends also on the characteristics of the dataset and the size of the ensemble.}
Finding F2 can be observed for label flipping by comparing downward the right\-/hand side of Tables~\ref{tab:deltaFlip}(a)--(d).
Being M2 and DR smaller than AM and SB in terms of cardinality and sparsity, \deltaf\ worsens more rapidly as the percentage of poisoned data points increases. In addition, \deltaf\ also has a smaller improvement on M2 and DR as \nbasemodel\ increases,  because the cardinality of the individual partitions and training sets is increasingly reduced.
Finding F2 can be observed for other perturbations by comparing the trends in Figure~\ref{fig:other-overall}. In this case, accuracy worsens at a higher rate in M2 and SB (smaller than AM) as \nbasemodel\ increases, with \outofranging\ being the most effective perturbation.
This worsening trend can be observed also in DR despite its \accuracyclean\ being significantly lower.

    \item[F3:]\textbf{Ensemble of random forests is an adequate protection from untargeted label flipping.} Finding F3 can be observed by comparing the increase of \deltaf\ of our ensemble approach with regards to the monolithic model in Tables~\ref{tab:deltaFlip}(a)--(d). 
For instance, considering dataset AM, accuracy becomes $>$$96$ with at least $\nbasemodel$$=$9 random forests in our ensemble approach for $\extentpoint$$\le$$30$ of poisoned data points, and keeps increasing slightly with \nbasemodel.
Instead, for $\extentpoint$$>$$30$  of poisoned data points, our approach starts suffering of not\-/negligible accuracy decreases, being $<$$96$ in virtually all cases. We note that, for \nbasemodel$>$$15$, this decrease can be still considered negligible, being accuracy $\ge$$95$.
In general, the improvement provided by our ensemble approach is significant, as summarized in Figure~\ref{fig:flipping-overall}, where the highest lines corresponding to the monolithic model are always significantly worse than those of our ensemble approach.

    \item[F4:]\textbf{Ensemble of decision trees is an adequate protection from untargeted perturbations zeroing, noising, out-of-ranging.} Finding F4 is a direct consequence of F1 and can be observed by comparing the accuracy of monolithic and ensemble models in Figure~\ref{fig:other-overall}. Perturbations \zeroing, \noising, and \outofranging\ introduce a minor accuracy decrease, which largely fails to make the monolithic model unusable in practice. Accuracy variation of the poisoned monolithic model with regards to the original accuracy is in fact always less than 4.784.
This implies that our ensemble approach is redundant in this scenario, and explains the accuracy decrease we observed as \nbasemodel\ increases. In practice, our approach only reduces the cardinality of the training set of each base model from $\vert D \vert$ to $\vert D \vert/\nbasemodel$, hence affecting classification accuracy. This tradeoff is advantageous in label flipping, when the accuracy decrease caused by the smaller training set is balanced by containing the accuracy decrease caused by poisoning. It is detrimental for other perturbations where the accuracy decrease caused by poisoning is negligible.

    \item[F5:]\textbf{Our ensemble approach is sustainable.} Finding F5 can be observed by analyzing the growth of resources consumption when  \nbasemodel\ increases. Consumed resources (CPU and RAM) stay mostly stable and are affected by the dimensions (number of data points and features) of the dataset.
In addition, our ensemble largely benefits from parallelism. This is noticeable by comparing CPU user time in Figures~\ref{fig:sustainability-all}(a), (c), and (e) with real execution time in Figures~\ref{fig:exp-resource-time-varying-n}, \ref{fig:exp-resource-time-varying-cardinality}, and~\ref{fig:exp-resource-time-varying-features}: high CPU user time still corresponds to low execution time.
This means that size and therefore robustness of our ensemble approach can be tuned according to the scenario, incurring in a \emph{constant} overhead. \end{description}

\vspace{0.5em}

From the above findings, we can conclude that random forest (a native ensemble algorithm) provides an empirically\-/strong robustness against \zeroing, \noising, \outofranging\ attacks. When label flipping is considered, an ensemble of random forests is needed to ensure robustness. The size of the ensemble must however be carefully balanced to avoid accuracy decrease due to an oversized ensemble approach.
In short, even untargeted poisoning attacks requiring little to none knowledge and resources on the attacker side can be dangerous and untractable with dataset strengthening defenses~\cite{paudice2018detection}; these attacks can be rather counteracted with a sustainable model strengthening defense.

Finally, we note that the results in this paper, although novel, are in line with other work on ML robustness, for instance~\cite{su12166434, jia2021intrinsic}, claiming that \emph{i)} if the amount of poisoning is reasonable, an ensemble strategy can reduce the influence of poisoned data points to the resulting model, and \emph{ii)} random forests are, in some cases, more robust than other models (e.g., naive bayes, neural networks)~\cite{su12166434, Zhang2021, YERLIKAYA2022118101}.
 \section{Comparison with the State of the Art}\label{sec:comparison}
Random forest, being one of the most popular algorithms for tabular datasets, has been studied from different angles including:
\begin{enumerate*}
    \item explainability~\cite{10.1145/2339530.2339556,10.1145/2939672.2939778,chipmanmaking,Deng2019,FERNANDEZ2020196,Gossen2021}; 
    \item fairness~\cite{10.1145/3540250.3549093,10.1007/978-3-030-75765-6_20,9462068,https://doi.org/10.48550/arxiv.2212.04593};
    \item sustainability~\cite{doi:10.1137/1.9781611975673.6, 9923840,doi:10.1142/S0218213009000305,ORBi-b468d9da-a210-4a15-9308-3e4b320b5675,Painsky2019};
    \item robustness~\cite{su12166434, Taheri2020,Zhang2021, VERDE20212624, 9652959, 10.1145/3468218.3469050, YERLIKAYA2022118101, https://doi.org/10.48550/arxiv.2208.08433, 10020528}.
\end{enumerate*}

\begin{table}[!t]
    \caption{Comparison with related work on data poisoning attacks and defenses against random forest. Threat model type (\emph{Type}): untargeted \emph{\typeUnTargeted} or targeted \emph{\typeTargeted}; selection strategy (\emph{Select.}): random \emph{\strategyRandom} or with a specific strategy \emph{\strategyNonRandom}; perturbation strategy (\emph{Pert.}): (random \emph{\strategyRandom} or with a specific strategy \emph{\strategyNonRandom}).}
    \label{tbl:related-work}
    \begin{adjustbox}{max totalsize={\linewidth}{\textheight},center}
        \begin{tabular}{l | c c | c c | c c c | l}
            \hline
            \multirow{2}{*}{\textbf{Ref.}} &
                \multicolumn{2}{c|}{\textbf{Poisoning type}} &
                \multicolumn{2}{c|}{\textbf{Defense type}} &
                \multicolumn{3}{c|}{\textbf{Threat model}}&
            \multirow{2}{*}{\textbf{Domain}} \\
            & \textbf{Features} & \textbf{Labels} & \textbf{Dataset} & \textbf{Model} &  \textbf{Type} & \textbf{Select.} & \textbf{Pert.} & \\
            \hline
            \cite{su12166434} & \notok & \ok & \notok & \notok & \typeUnTargeted & \strategyRandom & \strategyRandom & \domainTypeSingle\\ 
            \cite{Taheri2020} & \notok & \ok & \ok & \notok & \typeUnTargeted & \strategyNonRandom & \strategyNonRandom & \domainTypeSingle\\ 
            \cite{Zhang2021} & \notok & \ok & \notok & \notok & \typeTargeted & \strategyNonRandom & \strategyNonRandom & \domainTypeSingle\\
            \cite{VERDE20212624} & \ok & \notok & \notok & \notok & \typeUnTargeted & \strategyRandom & \strategyNonRandom & \domainTypeSingle\\
            \cite{9652959} & \ok & \notok & \notok & \notok & \typeUnTargeted & \strategyRandom & \strategyNonRandom & \domainTypeSingle\\
            \cite{10.1145/3468218.3469050} & \ok & \notok & \notok & \notok & \typeUnTargeted & \strategyRandom & \strategyNonRandom & \domainTypeSingle\\
            \cite{YERLIKAYA2022118101} & \notok & \ok & \notok & \notok & \typeUnTargeted & \strategyRandom,\strategyNonRandom & \strategyRandom,\strategyNonRandom & \domainTypeMulti \\
            \cite{https://doi.org/10.48550/arxiv.2208.08433} & \notok & \ok & \ok & \notok & \typeUnTargeted & \strategyNonRandom & \strategyNonRandom & \domainTypeSingle\\
            \cite{10020528} & \notok & \ok & \notok & \notok & \typeUnTargeted & \strategyRandom & \strategyRandom & \domainTypeSingle\\
            This & \ok & \ok & \notok & \ok & \typeUnTargeted & \strategyRandom & \strategyNonRandom & \domainTypeMulti \\ 
            \hline
\end{tabular}
    \end{adjustbox}
\end{table} 
In terms of robustness different solutions have been defined on random forest, though none of them can be easily experimentally compared against the ensemble approach in this paper. Table~\ref{tbl:related-work} shows how these solutions compare with our approach in terms of \emph{Poisoning type}, the portion of the data points (features or label) affected by poisoning; \emph{Defense type}, the type of defense, either dataset or model strengthening; \emph{threat model}, the attacker capabilities. 
The latter is composed of three main dimensions: \begin{enumerate*}
        \item the threat model type (untargeted \emph{\typeUnTargeted} or targeted \emph{\typeTargeted}), denoted as \emph{Type};
        \item the strategy used to select the data points to be poisoned (random \emph{\strategyRandom} or with a specific strategy \emph{\strategyNonRandom}), denoted as \emph{Select.}; and
        \item the strategy used to poison the selected data points (random \emph{\strategyRandom} or with a specific strategy \emph{\strategyNonRandom}), denoted as \emph{Pert.}
    \end{enumerate*}

Most of the surveyed approaches have considered poisoning attacks (no defenses) against standard random forests in a single domain~\cite{su12166434,Zhang2021,VERDE20212624,9652959,10.1145/3468218.3469050,10020528}, with few papers evaluating different datasets from different domains~\cite{YERLIKAYA2022118101}. In addition, most evaluations consider attacks affecting labels only~\cite{su12166434,Taheri2020,Zhang2021,YERLIKAYA2022118101,https://doi.org/10.48550/arxiv.2208.08433,10020528}.
Few defenses have been proposed and evaluated, whose threat model implements label flipping attack against data points selected according to a specific strategy.
Taheri et al.~\cite{Taheri2020} presented two dataset strengthening defenses in the domain of Android malware detection, based on healing suspicious data points according to label propagation and clustering.
Shahid et al.~\cite{https://doi.org/10.48550/arxiv.2208.08433} proposed a dataset strengthening defense in the domain of human activity recognition from sensors data. The latter is based on the approach by Paudice et al.~\cite{paudice2018label}, where a clustering model trained on a trusted dataset is used to heal suspicious data points.
We also note that there exist some approaches where random forest is part of the defense such as in~\cite{Ding_2019_CVPR_Workshops,Taheri2020}.

The ensemble defense in this paper (last row in Table~\ref{tbl:related-work}) departed from assumptions and configurations in the state of the art, making a comparative experimental evaluation meaningless. First, our threat model considers untargeted attacks, where data points are selected according to a random strategy and poisoning affects both labels and features. Second, our approach strengthens the ML model rather than the training set. Third, our approach focused on significantly different domains and datasets, rather than on a single domain with one~\cite{https://doi.org/10.48550/arxiv.2208.08433} or more~\cite{Taheri2020} datasets.
 \section{Conclusions}\label{sec:conclusion}
Machine learning models play an increasingly vital role in the digital services we interact with. As a consequence, the need for properly securing such models from attacks is a key issue being investigated by the research community. 
This paper aimed to shed new light on the usage of ensembles as a means of protecting random forests against poisoning attacks. While ensembles have been already proposed in the context of certified protection in the domain of image recognition, little has been done in the context of random forests.
Throughout fine\-/grained experiments, we show that label flipping, even if performed with no strategy, is a very dangerous type of perturbation, significantly degrading the performance of plain random forests. A simple yet effective and sustainable countermeasure consists in training models on disjoint training sets, then aggregating their predictions with majority voting. Other perturbations are less effective, and random forest is already an effective countermeasure.
The paper leaves space for future work. First, we plan to enrich our set of perturbations with targeted attacks, including backdoor poisoning. Second, we plan to fine\-/tune the hyperparameters of random forest to find relevant trends with respect to the considered threat model and datasets. Third, we plan to develop a complete benchmark considering the new perturbations and different hash function.

\ifCLASSOPTIONcompsoc
\section*{Acknowledgments}
  Research supported, in parts, by \emph{i)} Technology Innovation Institute (TII) under Grant 8434000394, \emph{ii)} the project MUSA - Multilayered Urban Sustainability Action - project, funded by the European Union - NextGenerationEU, under the National Recovery and Resilience Plan (NRRP) Mission 4 Component 2 Investment Line 1.5: Strengthening of research structures and creation of R\&D ``innovation ecosystems'', set up of ``territorial leaders in R\&D'' (CUP  G43C22001370007, Code ECS00000037), and \emph{iii)} the project SERICS (PE00000014) under the NRRP MUR program funded by the EU - NextGenerationEU.
\else
\section*{Acknowledgment}
  Research supported, in parts, by \emph{i)} Technology Innovation Institute (TII) under Grant 8434000394, \emph{ii)} the project MUSA - Multilayered Urban Sustainability Action - project, funded by the European Union - NextGenerationEU, under the National Recovery and Resilience Plan (NRRP) Mission 4 Component 2 Investment Line 1.5: Strengthening of research structures and creation of R\&D ``innovation ecosystems'', set up of ``territorial leaders in R\&D'' (CUP  G43C22001370007, Code ECS00000037), and \emph{iii)} the project SERICS (PE00000014) under the NRRP MUR program funded by the EU - NextGenerationEU.
\fi

\bibliographystyle{IEEEtran}
\bibliography{bibliography}

\begin{thebibliography}{10}
\providecommand{\url}[1]{#1}
\csname url@samestyle\endcsname
\providecommand{\newblock}{\relax}
\providecommand{\bibinfo}[2]{#2}
\providecommand{\BIBentrySTDinterwordspacing}{\spaceskip=0pt\relax}
\providecommand{\BIBentryALTinterwordstretchfactor}{4}
\providecommand{\BIBentryALTinterwordspacing}{\spaceskip=\fontdimen2\font plus
\BIBentryALTinterwordstretchfactor\fontdimen3\font minus \fontdimen4\font\relax}
\providecommand{\BIBforeignlanguage}[2]{{%
\expandafter\ifx\csname l@#1\endcsname\relax
\typeout{** WARNING: IEEEtran.bst: No hyphenation pattern has been}%
\typeout{** loaded for the language `#1'. Using the pattern for}%
\typeout{** the default language instead.}%
\else
\language=\csname l@#1\endcsname
\fi
#2}}
\providecommand{\BIBdecl}{\relax}
\BIBdecl

\bibitem{app9245574}
F.~Rundo, F.~Trenta, A.~L. di~Stallo, and S.~Battiato, ``{Machine Learning for Quantitative Finance Applications: A Survey},'' \emph{Applied Sciences}, vol.~9, no.~24, 2019.

\bibitem{8625421}
E.~Hossain, I.~Khan, F.~Un-Noor, S.~S. Sikander, and M.~S.~H. Sunny, ``{Application of Big Data and Machine Learning in Smart Grid, and Associated Security Concerns: A Review},'' \emph{IEEE Access}, vol.~7, pp. 13\,960--13\,988, 2019.

\bibitem{atmos11070676}
R.~Chen, W.~Zhang, and X.~Wang, ``{Machine Learning in Tropical Cyclone Forecast Modeling: A Review},'' \emph{Atmosphere}, vol.~11, no.~7, 2020.

\bibitem{mio2019signal}
C.~Mio and G.~Gianini, ``Signal reconstruction by means of embedding, clustering and autoencoder ensembles,'' in \emph{Proc. of IEEE ISCC 2019}, Barcelona, Spain, July 2019.

\bibitem{KOUROU20158}
K.~Kourou, T.~P. Exarchos, K.~P. Exarchos, M.~V. Karamouzis, and D.~I. Fotiadis, ``Machine learning applications in cancer prognosis and prediction,'' \emph{Computational and Structural Biotechnology Journal}, vol.~13, 2015.

\bibitem{KatraAnisetti22}
P.~Katrakazas, A.~Ballas, M.~Anisetti, and I.~Spais, ``An artificial intelligence outlook for colorectal cancer screening,'' in \emph{Proc. of IEEE BigDataService 2022}, San Francisco, CA, USA, August 2022.

\bibitem{Richens2020}
J.~G. Richens, C.~M. Lee, and S.~Johri, ``Improving the accuracy of medical diagnosis with causal machine learning,'' \emph{Nature Communications}, vol.~11, no.~1, Aug 2020.

\bibitem{sma2020rf}
J.~Y. Chang and E.~G. Im, ``{Data Poisoning Attack on Random Forest Classification Model},'' in \emph{Proc. of SMA 2020}, Ramada Plaza Jeju, Jeju, Republic of Korea, September 2020.

\bibitem{chen2017targeted}
X.~Chen, C.~Liu, B.~Li, K.~Lu, and D.~Song, ``{Targeted Backdoor Attacks on Deep Learning Systems Using Data Poisoning},'' \emph{arXiv preprint arXiv:1712.05526}, 2017.

\bibitem{6868201}
M.~Mozaffari-Kermani, S.~Sur-Kolay, A.~Raghunathan, and N.~K. Jha, ``{Systematic Poisoning Attacks on and Defenses for Machine Learning in Healthcare},'' \emph{IEEE Journal of Biomedical and Health Informatics}, vol.~19, no.~6, 2015.

\bibitem{263874}
R.~Schuster, C.~Song, E.~Tromer, and V.~Shmatikov, ``{You Autocomplete Me: Poisoning Vulnerabilities in Neural Code Completion},'' in \emph{Proc. of USENIX Security 2021}, Virtual, Aug. 2021.

\bibitem{pmlr-v20-biggio11}
B.~Biggio, B.~Nelson, and P.~Laskov, ``{Support Vector Machines Under Adversarial Label Noise},'' in \emph{Proc. of ACML 2011}, Taoyuan, Taiwan, November 2011.

\bibitem{su12166434}
C.~Dunn, N.~Moustafa, and B.~Turnbull, ``{Robustness Evaluations of Sustainable Machine Learning Models against Data Poisoning Attacks in the Internet of Things},'' \emph{Sustainability}, vol.~12, no.~16, 2020.

\bibitem{8489495}
C.~Frederickson, M.~Moore, G.~Dawson, and R.~Polikar, ``{Attack Strength vs. Detectability Dilemma in Adversarial Machine Learning},'' in \emph{Proc. of IJCNN 2018}, Rio de Janeiro, Brazil, July 2018.

\bibitem{pmlr-v97-diakonikolas19a}
I.~Diakonikolas, G.~Kamath, D.~Kane, J.~Li, J.~Steinhardt, and A.~Stewart, ``{Sever: A Robust Meta-Algorithm for Stochastic Optimization},'' in \emph{Proc. of ICML 2019}, Long Beach, CA, USA, June 2019.

\bibitem{10.5555/3367471.3367701}
Y.~Ma, X.~Zhu, and J.~Hsu, ``{Data Poisoning against Differentially-Private Learners: Attacks and Defenses},'' in \emph{Proc. of IJCAI 2019}, Macao, China, August 2019.

\bibitem{pmlr-v119-rosenfeld20b}
E.~Rosenfeld, E.~Winston, P.~Ravikumar, and Z.~Kolter, ``{Certified Robustness to Label-Flipping Attacks via Randomized Smoothing},'' in \emph{Proc. of ICML 2020}, Virtual, June 2020.

\bibitem{10.1007/978-3-030-66415-2_4}
N.~Peri, N.~Gupta, W.~R. Huang, L.~Fowl, C.~Zhu, S.~Feizi, T.~Goldstein, and J.~P. Dickerson, ``{Deep k-NN Defense Against Clean-Label Data Poisoning Attacks},'' in \emph{Proc. of ECCV 2020}, August 2020.

\bibitem{Koh2021}
P.~W. Koh, J.~Steinhardt, and P.~Liang, ``Stronger data poisoning attacks break data sanitization defenses,'' \emph{Machine Learning}, vol. 111, no.~1, Nov 2021.

\bibitem{jia2021intrinsic}
J.~Jia, X.~Cao, and N.~Z. Gong, ``{Intrinsic Certified Robustness of Bagging against Data Poisoning Attacks},'' in \emph{Proc. of AAAI 2021}, Virtual, February 2021.

\bibitem{levine2021deep}
A.~Levine and S.~Feizi, ``{Deep Partition Aggregation: Provable Defenses against General Poisoning Attacks},'' in \emph{Proc. of ICLR 2021}, Vienna, Austria, May 2021.

\bibitem{wang2022improved}
W.~Wang, A.~Levine, and S.~Feizi, ``{Improved Certified Defenses against Data Poisoning with (Deterministic) Finite Aggregation},'' in \emph{Proc. of ICML 2022}, Baltimore, MD, USA, July 2022.

\bibitem{pmlr-v162-chen22k}
R.~Chen, Z.~Li, J.~Li, J.~Yan, and C.~Wu, ``{On Collective Robustness of Bagging Against Data Poisoning},'' in \emph{Proc. of ICML 2022}, Baltimore, MD, USA, July 2022.

\bibitem{why}
L.~Grinsztajn and G.~V. Edouard~Oyallon, ``Why do tree-based models still outperform deep learning on typical tabular data?'' in \emph{Proc. of NeurIPS 2022}, New Orleans, LA, USA, November--December 2022.

\bibitem{10.1145/2939672.2939778}
M.~T. Ribeiro, S.~Singh, and C.~Guestrin, ``"why should i trust you?": Explaining the predictions of any classifier,'' in \emph{Proc. of ACM KDD 2016}, San Francisco, CA, USA, August 2016.

\bibitem{10.1145/3540250.3549093}
Z.~Chen, J.~M. Zhang, F.~Sarro, and M.~Harman, ``Maat: A novel ensemble approach to addressing fairness and performance bugs for machine learning software,'' in \emph{Proc. of ACM ESEC/FSE 2022}, Singapore, Singapore, November 2022.

\bibitem{9923840}
A.~Prasad, S.~Rajendra, K.~Rajan, R.~Govindarajan, and U.~Bondhugula, ``Treebeard: An optimizing compiler for decision tree based ml inference,'' in \emph{Proc. of IEEE/ACM MICRO 2022}, Chicago, IL, USA, October 2022.

\bibitem{Zhang2021}
H.~Zhang, N.~Cheng, Y.~Zhang, and Z.~Li, ``Label flipping attacks against naive bayes on spam filtering systems,'' \emph{Applied Intelligence}, vol.~51, no.~7, Jul 2021.

\bibitem{VERDE20212624}
L.~Verde, F.~Marulli, and S.~Marrone, ``Exploring the impact of data poisoning attacks on machine learning model reliability,'' \emph{Procedia Computer Science}, vol. 192, 2021.

\bibitem{9652959}
K.~Talty, J.~Stockdale, and N.~D. Bastian, ``A sensitivity analysis of poisoning and evasion attacks in network intrusion detection system machine learning models,'' in \emph{Proc. of IEEE MILCOM 2021}, San Diego, CA, USA, November 2021.

\bibitem{10.1145/3468218.3469050}
A.~Prud'Homme and B.~Kantarci, ``{Poisoning Attack Anticipation in Mobile Crowdsensing: A Competitive Learning-Based Study},'' in \emph{Proc. of ACM WiseML 2021}, Abu Dhabi, UAE, June 2021.

\bibitem{YERLIKAYA2022118101}
F.~A. Yerlikaya and Şerif Bahtiyar, ``Data poisoning attacks against machine learning algorithms,'' \emph{Expert Systems with Applications}, vol. 208, 2022.

\bibitem{Taheri2020}
R.~Taheri, R.~Javidan, M.~Shojafar, Z.~Pooranian, A.~Miri, and M.~Conti, ``On defending against label flipping attacks on malware detection systems,'' \emph{Neural Computing and Applications}, vol.~32, no.~18, Sep 2020.

\bibitem{https://doi.org/10.48550/arxiv.2208.08433}
A.~R. Shahid, A.~Imteaj, P.~Y. Wu, D.~A. Igoche, and T.~Alam, ``{Label Flipping Data Poisoning Attack Against Wearable Human Activity Recognition System},'' \emph{arXiv preprint arXiv:2208.08433}, 2022.

\bibitem{damiani2020certified}
E.~Damiani and C.~A. Ardagna, ``{Certified Machine-Learning Models},'' in \emph{Proc. of SOFSEM 2020}, Limassol, Cyprus, January 2020.

\bibitem{anisetti2020methodology}
M.~Anisetti, C.~A. Ardagna, E.~Damiani, and P.~G. Panero, ``{A Methodology for Non-Functional Property Evaluation of Machine Learning Models},'' in \emph{Proc. of MEDES 2020}, Abu Dhabi, UAE, November 2020.

\bibitem{10.1145/3446331}
L.~Mauri and E.~Damiani, ``{Estimating Degradation of Machine Learning Data Assets},'' \emph{Journal of Data and Information Quality}, vol.~14, no.~2, dec 2021.

\bibitem{10.1145/3585385}
A.~E. Cin\`{a}, K.~Grosse, A.~Demontis, S.~Vascon, W.~Zellinger, B.~A. Moser, A.~Oprea, B.~Biggio, M.~Pelillo, and F.~Roli, ``{Wild Patterns Reloaded: A Survey of Machine Learning Security against Training Data Poisoning},'' \emph{ACM CSUR}, 2023.

\bibitem{42503}
C.~Szegedy, W.~Zaremba, I.~Sutskever, J.~Bruna, D.~Erhan, I.~Goodfellow, and R.~Fergus, ``Intriguing properties of neural networks,'' in \emph{Proc. of ICLR 2014}, Banff, Canada, April 2014.

\bibitem{zhang2017understanding}
C.~Zhang, S.~Bengio, M.~Hardt, B.~Recht, and O.~Vinyals, ``{Understanding Deep Learning Requires Rethinking Generalization},'' in \emph{Proc. of ICLR 2017}, Toulon, France, April 2017.

\bibitem{paudice2018label}
A.~Paudice, L.~Mu{\~n}oz-Gonz{\'a}lez, and E.~C. Lupu, ``{Label Sanitization Against Label Flipping Poisoning Attacks},'' in \emph{Proc. of ECML PKDD 2018 Workshops}, Dublin, Ireland, September 2018.

\bibitem{NEURIPS2018_22722a34}
A.~Shafahi, W.~R. Huang, M.~Najibi, O.~Suciu, C.~Studer, T.~Dumitras, and T.~Goldstein, ``{Poison Frogs! Targeted Clean-Label Poisoning Attacks on Neural Networks},'' in \emph{Proc. of NeurIPS 2018}, Montréal, Canada, December 2018.

\bibitem{4531146}
G.~F. Cretu, A.~Stavrou, M.~E. Locasto, S.~J. Stolfo, and A.~D. Keromytis, ``{Casting out Demons: Sanitizing Training Data for Anomaly Sensors},'' in \emph{Proc. of IEEE SP 2008}, Oakland, CA, USA, may 2008.

\bibitem{Barreno2010}
M.~Barreno, B.~Nelson, A.~D. Joseph, and J.~D. Tygar, ``The security of machine learning,'' \emph{Machine Learning}, vol.~81, no.~2, 2010.

\bibitem{prasad2020robust}
A.~Prasad, A.~S. Suggala, S.~Balakrishnan, and P.~Ravikumar, ``Robust estimation via robust gradient estimation,'' \emph{Journal of the Royal Statistical Society: Series B (Statistical Methodology)}, vol.~82, no.~3, 2020.

\bibitem{biggio2013evasion}
B.~Biggio, I.~Corona, D.~Maiorca, B.~Nelson, N.~{\v{S}}rndi{\'c}, P.~Laskov, G.~Giacinto, and F.~Roli, ``{Evasion Attacks Against Machine Learning at Test Time},'' in \emph{Proc. of ECML PKDD 2013}, Prague, Czech Republic, September 2013.

\bibitem{hong2020effectiveness}
S.~Hong, V.~Chandrasekaran, Y.~Kaya, T.~Dumitraş, and N.~Papernot, ``{On the Effectiveness of Mitigating Data Poisoning Attacks with Gradient Shaping},'' \emph{arXiv preprint arXiv:2002.11497}, 2020.

\bibitem{10020528}
K.~Aryal, M.~Gupta, and M.~Abdelsalam, ``{Analysis of Label-Flip Poisoning Attack on Machine Learning Based Malware Detector},'' in \emph{Proc. of IEEE Big Data 2022}, Osaka, Japan, December 2022.

\bibitem{enisaetl2022}
{Eurpean Union Agency for Cybersecurity}, ``{ENISA Threat Landscape 2022},'' Eurpean Union Agency for Cybersecurity, Tech. Rep., 10 2022.

\bibitem{Dua:2019}
\BIBentryALTinterwordspacing
D.~Dua and C.~Graff, ``{UCI} machine learning repository,'' 2017. [Online]. Available: \url{http://archive.ics.uci.edu/ml}
\BIBentrySTDinterwordspacing

\bibitem{Quinlan1986}
J.~R. Quinlan, ``{Induction of Decision Trees},'' \emph{Machine Learning}, vol.~1, no.~1, 1986.

\bibitem{DBLP:journals/ml/HushSS07}
D.~R. Hush, C.~Scovel, and I.~Steinwart, ``Stability of unstable learning algorithms,'' \emph{Machine Learning}, vol.~67, no.~3, 2007.

\bibitem{DBLP:conf/icml/WangW02}
Y.~Wang and I.~H. Witten, ``Modeling for optimal probability prediction,'' in \emph{Proc. of ICML 2002}, Sidney, Australia, July 2002.

\bibitem{DBLP:conf/esann/ChristosS04}
C.~Dimitrakakis and S.~Bengio, ``Online policy adaptation for ensemble classifiers,'' in \emph{Proc. of ESANN 2004}, Bruges, Belgium, April 2004.

\bibitem{DBLP:journals/kbs/AntalH14}
B.~Antal and A.~Hajdu, ``An ensemble-based system for automatic screening of diabetic retinopathy,'' \emph{Knowledge-Based Systems}, vol.~60, 2014.

\bibitem{decenciere2014feedback}
E.~Decenci{\`e}re, X.~Zhang, G.~Cazuguel, B.~Lay, B.~Cochener, C.~Trone, P.~Gain, R.~Ordonez, P.~Massin, A.~Erginay \emph{et~al.}, ``Feedback on a publicly distributed image database: the messidor database,'' \emph{Image Analysis \& Stereology}, vol.~33, no.~3, 2014.

\bibitem{DBLP:journals/sigkdd/HallFHPRW09}
M.~A. Hall, E.~Frank, G.~Holmes, B.~Pfahringer, P.~Reutemann, and I.~H. Witten, ``The {WEKA} data mining software: an update,'' \emph{{SIGKDD} Explorations Newsletter}, vol.~11, no.~1, 2009.

\bibitem{paudice2018detection}
A.~Paudice, L.~Muñoz-González, A.~Gyorgy, and E.~C. Lupu, ``{Detection of Adversarial Training Examples in Poisoning Attacks through Anomaly Detection},'' \emph{arXiv preprint arXiv:1802.03041}, 2018.

\bibitem{10.1145/2339530.2339556}
Y.~Lou, R.~Caruana, and J.~Gehrke, ``Intelligible models for classification and regression,'' in \emph{Proc. of ACM KDD 2012}, Beijing, China, August 2012.

\bibitem{chipmanmaking}
H.~Chipman, E.~George, and McCulloch, ``Making sense of a forest of trees,'' \emph{Computing Science and Statistics}, 1998.

\bibitem{Deng2019}
H.~Deng, ``Interpreting tree ensembles with intrees,'' \emph{JDSA}, vol.~7, no.~4, Jun 2019.

\bibitem{FERNANDEZ2020196}
R.~R. Fernández, I.~{Martín de Diego}, V.~Aceña, A.~Fernández-Isabel, and J.~M. Moguerza, ``Random forest explainability using counterfactual sets,'' \emph{Information Fusion}, vol.~63, 2020.

\bibitem{Gossen2021}
F.~Gossen and B.~Steffen, ``Algebraic aggregation of random forests: towards explainability and rapid evaluation,'' \emph{STT}, 2021.

\bibitem{10.1007/978-3-030-75765-6_20}
W.~Zhang, A.~Bifet, X.~Zhang, J.~C. Weiss, and W.~Nejdl, ``Farf: A fair and adaptive random forests classifier,'' in \emph{Proc. of PAKDD 2021}, Virtual, May 2021.

\bibitem{9462068}
P.~J. Kenfack, A.~M. Khan, S.~A. Kazmi, R.~Hussain, A.~Oracevic, and A.~M. Khattak, ``Impact of model ensemble on the fairness of classifiers in machine learning,'' in \emph{Proc. of ICAPAI 2021}, Halden, Norway, May 2021.

\bibitem{https://doi.org/10.48550/arxiv.2212.04593}
U.~Gohar, S.~Biswas, and H.~Rajan, ``Towards understanding fairness and its composition in ensemble machine learning,'' in \emph{Proc. of IEEE/ACM ICSE 2023}, Melbourne, Australia, May 2023.

\bibitem{doi:10.1137/1.9781611975673.6}
J.~Browne, D.~Mhembere, T.~M. Tomita, J.~T. Vogelstein, and R.~Burns, ``Forest packing: Fast parallel, decision forests,'' in \emph{Proc. of SDM 2019}, Calgary, Canada, May 2019.

\bibitem{doi:10.1142/S0218213009000305}
A.~H. Peterson and T.~R. Martinzed, ``Reducing decision tree ensemble size using parallel decision dags,'' \emph{International Journal on Artificial Intelligence Tools}, vol.~18, no.~04, 2009.

\bibitem{ORBi-b468d9da-a210-4a15-9308-3e4b320b5675}
A.~Joly, F.~Schnitzler, P.~Geurts, and L.~Wehenkel, ``L1-based compression of random forest models,'' in \emph{Proc. of ESANN 2012}, Bruges, Belgium, April 2012.

\bibitem{Painsky2019}
A.~Painsky and S.~Rosset, ``Lossless compression of random forests,'' \emph{Journal of Computer Science and Technology}, vol.~34, no.~2, 2019.

\bibitem{Ding_2019_CVPR_Workshops}
Y.~Ding, L.~Wang, H.~Zhang, J.~Yi, D.~Fan, and B.~Gong, ``{Defending Against Adversarial Attacks Using Random Forest},'' in \emph{Proc. of IEEE/CVF CVPR 2019}, Long Beach, CA, USA, June 2019.

\end{thebibliography}

\vskip -3\baselineskip plus -1fil

\begin{IEEEbiography}[{\includegraphics[width=1in,height=1.25in,clip]{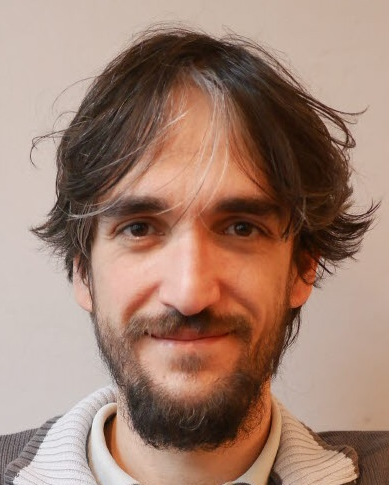}}]{Marco Anisetti} is Full Professor at the Università degli Studi di Milano. His research interests are in the area of computational intelligence, and its application to the design and evaluation of complex systems. He has been investigating innovative solutions in the area of cloud security assurance evaluation. In this area he defined a new scheme for continuous and incremental cloud security certification, based on distributed assurance evaluation architecture.
\end{IEEEbiography}

\vskip -3\baselineskip plus -1fil

\begin{IEEEbiography}[{\includegraphics[width=1in,height=1.25in,clip]{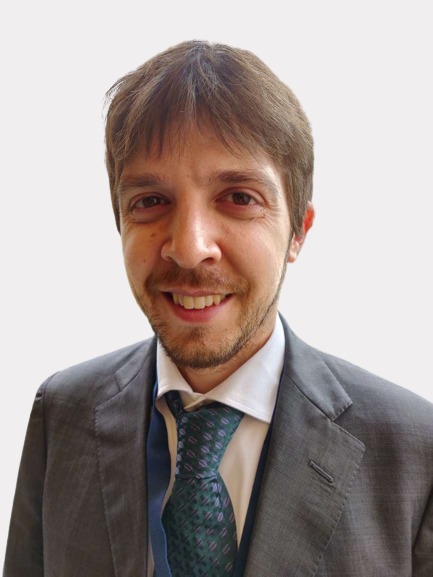}}]{Claudio A. Ardagna} is Full Professor at the Università degli Studi di Milano, the Director of the CINI National Lab on Big Data, and co-founder of Moon Cloud srl. His research interests are in the area of cloud-edge security and assurance, and data science. He has published more than 140 contributions in international journals, conference/workshop proceedings, and chapters in international books. He has been visiting professor at Universit\'e Jean Moulin Lyon 3 and visiting researcher at Beijing University of Posts and Telecommunications, Khalifa University, George Mason University. He is member of the Steering Committee of IEEE TCC, member of the editorial board of the IEEE TCC and IEEE TSC, and secretary of the IEEE Technical Committee on Services Computing.
\end{IEEEbiography}

\vskip -2\baselineskip plus -1fil

\begin{IEEEbiography}[{\includegraphics[width=1in,height=1.25in,clip]{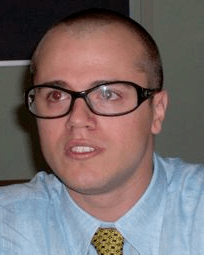}}]{Alessandro Balestrucci} is a postdoctoral researcher at CINI. His research interests are in the area of security of intelligent systems and applied artificial intelligence to disinformation prevention and social network analysis. He got his PhD in Computer Science at Gran Sasso Science Institute and qualification of IT engineer at Ca'Foscari University of Venice. He has participated/is participating to national and European projects (H2020) cooperating with IMT Lucca, CNR and University of Milan. 
\end{IEEEbiography}

\vskip -2\baselineskip plus -1fil

\begin{IEEEbiography}[{\includegraphics[width=1in,height=1.25in,clip]{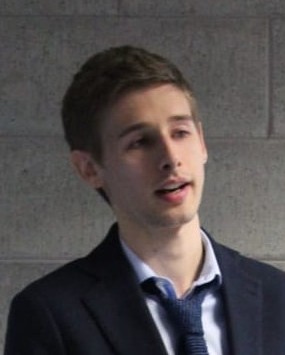}}]{Nicola Bena} is a Ph.D. student at the Università degli Studi di Milano. His research interests are in the area of security of modern distributed systems with particular reference to certification, assurance, and risk management techniques. He has participated/is participating to several national and European projects, including H2020 Project CONCORDIA, one of the four European projects aimed to establish the European Cybersecurity Competence Network. He has been visiting scholar at Khalifa University, UAE, and INSA Lyon, France.
\end{IEEEbiography}

\vskip -2\baselineskip plus -1fil

\begin{IEEEbiography}[{\includegraphics[width=1in,height=1.25in,clip]{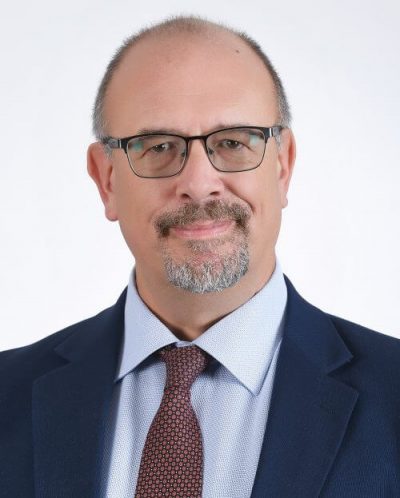}}]{Ernesto Damiani} is Full Professor at the Department of Computer Science, Università degli Studi di Milano, where he leads the Secure
Service-oriented Architectures Research (SESAR) Laboratory. He is also the Founding Director of the Center for Cyber-Physical Systems, Khalifa University, United Arab Emirates. He received an Honorary Doctorate from Institute National des Sciences Appliquées de Lyon, France, in 2017, for his contributions to research and teaching on big data analytics.  He serves as Editor in Chief for IEEE Transactions on Services Computing. His research interests include cybersecurity, big data, and cloud/edge processing, and he has published over 680 peer-reviewed articles and books. He is a Distinguished Scientist of ACM and was a recipient of the 2017 Stephen Yau Award.
\end{IEEEbiography}

\vskip -2\baselineskip plus -1fil

\begin{IEEEbiography}[{\includegraphics[width=1in,height=1.25in,clip]{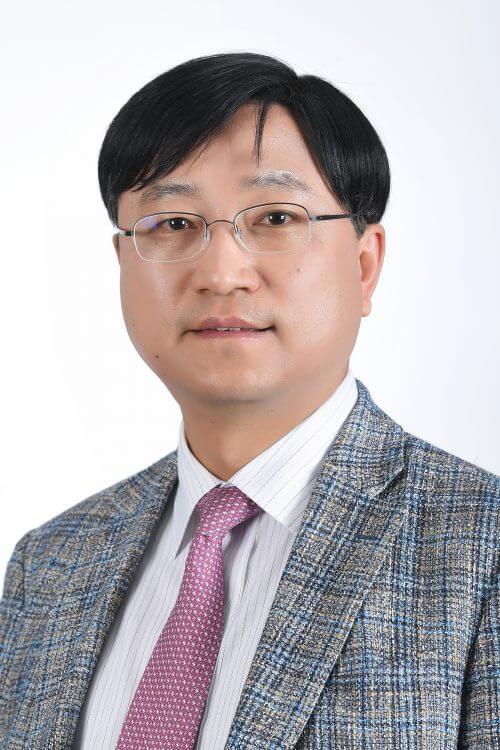}}]{Chan Yeob Yeun} (Senior Member, IEEE) received the M.Sc. and Ph.D. degrees in information security from Royal Holloway, University of London, in 1996 and 2000, respectively. After his Ph.D. degree, he joined Toshiba TRL, Bristol, U.K., and later, he became the Vice President at the Mobile Handset Research and Development Center, LG Electronics, Seoul, South Korea, in 2005. He was responsible for developing mobile TV technologies and related security. He left LG Electronics, in 2007, and joined ICU (merged with KAIST), South Korea, until August 2008, and then the Khalifa University of Science and Technology, in September 2008. He is currently a Researcher in cybersecurity, including the IoT/USN security, cyber-physical system security, cloud/fog security, and cryptographic techniques, an Associate Professor with the Department of Electrical Engineering and Computer Science, and the Cybersecurity Leader of the Center for Cyber-Physical Systems (C2PS). He also enjoys lecturing for M.Sc. degree in cyber security and Ph.D. degree in engineering courses at Khalifa University. He has published more than 140 journal articles and conference papers, nine book chapters, and ten international patent applications. He also works on the editorial board of multiple international journals and on the steering committee of international conferences.
\end{IEEEbiography}

\balance

\end{document}